\documentclass[10pt,twocolumn,letterpaper]{article}

\usepackage{cvpr}              % To produce the CAMERA-READY version

\usepackage{comment}
\usepackage[accsupp]{axessibility}

\definecolor{amethyst}{rgb}{0.6, 0.4, 0.8}
\definecolor{darkpastelgreen}{rgb}{0.01, 0.75, 0.24}
\definecolor{amber}{rgb}{1.0, 0.75, 0.0}
\definecolor{cadmiumorange}{rgb}{0.93, 0.53, 0.18}
\definecolor{lawngreen}{rgb}{0.49, 0.99, 0.0}
\definecolor{limegreen}{rgb}{0.2, 0.8, 0.2}
\definecolor{neongreen}{rgb}{0.22, 0.88, 0.08}
\definecolor{amethyst}{rgb}{0.6, 0.4, 0.8}
\definecolor{darkpastelgreen}{rgb}{0.01, 0.75, 0.24}
\definecolor{greenbest}{RGB}{88,137,15}
\definecolor{redworst}{rgb}{0.83, 0.3, 0.30}
\definecolor{royalazure}{rgb}{0.25, 0.41, 0.88}

\usepackage{graphicx} % Required for inserting images
\usepackage{booktabs} % For \toprule, \midrule, etc.
\usepackage{multirow}
\usepackage{xcolor}
\usepackage[table]{xcolor}
\usepackage{pifont} % For \ding symbols
\newcommand{\cmark}{\textcolor{green!60!black}{\ding{51}}}
\newcommand{\xmark}{\textcolor{red!70!black}{\ding{55}}}

\usepackage{float}

\makeatletter
\def\blfootnote{\gdef\@thefnmark{}\@footnotetext}
\makeatother

\definecolor{cvprblue}{rgb}{0.21,0.49,0.74}
\usepackage[pagebackref,breaklinks,colorlinks,allcolors=cvprblue]{hyperref}

\definecolor{gold}{RGB}{255,179,179}
\definecolor{silver}{RGB}{255,217,179}
\definecolor{bronze}{RGB}{255,255,179}
\newcommand{\G}[1]{\cellcolor{gold}{#1}}   % 1st place + bold
\newcommand{\Z}[1]{\cellcolor{silver}{#1}} % Silver = 2nd place
\newcommand{\B}[1]{\cellcolor{bronze}{#1}}

\def\paperID{9341} % *** Enter the Paper ID here
\def\confName{CVPR}
\def\confYear{2026}

\newcommand{\Bcol}[1]{\textcolor{teal}{#1}}
\newcommand{\Ecol}[1]{\textcolor{red}{#1}}
\newcommand{\Acol}[1]{\textcolor{orange}{#1}}

\newcommand{\methodname}{
    \textbf{\Bcol{B}\Ecol{E}\Acol{A}}-GS}

\newcommand{\methodnametable}{
    \textbf{\Bcol{B}\Ecol{E}\Acol{A}}-GS~}

\title{
\textbf{\methodname: \Bcol{B}\Ecol{E}yond R\Acol{A}diance Supervision in 3DGS\\for Precise Object Extraction}}

\author{
    \begin{tabular}{c c c}
        Alessio Mazzucchelli$^{1,2}$ & Maria Naranjo-Almeida$^{1}$ & Jorge Bustos-Sanchez$^{1}$
    \end{tabular}
    \vspace{0.1cm} \\
    \begin{tabular}{c c c c}
        Mariella Dimiccoli$^{2}$ & Francesc Moreno-Noguer$^{2,*}$ & Jordi Sanchez-Riera$^{2}$ &  Adrian Penate-Sanchez$^{3}$
    \end{tabular}
    \vspace{0.3cm} \\
    \small $^{1}$Arquimea Research Center, $^{2}$Institut de Robòtica i Informàtica Industrial (CSIC-UPC) \\ [-0.12cm] \small $^{3}$Universidad de las Palmas de Gran Canaria (IUSIANI)
}

\begin{document}

\twocolumn[{%
    \renewcommand\twocolumn[1][]{#1}%
    \maketitle
    \begin{center}
        \vspace{-3mm}
        \includegraphics[width=1.0\linewidth, trim=2mm 0 1mm 6mm, clip]{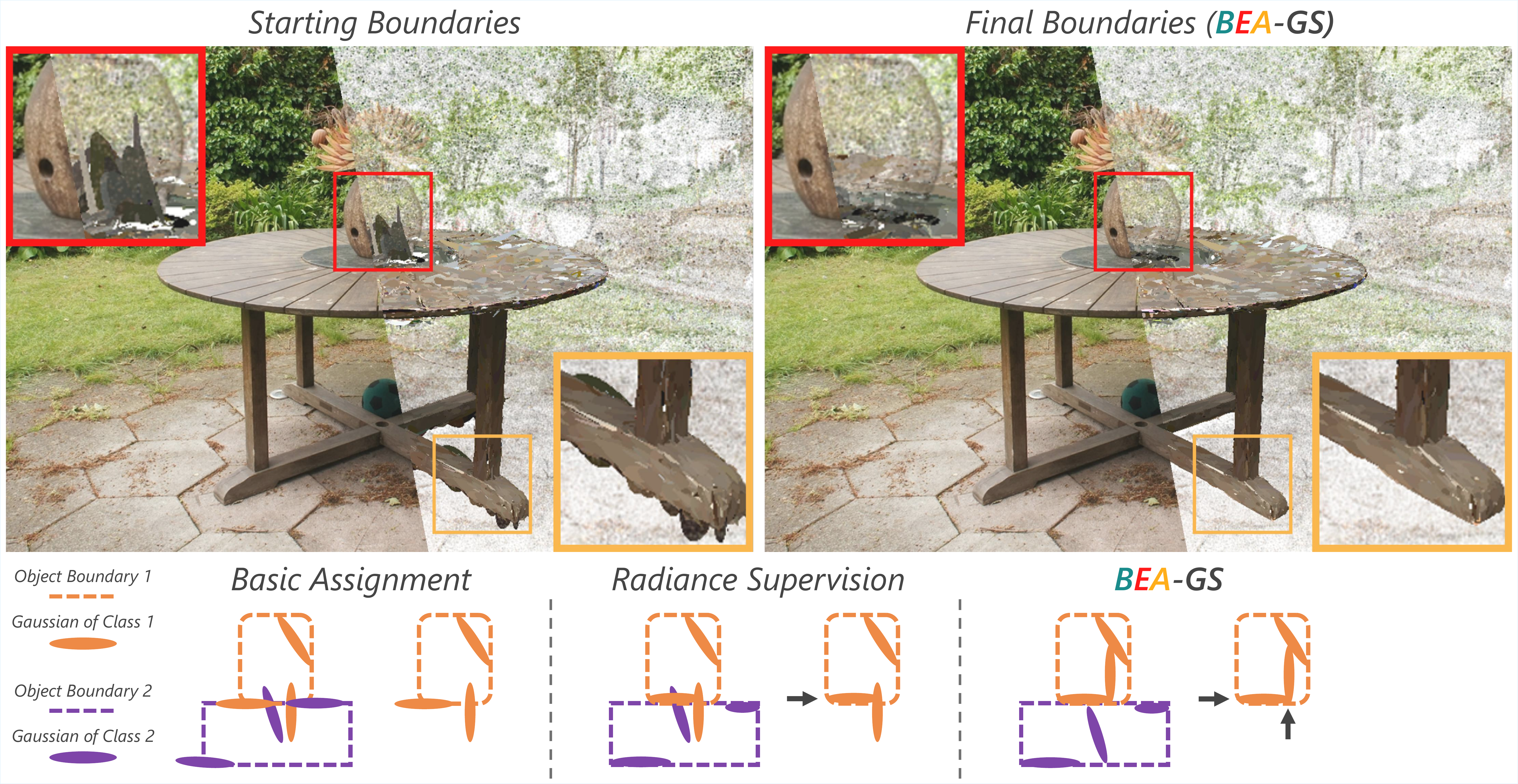}
        \vspace{-6mm}
    	\captionof{figure}{\textit{Top row}: Lifting 2D semantic segmentations to 3DGS scenes often produces incorrect object boundaries due to scenes being constructed as a single entity (left). \textit{Bottom row}: Early approaches freeze the geometry and only assign semantic labels to existing Gaussians, often leading to inaccurate object extractions (bottom left). This limitation is partially mitigated by radiance-based supervision that allows visible Gaussians to be refined (bottom center). In contrast,\methodname~proposes a novel solution that optimizes both seen and partly unseen Gaussians to correctly represent the boundaries between objects.}
		\label{fig:teaser}
	\end{center}
    \vspace{2mm}
	}
]

\thispagestyle{empty}

\maketitle

\blfootnote{$^*$work started before joining Amazon}

\begin{abstract}
%\vspace{-3mm}

Most Gaussian Splatting techniques that provide a 3D semantic representation of the scene do not optimize the underlying 3D geometry, making object-level editing or asset extraction challenging. Recent methods, such as~\cite{COBGS, Trace3D, ObjectGS}, acknowledge this limitation and propose approaches that modify the scene's geometry to represent the underlying semantics. We advance this concept  further by proposing a novel solution that provides near perfect boundaries in object extraction. We do so by introducing two new losses in the optimization that take care of: 1) a loss that modifies the geometry of visible Gaussians to respect semantic boundaries, and 2) a loss that adjusts the geometry of non-visible Gaussians that appear once the object is extracted. Our first loss propagates gradients directly through the rasterization, allowing for seamless integration within the optimization of the Gaussian parameters. The second loss also propagates gradients to Gaussian parameters but does so without passing through the rasterization, enabling modification of the scene's geometry even when little transmittance reaches a Gaussian (partial or non-visible). Exhaustive comparisons with 12 state of the art methods across 4 datasets, using six metrics, demonstrate that our approach produces overall the best boundary segmentation to date.

\end{abstract}    
\vspace{-3mm}
\section{Introduction}
\label{sec:intro}

3D Gaussian Splatting (3DGS) ~\cite{kerbl3Dgaussians} has emerged as an effective way of building realistic 3D models from just a set of real-world images. While this eliminates the need for manual 3D modeling, applications based on 3DGS lack inherent support for object-level editing and asset extraction.
To address these limitations, several approaches~\cite{gaussian_grouping,  flashsplat, wu2024opengaussian, DrSplat, jain2024gaussiancut, Unilift, ILGS} have integrated 2D Vision Foundation Models, such as SAM~\cite{SAM, SAM2} and CLIP~\cite{clip2021}, into the 3DGS representation, enabling object-level editing and semantic understanding of scenes. However, these methods disregard the underlying scene's geometry and, as observed by~\cite{ClickGaussian}, Gaussians learned without semantic constraints often contribute to multiple objects, hindering semantic learning. To overcome this challenge, recent methods, \cite{COBGS}, \cite{Trace3D}, and \cite{ObjectGS}, have introduced  semantic constraints to modify the geometry, ensuring that each Gaussian contributes only to a single object. 
While this approach was expected to resolve such issues, we observed that during object extraction, some Gaussians still extend beyond their intended boundaries. This seemingly minor issue becomes critical when extracting and reusing objects in new scenes, as these hidden Gaussians become visible and create unexpected artifacts as seen in Fig~\ref{fig:teaser}. In this work, we analyze the root causes of this phenomenon.

Previous methods rely on the alpha-blending rasterizer to propagate semantic changes to the Gaussians, which means that only Gaussians that receive a certain amount of radiance are modified.  Gaussians with minimal radiance reception cannot be altered through the gradients of alpha-blending.  These partially visible Gaussians remain "stuck" beneath  object surfaces and appear when extracting objects from a scene. This occurs because, unlike NeRF \cite{mildenhall2020nerf, barron2022mipnerf360, InstantNGP}, where geometry and appearance are mostly decoupled, in 3DGS these aspects are tightly linked, as described in~\cite{Chao_2025_CVPR}. Each Gaussian's appearance depends directly on its shape and size, so when trained only with a photometric loss, it may stretch or drift into nearby or unseen regions to better match the observed color, rather than  simply adjusting its appearance parameters.

Motivated by these observations, we propose a Gaussian Splatting framework that learns a robust Gaussian arrangement resilient to object extraction. Our approach combines a 2D boundary loss, that penalizes cross-class contributions for visible Gaussians, with a 3D occupancy/visibility loss that regularizes non-visible or partially visible Gaussians. This dual optimization strategy effectively constrains Gaussians that don't contribute to scene rendering, resulting in a more stable structure for object extraction. Our main contributions are: 

\begin{itemize}
\item  A boundary rasterization loss that operates directly with the rasterizer, requiring only a single additional channel even with multiple classes. The boundary loss enforces visible Gaussians to respect semantic boundaries between objects, thus, creating clean boundaries between them.
\item An occupancy loss that refines the underlying geometry of objects receiving minimal or no radiance. This loss addresses partially visible Gaussians that the boundary loss cannot handle, as the latter is based on radiance contributions.

\item An extensive evaluation across 4 datasets, 6 metrics, and 12 state-of-the-art methods under identical 2D segmentation masks.\methodname~consistently achieves the best results over all datasets and metrics. %We believe this shows the proposed approach's robustness, becoming a useful contribution to the community.

\end{itemize}

\section{Related Work}
\label{sec:related}

\begin{figure*}[t]
    \centering
    \includegraphics[width=\textwidth]{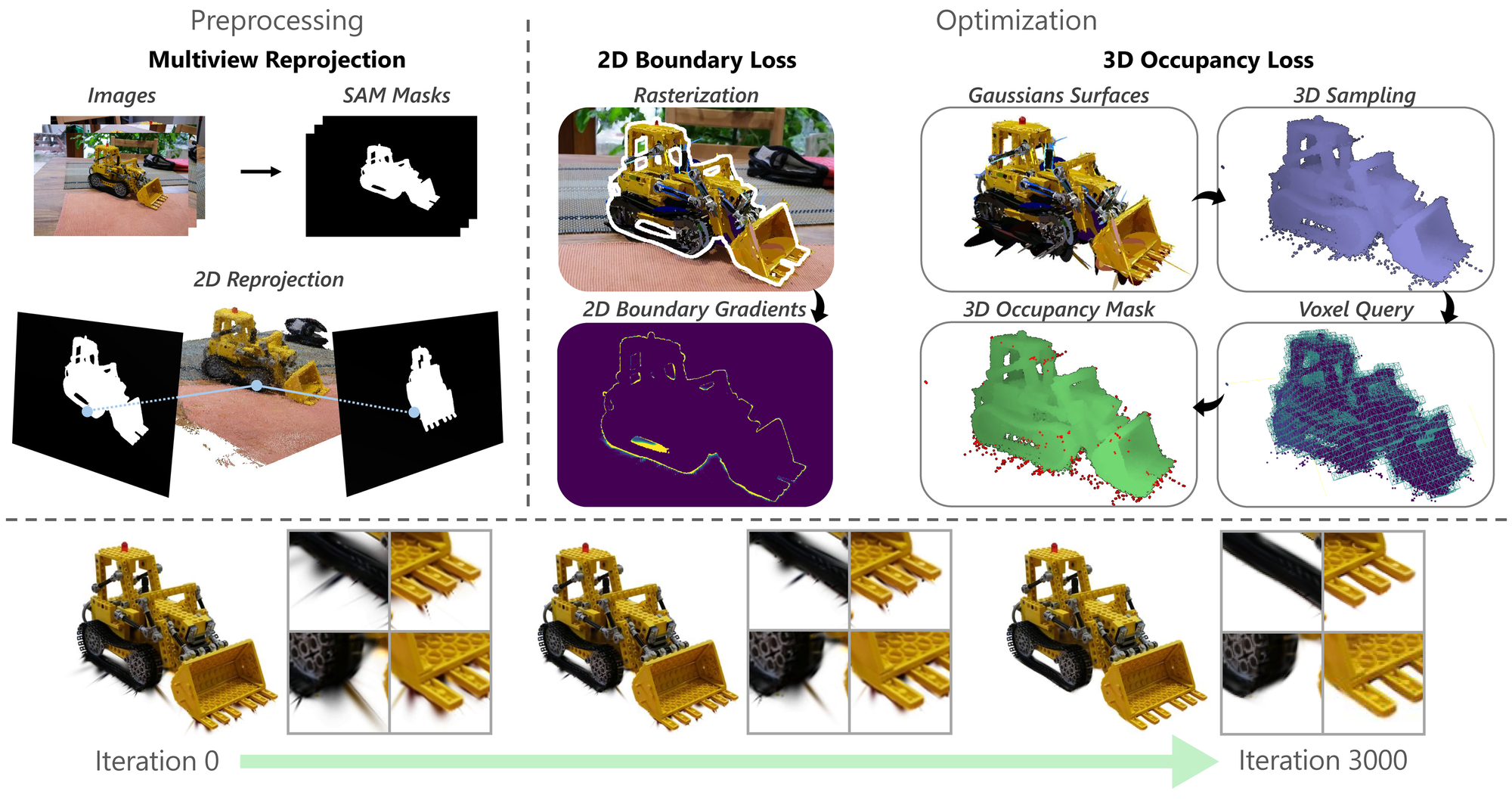}
    \vspace{-8mm}
    \caption{Main Diagram: Given a pretrained 2DGS scene and two-stage segmentations masks obtained using SAM2~\cite{SAM2} our \textbf{Multiview Reprojection} improves mask consistency between images, see Sec.~\ref{ssec:labels}. Once masks have been corrected, we begin optimization for $3000$ iterations. Our optimization applies our \textbf{2D Boundary Loss} to improve visible object boundaries, see Sec.~\ref{ssec:semantic}, and our voxel \textbf{3D Occupancy Loss} to correct non-visible Gaussians, see Sec.~\ref{ssec:occupancy}.  }
    \label{fig:main_diagram}
    \vspace{-0mm}
\end{figure*}

\textbf{3D Gaussian Splatting}. 3D Gaussian Splatting (3DGS) methods~\cite{kerbl3Dgaussians, Huang2DGS2024, scaffoldgs, 3dgrt2024, wu20253dgut, MonteCarloGS} have become the de facto option for novel view synthesis of 3D scenes. Gaussian splatting offers an automatic approach to obtain realistic 3D representations from just a set of images. In order for such scenes to be used by 3D artists, they need to adapt to the changing requirements of media production. This requires gaussian splatting scenes to be capable of operations like 3D object removal~\cite{AG2aussian, gaussianeditor_cvpr_2024, zhang20243DitScene}, inpainting~\cite{Huang_2025_CVPR, SplatFill, ZHOU2025104362}, and more generally editing~\cite{mirzaei2024reffusionreferenceadapteddiffusion, wang2024gscream, gaussctrl2024, jaganathan24iceg, guedon2024frosting, chen2024dge, wang2024view, Kim_2025_ICCV, mazzucchelli2026virgi}. The main challenge to unlock the edition capabilities of this technology is to obtain a segmentation of the 3D scene that defines where one object finishes and another begins.

\vspace{1mm}
\noindent \textbf{Semantic Feature Learning}. Early methods~\cite{qin2023langsplat, zhou2024feature, chacko2025wacv, cen2023saga, shi2024language} address the challenge of correlating the 2D masks extracted by SAM~\cite{SAM} across different views, primarily by learning a suitable feature representation. One of the first solutions, LangSplat~\cite{qin2023langsplat}, leveraged CLIP~\cite{clip2021} features to establish correspondences between multiview masks. By attaching language-aligned features to each Gaussian they enable rendering-based supervision, using SAM for boundary alignment and CLIP for cross-view semantic consistency during optimization. However, as noted by OpenGaussian~\cite{wu2024opengaussian}, while such approaches perform well in 2D pixel-level segmentation, the alpha-blending process in 3D Gaussian Splatting produces weakly expressive features and inaccurate 2D–3D associations. They achieve strong 2D results but struggle in 3D point-level understanding.

To overcome these limitations, methods such as~\cite{wu2024opengaussian, InstaScene, ClickGaussian, PanoGS, FMGS_Zuo_IJCV2024} focus on enhancing the expressiveness of 3D features. OpenGaussian~\cite{wu2024opengaussian}, for example, discretizes feature space via a learned codebook so that each Gaussian’s feature aligns closely with the correct semantic embedding. Other works propose alternative mechanisms to enrich 3D feature representations (e.g., spatial contrastive learning in InstaScene~\cite{InstaScene} or global feature candidates in ClickGaussian~\cite{ClickGaussian}) to achieve more distinguishable features. 

Another family of methods~\cite{lyu2024gaga, flashsplat, DrSplat, jain2024gaussiancut} demonstrates that segmentation can be tackled without the need for feature optimization. GaussianCut~\cite{jain2024gaussiancut}, starting from ID-correlated masks, represents the scene as a graph and then applies a graph-cut algorithm to minimize an energy function, effectively classifying the Gaussians. In contrast, Dr. Splat~\cite{DrSplat} and FlashSplat~\cite{flashsplat} show that Gaussian association can be achieved by leveraging radiance information. Dr. Splat assigns CLIP features to Gaussians based on their radiance contribution, while FlashSplat, given ID-correlated masks, reformulates the problem as an Integer Linear Programming task.

\vspace{1mm}
\noindent \textbf{Adapting Geometry}.  While assigning correct features or labels to each Gaussian without modifying geometry is important,  it relies on a 3DGS structure that was built without semantic information.  Recognizing these limitations, several methods~\cite{gaussian_grouping, Unilift, ILGS, COBGS, Trace3D, ObjectGS, SAGD} incorporate geometry modification into their optimization  process. Among these, Gaussian Grouping~\cite{gaussian_grouping}, and works built up on it like UniLift~\cite{Unilift} and ILGS~\cite{ILGS}, rely on a multi-layer perceptron to classify alpha-blended features. This classification then guides the refinement of Gaussian geometry to better align with supervised object masks. However, despite their effectiveness in 2D segmentation, these approaches share the limitation highlighted by OpenGaussian: relying on classification of 2D rendered features does not translate effectively to 3D understanding of Gaussians.

In contrast, ~\cite{COBGS, ObjectGS} focus on directly learning each Gaussian’s label within the 3D space. Each Gaussian is assigned a probability of belonging to a certain class, allowing classification to occur directly in 3D. Building on this idea, these methods also explicitly address the boundary problem in 3D Gaussian Splatting by enforcing that each Gaussian may contribute to only a single object. ObjectGS~\cite{ObjectGS} and COB-GS~\cite{COBGS} use the rendered probability maps as supervision to refine Gaussian boundaries. Additionally, COB-GS~\cite{COBGS} introduces a heuristic to detect Gaussians with ambiguous boundaries, those receiving conflicting optimization signals from multiple objects, and splits them accordingly. Similarly, Trace3D~\cite{Trace3D} performs geometry refinement by analyzing each Gaussian’s contribution to the image space, splitting or pruning those that contribute to multiple object masks. Despite their differences in design choices and optimization strategies, all these methods share the same fundamental limitation: they operate within the constraints imposed by alpha-blending, and therefore by the radiance absorbed by each Gaussian. Every decision, whether gradient-based or heuristic, is ultimately governed by radiance. A Gaussian, or a part of a Gaussian, that is not visible remains unaffected during optimization. Consequently, when performing object extraction, these methods often produce artifacts such as Gaussian spikes from regions that received little or no radiance supervision during training. This is precisely the gap our method aims to bridge.

\begin{figure}[t]
    \centering
    \includegraphics[width=\columnwidth]{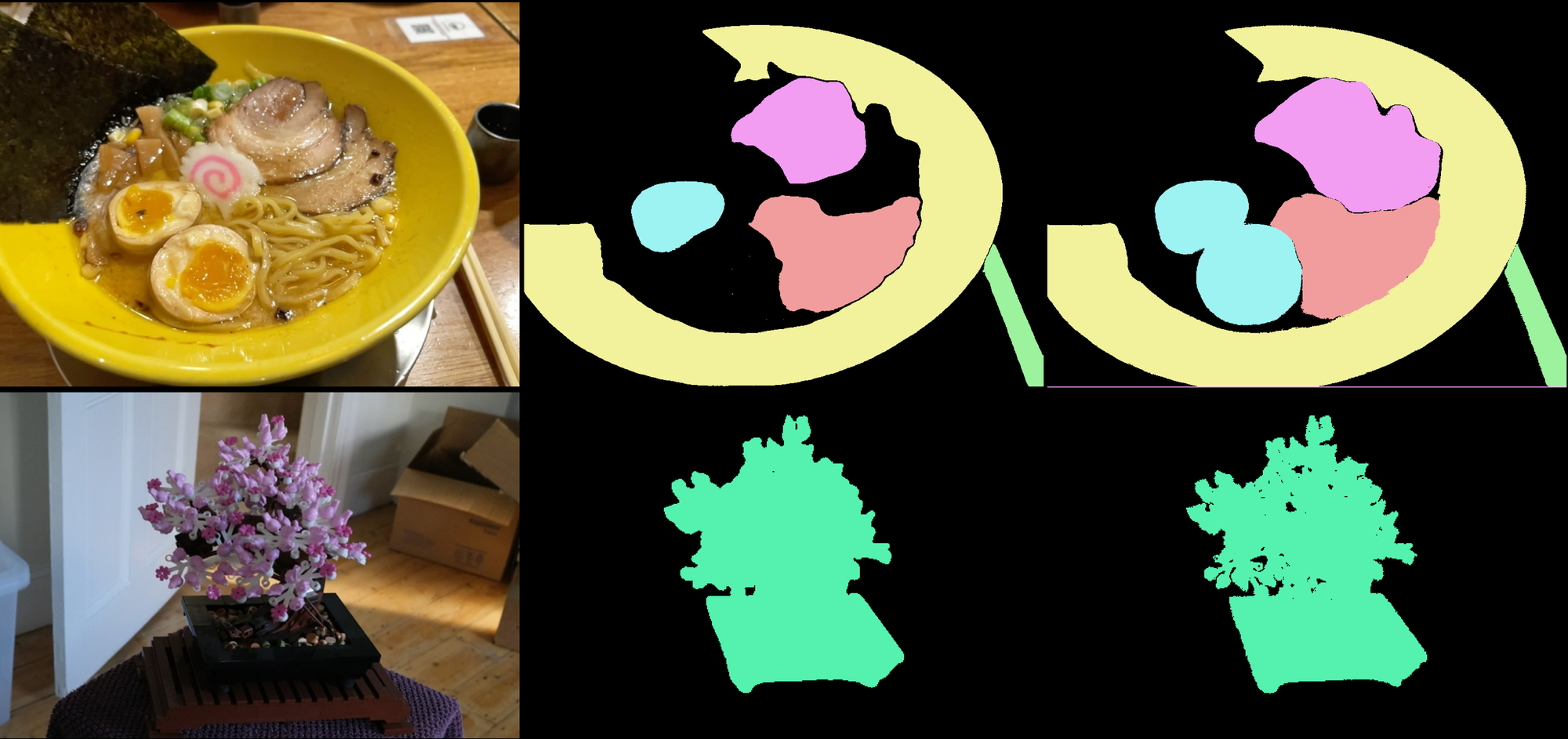}
    \vspace{-5mm}
    \caption{\textbf{Multiview Reprojection}. \textit{First column:} original RGB images. \textit{Second column:} two-stage segmentation masks obtained using SAM2~\cite{SAM2}. \textit{Third column:} improved segmentation masks using our multiview reprojection. The first row shows how our approach incorporates regions that were missed. The second row shows that, by understanding the 3D structure of the scene, our approach yields much more fine grained details.}
    \label{fig:reprojection_example}
    \vspace{-0mm}
\end{figure}

\section{Method}
\label{sec:method}

This section presents our method, illustrated in Fig.~\ref{fig:main_diagram}.  We begin in Section~\ref{ssec:background} with a description of 2D Gaussian Splatting~\cite{Huang2DGS2024} (2DGS). %Section~\ref{ssec:labels} outlines our approach to generating initial object masks for initializing the 2DGS framework. 
Section~\ref{ssec:labels} describes our proposed approach to obtain segmentation masks that are multiview consistent. Next, in Sections~\ref{ssec:semantic} and~\ref{ssec:occupancy}, we introduce our boundary and occupancy loss terms, which ensure Gaussians remain within their assigned object boundaries, even when Gaussians are partially occluded. Finally, Section \ref{ssec:final_loss}, presents the overall objective loss function.

\subsection{Background}
\label{ssec:background}

We reconstruct scenes using 2DGS from posed RGB images. The method uses $N$ 2D Gaussians embedded in 3D space, each parameterized by: center point $p_i$, 
orientation $R_i$, scale $s_i$, color $c_i$, and opacity $\alpha_i$. The color of a pixel $u$ is computed by splatting Gaussians: 
\begin{align}
    c(u) &= \sum_{i=1}^{N} c_i \alpha_i \hat{\mathcal{G}}_i(x) 
    \prod_{j}^{i-1} \left(1 - \alpha_j \hat{\mathcal{G}}_j(x)\right)\;,
\end{align}
where $\hat{\mathcal{G}}_i(x)$ is the density of the $i$-th Gaussian at 3D position $x$, derived from its position, orientation, and scale ${(p_i, R_i, s_i)}$. It additionally incorporates depth distortion $\mathcal{L}_{depth}$ and normal consistency $\mathcal{L}_{norm}$ terms to enhance reconstruction quality. The complete loss function is, $\mathcal{L}_{\text{2DGS}} = \mathcal{L}_{\text{rgb}} + \lambda_{depth}\,\mathcal{L}_{depth} + \lambda_{norm}\,\mathcal{L}_{norm}$ where $\lambda_{depth}$ and $\lambda_{norm}$ are weighting coefficients.

\subsection{Multiview Reprojection}
\label{ssec:labels}

For each training image $I$ we obtain 2D semantic masks $M_\phi$, where $\phi$ in $\{1..L\}$ indexes the semantic class, and $L$ is the total number of classes in the scene. We use the SAM2~\cite{SAM2} two-stage mask retrieval described in~\cite{COBGS} to minimize errors due to tracking being lost. While this alleviates issues related to tracking, it does not address boundary inconsistencies across views. To tackle this, we propose an approach that aims at enforcing the consistency of segmentation boundaries across views. First, we generate a 3D point cloud from the pretrained model using 2DGS depth and camera parameters. Afterwards, we lift each 2D pixel $u \in \mathbb{R}^2$ to a 3D point $x \in \mathbb{R}^3$ and associate it with its corresponding semantic label from the initial mask. Finally, we reproject these points into all available views and take the most frequent label per pixel, to obtain a refined set of semantic masks $M'=\textrm{argmax}(M_\phi)$. This approach improves the reliability of the segmentation as shown in Fig.~\ref{fig:reprojection_example}.

\subsection{2D Boundary Loss}
\label{ssec:semantic}

In order to obtain a boundary-aware Gaussian scene from the refined masks, we first need to extend each Gaussian with an additional parameter $\phi$, that tells us the class to which it belongs. This parameter is initialized using the refined masks $M'$, following the procedure described in~\cite{flashsplat}. We perform splatting over all training images and we accumulate the frequency with which each Gaussian contributes to pixels of a given class. The accumulation is weighted by the Gaussian’s contribution, yielding a more robust and consistent semantic assignment. Gaussian initialization finishes by assigning to each Gaussian the class to which it contributes the most. Importantly, our class parameter is kept non-trainable, enforcing that Gaussians adapt to the scene by adjusting their position, shape or opacity rather than altering their assigned class.

To ensure that each Gaussian subsequently contributes only to pixels of its class, we introduce a boundary loss that penalizes Gaussians contributing to pixels of different classes. Specifically, given $N$ Gaussians projected onto a pixel $u$, each Gaussian indexed by $i$ is characterized by a class $\phi_i$, an opacity value $\alpha_i$ and a density $\hat{\mathcal{G}_i}(x)$ at 3D position $x$. For the same pixel $u$, the segmentation mask value $M'(u)$ serves as supervision during training. The boundary loss is defined as:

\vspace{-6mm}
\begin{gather}
\begin{aligned}
    \mathcal{L}_{bound}(u) &= \sum_{i=1}^{N} H_i(u) \alpha_i \hat{\mathcal{G}}_i(x) \prod_{j=1}^{i-1} (1 - \alpha_j \hat{\mathcal{G}}_j(x)) \\
    H_i(u) &= 
    \begin{cases}
        0,  & \text{if } \phi_i = M'(u)\\
        1,  & \text{if } \phi_i \neq M'(u)
    \end{cases}
\label{eq:boundary_loss}
\end{aligned}
\end{gather}

When a Gaussian contributes to a pixel of a different class, the loss rewards reducing its opacity, changing its shape and/or shifting it away from the boundary. Other methods also propagate their loss through the rasterization like COB-GS~\cite{COBGS} and ObjectGS~\cite{ObjectGS}. We now highlight the key differences. Our approach naturally supports multi-class segmentation, whereas COB-GS is inherently restricted to binary extraction and must be optimized separately for each object. ObjectGS supports multi-class optimization but requires an additional per-class channel for each Gaussian, leading to increased memory usage and higher computational cost during rendering and backpropagation. Instead, our approach only requires a single additional channel in each Gaussian. Another critical difference is that both COB-GS and ObjectGS apply their loss over the entire segmentation mask, while our loss is only activated by parts of Gaussians that are outside of boundaries, as seen in Fig.~\ref{fig:main_diagram} under \textit{"2D Boundary Gradients"}.

\begin{figure}[t]
    \centering
    \includegraphics[width=\columnwidth, trim={3mm 3mm 3mm 3mm}, clip]{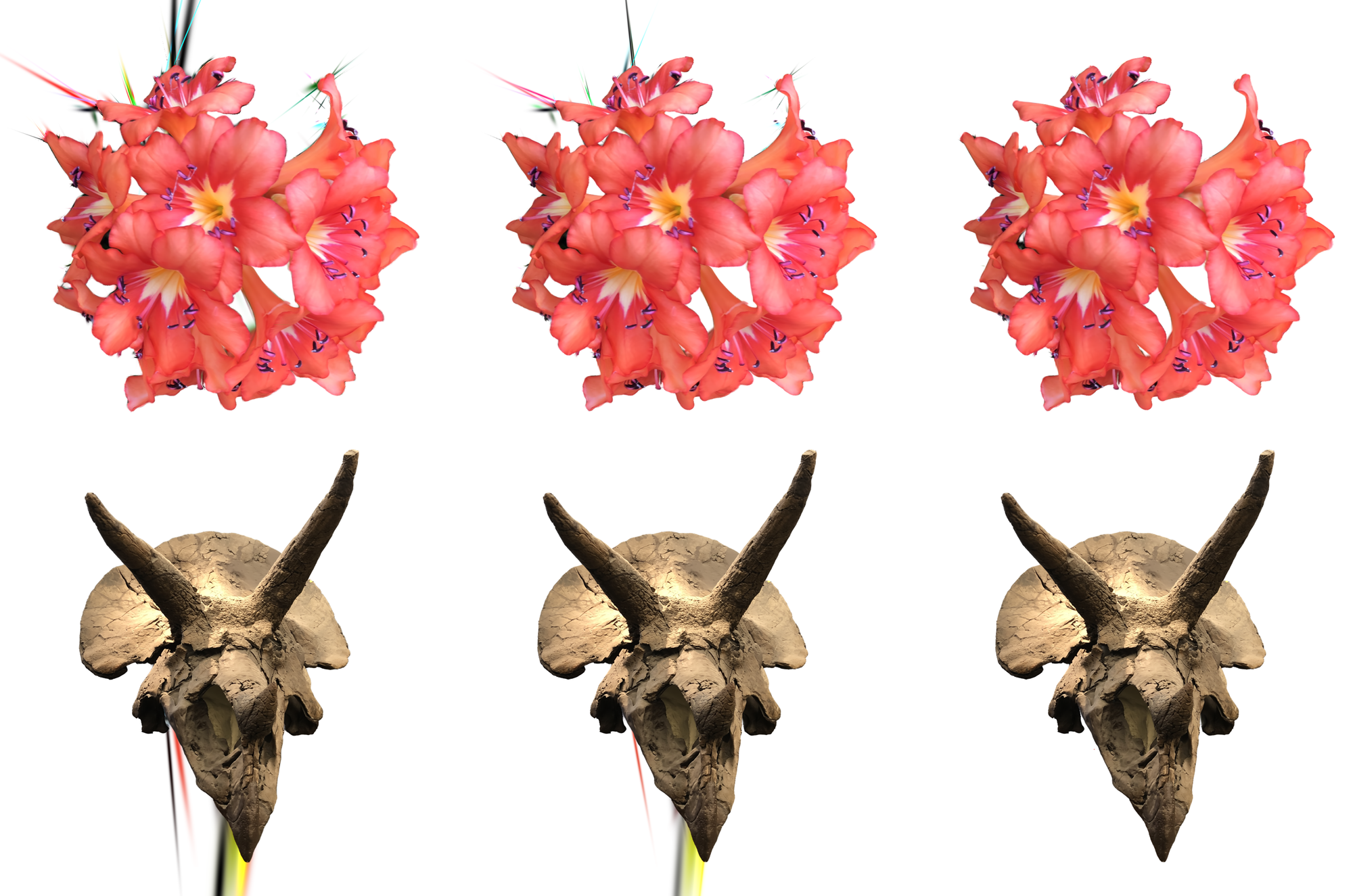}
    \vspace{-4mm}
    \caption{\textbf{Optimization Losses}. \textbf{First column:} original Gaussians assignment. \textbf{Second column:} Gaussians optimized using only the 2D Boundary Loss. \textbf{Third column:} Gaussians optimized with both the 2D Boundary Loss and the 3D Occupancy Loss.}
    \label{fig:split_contributions_figure}
    \vspace{-1mm}
\end{figure}

\subsection{3D Occupancy Loss}
\label{ssec:occupancy}

Up to this stage our regularization term, similar to previous methods~\cite{COBGS, Trace3D, ObjectGS}, relies on transmittance $\prod_{j=1}^{i-1} (1 - \alpha_j \hat{\mathcal{G}}_j(x))$, to propagate the desired changes to the Gaussians. Since the formulation is conditioned on transmittance, it affects only the visible parts of Gaussians. To address artifacts in non-visible regions that appear during object extraction, we need to introduce a new regularization term that is not limited by radiance. As shown in Fig.~\ref{fig:split_contributions_figure}, just applying the boundary loss $\mathcal{L}_{bound}$ does not lead to a perfect extraction. To address this, we sample the 2D surface of each Gaussian and determine if these samples correspond to valid geometry, or, to unseen areas that need to be penalized. This sampling is performed uniformly in the Gaussian UV space and each unseen sample is penalized proportionally to its density. To detect unseen areas, we construct a depth-based proxy of the scene geometry.

\vspace{1mm}
\noindent{\bf{Visibility Voxel Grid.}}
For each training image $I$ we backproject, using the rendered depth, all pixels of the semantic rendering obtained by applying alpha-blending to the semantic class $\phi$ of each Gaussian. We then merge all points with the same class, from all point clouds, into class-specific point clouds $P_\phi = \{ x \mid x \in P,\, \text{class}(x)=\phi \}$, representing the spatial support of class $\phi$ in the scene. To enable efficient queries during training, each $P_\phi$ is voxelized into a grid $V_\phi$. Each voxel grid $V_\phi$ acts as a geometric prior on visibility to detect regions that need to be penalized. The grids $V_\phi$ are only computed once before the optimization and voxels where several classes collide are considered empty. To reliably distinguish valid geometry from unseen regions, it is essential to define an appropriate voxel size $s$ for the spatial discretizations.

\vspace{1mm}
\noindent{\bf{Adaptive Voxel Size.}}
We wish to obtain a voxel size $s$ that is data-driven and tailored to each scene. To do so, we want the value of $s$ that contains $k$ points on average. This can be defined as $\rho = \frac{k}{s^3}$, where $\rho$ is the density and $s^3$ the volume of a voxel of side $s$. In order to find $s$ we need to establish the density $\rho$. Following~\cite{Loftsgaarden1965ANE, NIPS2014_a5549f3f}, we calculate the local spatial density $\rho$ from the average distance of each point to it's $k$-th nearest neighbor $d_k$ as, $\rho = k / (\tfrac{4}{3}\pi d_k^{3})$. We then solve $s = \sqrt[3]{(k / \rho)}$, obtaining the value of $s$ for the calculated density $\rho$, thus adapting the voxel size to the spatial density.

\vspace{1mm}
\noindent{\bf{Occupancy Loss.}}
During training we uniformly sample $Z$ points $x'_r$ with $r \in =\{1..Z\}$ from each Gaussian. For each sampled point we query the voxel grid $V_\phi$, were $\phi$ is the class of the Gaussian from which the point was sampled. We then check whether the voxel position that contains $x'_r$, or any of its neighbors, is occupied. The loss is defined as:

\vspace{-4mm}
\begin{gather}
\begin{aligned}
    \mathcal{L}_{occ} &= \sum_{i=1}^{N} \sum^{Z}_{r=1} \frac{\alpha_i \hat{\mathcal{G}}_i(x'_r) Q_i(x'_r, V_\phi)}{N \cdot Z} \\
    Q_i(x'_r, V_\phi) &= 
    \begin{cases}
        1,  & \text{if } D_{occ}(x'_r, V_\phi) = 0\\
        0,  & \text{if } D_{occ}(x'_r, V_\phi) > 0
    \end{cases}
\label{eq:occ_loss}
\end{aligned}
\end{gather}
where $Q_i()$ is the function that queries the voxel grid $V_\phi$. We define $D_{occ}(x'_r, V_\phi)$ as the number of occupied voxels in the neighborhood of the voxel containing $x'_r$ within $V_\phi$. $D_{occ}$ ranges between $\{0..27\}$ as it counts the neighbors and the voxel itself. If none of those voxels are occupied, $D_{occ}(x'_j, V_\phi) = 0$, we consider $x'_r$ to be a non-visible point. Each sample is penalized according to its opacity $\alpha_i$ and density of the Gaussian at that position $\hat{\mathcal{G}}(x'_r)$. With $\mathcal{L}_{occ}$, we have now obtained a term that is able to penalize the non-visible parts of Gaussians.

\subsection{Model Training}
\label{ssec:final_loss}

Our proposed losses are combined with the original loss from the 2DGS model, allowing end-to-end optimization of all parameters. This joint formulation ensures that updates are propagated to both visible and non-visible parts of every Gaussian. Accordingly, our final loss is defined as follows, $\mathcal{L} = \mathcal{L}_{2DGS} + \lambda_{bound} \mathcal{L}_{bound} + \lambda_{occ} \mathcal{L}_{occ}$, where $\lambda_{bound}$ and $\lambda_{occ}$ are the weights of each loss. The boundary loss is propagated through traditional Gaussian splatting, see Eq.~\ref{eq:boundary_loss}, updating the Gaussian parameters via the rasterization. In contrast, the occupancy loss is directly propagated to the Gaussian parameters, see Eq.~\ref{eq:occ_loss}. By optimizing $\mathcal{L}_{bound}$ and $\mathcal{L}_{occ}$ alongside $\mathcal{L}_{rgb}$ the optimization rewards boundary changes that respect rendering quality, and appearance changes that respect boundaries.

\newcommand{\Gb}[1]{\colorbox{gold}{\textbf{#1}}}
\newcommand{\Zb}[1]{\colorbox{silver}{\textbf{#1}}}
\newcommand{\Bb}[1]{\colorbox{bronze}{\textbf{#1}}}

\begin{table*}[t]
\centering
\vspace{-2mm}
\caption{Extracted metrics (left), proposed by~\cite{flashsplat} and Rendered metrics (right), proposed by~\cite{jain2024gaussiancut}. All results reported here are computed by using the same two-stage segmentations masks obtained using SAM2\cite{SAM2}. For this reason results may differ from those reported in the original papers. We report the latter in parenthesis only when they are better than those reported here. \textit{InstaScene} is marked with an asterisk, as it is evaluated using only its segmentation module, given that the in-situ diffusion-based generation component was not available. The best results are highlighted in \Gb{red}, the second best in \Zb{orange}, and the third best in \Bb{yellow}.
% Color codes: \textcolor{gold}{\textbf{Best}}, \textcolor{silver}{Second}, \textcolor{bronze}{Third}.
\vspace{-2mm}
}
\label{tab:quantitative_results}
\vspace{-0mm}
\setlength{\tabcolsep}{3pt}
\resizebox{\textwidth}{!}{%
\begin{tabular}{l |ccc| ccc| ccc |ccc| ccc| ccc}
\toprule
& \multicolumn{9}{c|}{Extracted 3D Metrics~\cite{flashsplat}} & \multicolumn{9}{c}{Rendered 3D Metrics~\cite{jain2024gaussiancut}} \\
\toprule
& \multicolumn{3}{c|}{Mip-NeRF 360} & \multicolumn{3}{c|}{LeRF} & \multicolumn{3}{c|}{LLFF}
& \multicolumn{3}{c|}{Mip-NeRF 360} & \multicolumn{3}{c|}{LeRF} & \multicolumn{3}{c}{LLFF} \\
\toprule
Method & Acc & IoU & BIoU & Acc & IoU & BIoU & Acc & IoU & BIoU & Acc & IoU & BIoU & Acc & IoU & BIoU & Acc & IoU & BIoU \\
\cmidrule(lr){1-10} \cmidrule(lr){11-19}
GaussGroup~\cite{gaussian_grouping}     & 92.0 & 58.8 & 44.7 & 93.1 & 30.2 & 19.1 & 92.2 & 73.9 & 49.6 & 98.9 & 87.8 & 79.3 & \G{99.8} & 86.7 & 80.9 & 98.0 & 89.0 & 71.8 \\
Flashsplat~\cite{flashsplat}            & \Z{98.7} & \Z{87.4} & \Z{78.6} & \Z{99.0} & 80.4 & 72.5 & \G{98.4(98.6)} & \B{91.8} & \Z{76.5} & 98.8 & 84.2 & 78.1 & 99.6 & 80.3 & 75.9 & 98.3 & 91.2 & 78.4 \\
OpenGauss~\cite{wu2024opengaussian}         & 71.6 & 30.0 & 16.9 & 98.2 & 67.1 & 57.1 & 84.5 & 65.1 & 46.5 & 83.9 & 58.5 & 51.3 & 99.4 & 84.1 & 77.8 & 85.4 & 71.9 & 60.4 \\
GaussCut~\cite{jain2024gaussiancut}         & 97.5 & 81.3 & 67.3 & 93.8 & 58.7 & 46.2 & 96.8 & 85.8 & 64.1 & \B{99.1} & \Z{88.7} & \B{82.9} & 96.3 & 82.7 & 76.7 & 98.2(98.4) & \G{91.1(92.5)} & 78.4 \\
Dr. Splat~\cite{DrSplat}      & 95.8 & 67.1 & 53.6 & 98.6 & 61.9 & 51.2 & 96.0 & 81.7 & 61.7 & 98.6 & 82.3 & 72.4 & 99.4 & 67.1 & 59.9 & 97.9 & 88.4 & 73.1 \\
UniLift~\cite{Unilift}            & 77.5 & 40.8 & 30.6 & 97.0 & 37.2 & 26.3 & 90.8 & 70.1 & 47.0 & 98.3 & 79.9 & 72.5 & 99.7 & 68.8 & 64.4 & 98.5 & \B{92.3} & \Z{81.7} \\
COB-GS~\cite{COBGS}            & 98.5 & 86.9 & \B{78.0} & 98.8 & \B{83.0} & \B{75.6} & \G{98.3(98.6)} & \Z{91.4(92.1)} & \B{75.1} & \Z{99.1} & \B{88.5} & \Z{83.8} & 99.7 & 86.5 & 81.9 & \G{98.5} & \B{92.3} & \B{81.5} \\
ILGS~\cite{ILGS}              & 65.8 & 21.6 & 12.1 & 84.0 & 17.0 & 9.6 & 81.0 & 50.5 & 29.3 & 98.7 & 83.2 & 68.5 & 99.6 & 73.8 & 66.0 & 98.4 & 91.7 & 74.0 \\
Trace3D~\cite{Trace3D}           & \B{98.6} & \B{87.1} & 75.4 & 98.3 & \Z{85.9} & \Z{80.1} & 97.3 & 86.1 & 60.1 & 98.7 & 82.9 & 75.6 & 99.7 & 88.3 & 83.1 & 98.2 & 90.4 & 77.5 \\
InstaScene*~\cite{InstaScene}        & 98.3 & 86.3 & 77.3 & \B{98.9} & 76.1 & 67.1 & 96.8 & 88.1 & 73.9 & 98.9 & 85.1 & 77.5 & 99.7 & \Z{89.6} & \B{84.0} & \G{98.5} & 92.2 & 67.2 \\
ObjectGS~\cite{ObjectGS}           & 96.4 & 76.7 & 61.3 & 98.1 & 69.1 & 58.5 & 93.7 & 71.9 & 49.8 & 98.9 & 86.8 & 80.9 & \G{99.8} & \Z{89.6} & \Z{86.7} & 98.2 & 90.2 & 73.4 \\
SAGD~\cite{SAGD}                        & 96.9 & 65.6 & 56.5 & 97.6 & 73.2 & 66.0 & 95.9 & 79.2 & 55.5 & 97.3 & 62.1 & 56.2 & 98.9 & 72.5 & 67.3 & 97.9 & 87.6 & 74.7 \\
\cmidrule(lr){1-10} \cmidrule(lr){11-19}
\methodnametable & \G{99.1} & \G{92.0} & \G{85.8} & \G{99.2} & \G{89.4} & \G{83.6} & \G{98.6} & \G{93.0} & \G{80.7} & \G{99.2} & \G{89.8} & \G{85.3} & \G{99.8} & \G{92.0} & \G{88.5} & \G{98.5} & \Z{92.4} & \G{82.9} \\
%Diff to 2nd & +0.4 & +4.5 & +7.1 & +0.2 & +3.5 & +3.5 & 0.0 & +0.9 & +4.3 & 0.0 & +1.1 & +1.5 & 0.0 & +2.3 & +4.2 & +0.0 & -0.1 & +1.0 \\
\bottomrule
\end{tabular}
\vspace{-0mm}
}
\end{table*}

\begin{table}[t]
\centering
\caption{Quantitative results on the 3DOVS dataset~\cite{3DOVS_dataset_paper}. Same color convention as in Tab. \ref{tab:quantitative_results}}
\label{tab:quantitative_results_3dovs}
\vspace{-2mm}
\setlength{\tabcolsep}{8pt}
\resizebox{\columnwidth}{!}{
\begin{tabular}{l |ccc| ccc}
\toprule
& \multicolumn{3}{c|}{Extracted 3D Metrics~\cite{flashsplat}} & \multicolumn{3}{c}{Rendered 3D Metrics~\cite{jain2024gaussiancut}} \\
\midrule
Method & Acc & IoU & BIoU & Acc & IoU & BIoU \\
\midrule
Flashsplat~\cite{flashsplat}    & \B{99.3} & \B{86.8} & \B{75.7} & 99.5 & 88.1 & 78.2 \\
COB-GS~\cite{COBGS}             & 99.2 & 85.8 & 73.9 & \G{99.8} & \Z{94.1} & \Z{87.9} \\
Trace3D~\cite{Trace3D}          & \Z{99.4} & \Z{90.2} & \Z{81.1} & 99.7 & 92.1 & 84.9 \\
ObjectGS~\cite{ObjectGS}        & 98.6 & 79.2 & 68.2 & \G{99.8} & \G{95.3} & \G{90.7} \\
\midrule
\methodnametable                & \G{99.7} & \G{93.2} & \G{87.3} & \G{99.8} & \B{93.7} & \B{87.8} \\
\bottomrule
\end{tabular}
\vspace{-0mm}
}
\end{table}

\begin{table}[t]
\centering
\caption{Ablation study. \textbf{R}: Multiview Reprojection, \textbf{B}: Boundary loss, \textbf{O}: Occupancy loss.}
\label{tab:ablation_study}
\vspace{-2mm}
\setlength{\tabcolsep}{1pt}
\resizebox{\columnwidth}{!}{
\begin{tabular}{ccc@{\hskip 10pt}ccc@{\hskip 10pt}ccc@{\hskip 10pt}ccc@{\hskip 10pt}ccc}
\toprule
\multicolumn{3}{c}{Module} & \multicolumn{3}{c}{Mip-NeRF 360} & \multicolumn{3}{c}{LeRF} & \multicolumn{3}{c}{LLFF} & \multicolumn{3}{c}{3DOVS} \\
%\cmidrule(lr){1-3}\cmidrule(lr){4-7}\cmidrule(lr){8-11}\cmidrule(lr){12-15}
\midrule
\textbf{R} & \textbf{B} & \textbf{O} & Acc & IoU & BIoU & Acc & IoU & BIoU & Acc & IoU & BIoU & Acc & IoU & BIoU \\
\midrule
\xmark & \xmark & \xmark   & 98.1 & 83.9 & 72.8 & 98.3 & 66.3 & 56.6 & 97.2 & 86.8 & 66.4 & 98.3 & 76.8 & 65.6\\
\cmark & \cmark & \xmark   & 99.0 & 90.5 & 83.6 & 98.7 & 78.2 & 70.1 & 98.2 & 90.8 & 75.0 & 98.9 & 82.6 & 73.9\\
\cmark & \xmark & \cmark   & 98.9 & 90.1 & 81.9 & 99.2 & 85.0 & 77.2 & 98.5 & 92.8 & 80.0 & 99.7 & 93.2 & 87.0\\
\xmark & \cmark & \cmark   & 99.0 & 90.9 & 84.5 & 99.2 & 86.6 & 80.3 & 98.6 & 93.0 & 80.4 & 99.7 & 93.3 & 87.0 \\
\cmark & \cmark & \cmark   & 99.1 & 92.0 & 85.8 & 99.2 & 89.4 & 83.6 & 98.6 & 93.0 & 80.7 & 99.7 & 93.2 & 87.3  \\
\bottomrule
\end{tabular}
}
\vspace{-1mm}
\end{table}

\section{Results}
\label{sec:results}

To validate our approach we performed an exhaustive quantitative assessment based on 6 metrics over 4 benchmark datasets with comparisons to 12 state of the art baselines. We also report qualitative results to compare the quality of object boundaries obtained through~\methodname~with those of the best performing methods over several scenes. Finally, we provide an ablation study that provides in-depth analysis of our contributions.

\subsection{Datasets, Metrics and Baselines}
\label{sec:datasets_metrics}

We evaluate on four datasets, LeRF~\cite{lerf2023} with segmentation masks from Gaussian Grouping~\cite{gaussian_grouping}, LLFF~\cite{mildenhall2019llff} with masks from NVOS~\cite{nvos_ren_cvpr2022}, Mip-NeRF 360~\cite{barron2022mipnerf360} with our generated ground-truth masks, and 3DOVS~\cite{3DOVS_dataset_paper} with its original ground-truth masks. All methods use identical training masks for a fair comparison. We use the 3D segmentation mean IoU metric and the mean accuracy as proposed by Flashsplat~\cite{flashsplat}, and used by~\cite{COBGS, Trace3D, wu2024opengaussian, DrSplat, SAGD}, the hyper-parameters are the same as defined in~\cite{flashsplat}. Flashsplat's metrics~\cite{flashsplat} first assign classes in 3D space and then render the object on its own to clearly assess the quality of the semantic boundaries. Then, they compute the mean intersection over the union (IoU) and the mean accuracy (Acc) values. We also compute the mean value of the BIoU metric~\cite{Cheng_2021_CVPR} (BIoU) which gives a better assessment of the quality of the produced boundaries. We call these metrics \textit{Extracted 3D Metrics}. A similar metric was used by GaussianCut~\cite{jain2024gaussiancut}, and also used by~\cite{ObjectGS}, were classes are assigned to Gaussians in 3D space, but instead of rendering each object on its own they render the whole scene. They then compute the mean (Acc, IoU) scores; in our case, we additionally report the mean (BIoU) scores. We call these metrics \textit{Rendered 3D Metrics}. \textit{Extracted 3D Metrics} measure the quality of segmentation when extracting individual objects, while \textit{Rendered 3D Metrics} assess the quality of the segmentation when rendering the object within the scene. In all our experiments we respect the original train/test splits for all datasets to assure no test image has been used during training for either reconstruction or semantic segmentation.
If the original paper has better results for a given metric than the one we obtained we show both results in the table to clearly disclose the discrepancy.

\subsection{Quantitative Results}
\label{sec:quantitative_results}

Quantitative results are presented in Tab.~\ref{tab:quantitative_results} for LLFF, LeRF and the Mip-NeRF 360 datasets, and in Tab.~\ref{tab:quantitative_results_3dovs} for the 3DOVS dataset. Our proposed approach outperforms the state of the art in 21 out of 24 evaluated cases. The improvement in the BIoU metric is quite noticeable as it shows that our approach obtains the cleanest boundaries when the object is extracted. Another significant achievement is that some methods perform better under certain datasets and/or metrics but our approach is consistently obtaining the best results regardless of the dataset employed and the specificity of the metric. We also address the possible loss of rendering quality, a major concern when modifying the geometry of a Gaussian Splatting scene. We report the (before/after) mean PSNR values for each dataset for the test set: Mip-NeRF 360 ($29.1/29.1$), LeRF ($25.2/25.0$), LLFF ($25.0/24.9$), 3DOVS ($26.7/26.5$). It can be seen that our approach has near zero loss in rendering quality.

\subsection{Qualitative Results}
\label{sec:qualitative_results}

\begin{figure*}[t]
    \centering
    \setlength{\tabcolsep}{0.5pt}
    \renewcommand{\arraystretch}{0.95}
    \begin{tabular}{ccccc}

    %\vspace{-2pt}
    \raisebox{11mm}{\rotatebox{90}{\textbf{GT}}} &
    \includegraphics[width=0.237\linewidth]{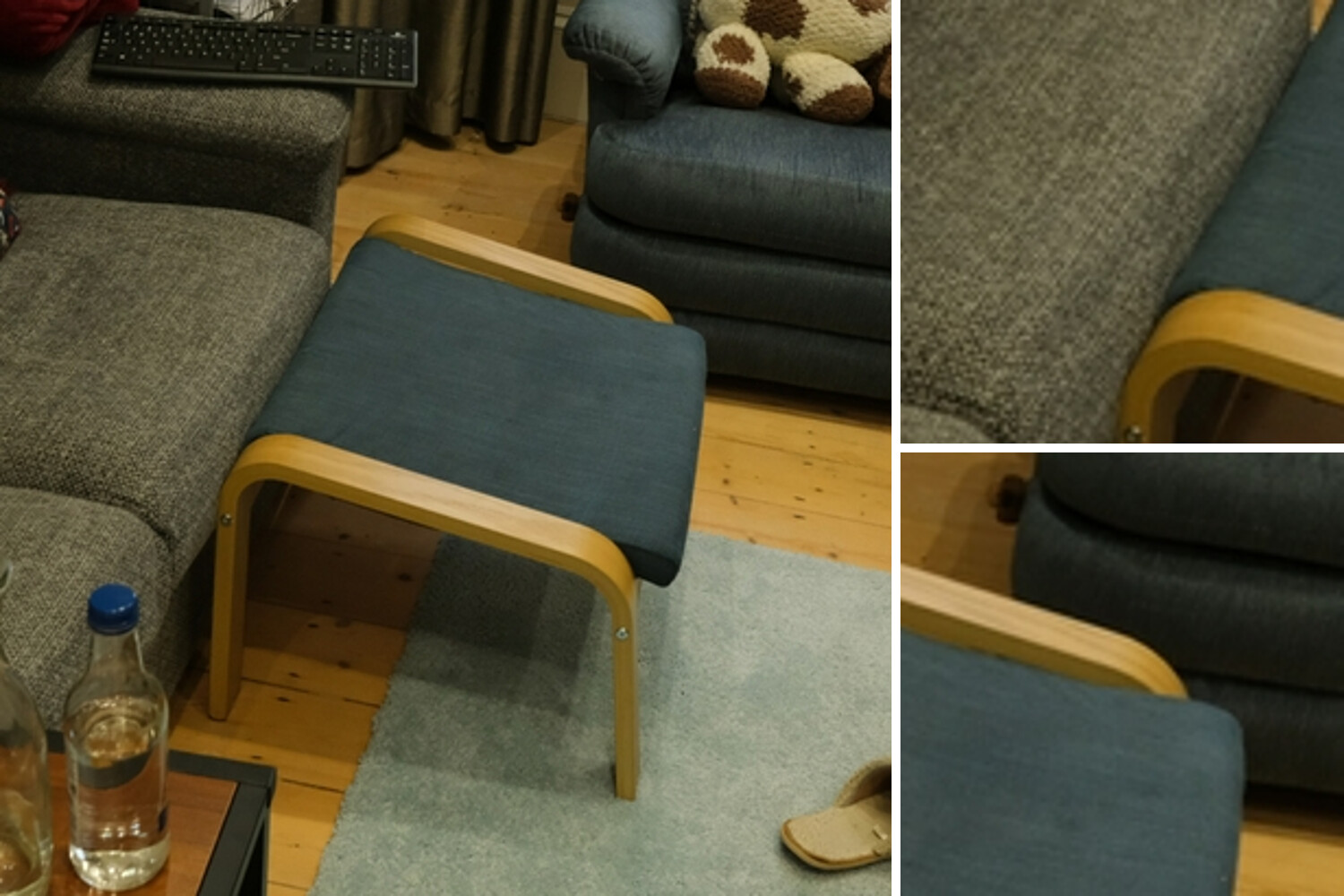} &
    \includegraphics[width=0.237\linewidth]{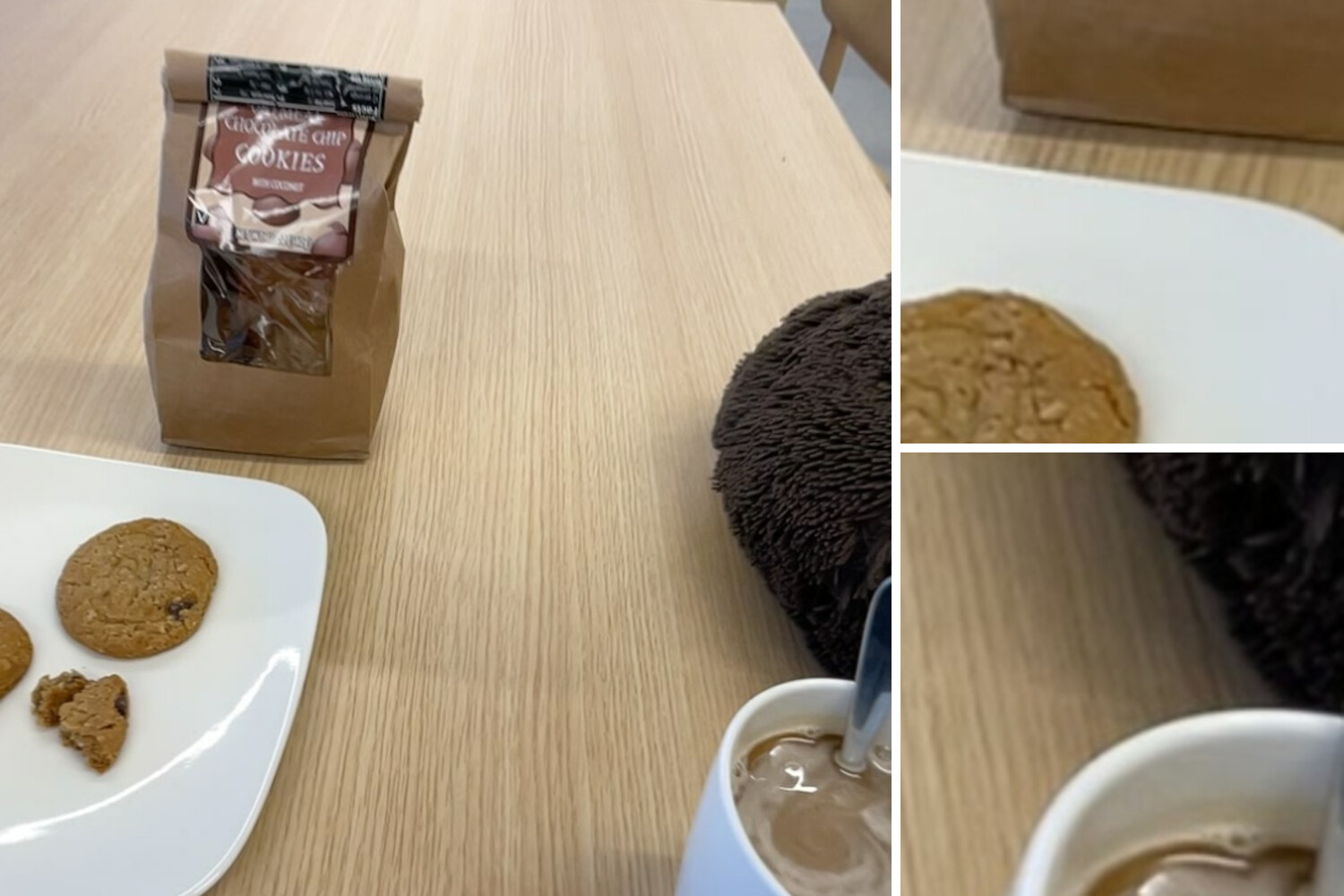} &
    \includegraphics[width=0.237\linewidth]{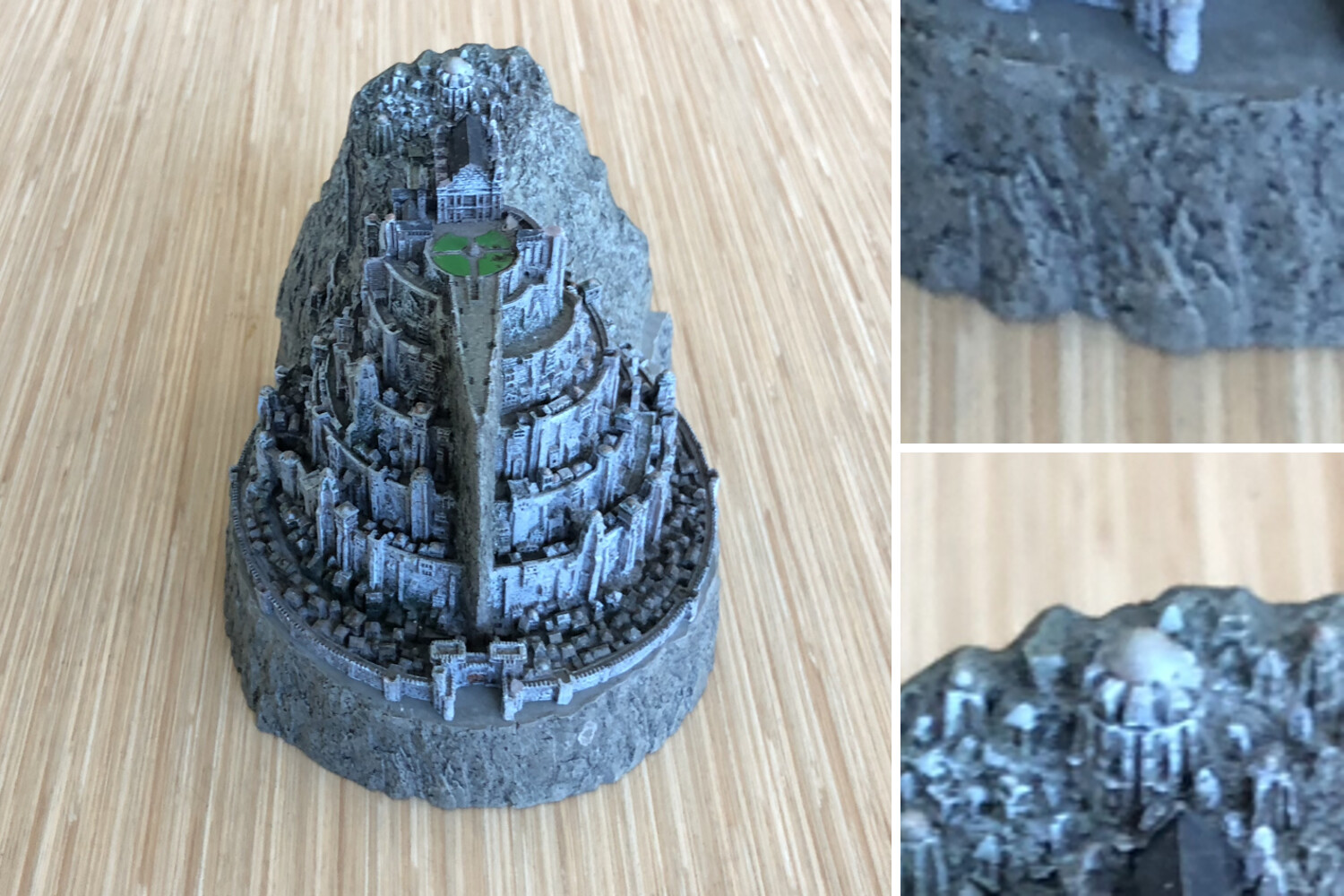} &
    \includegraphics[width=0.237\linewidth]{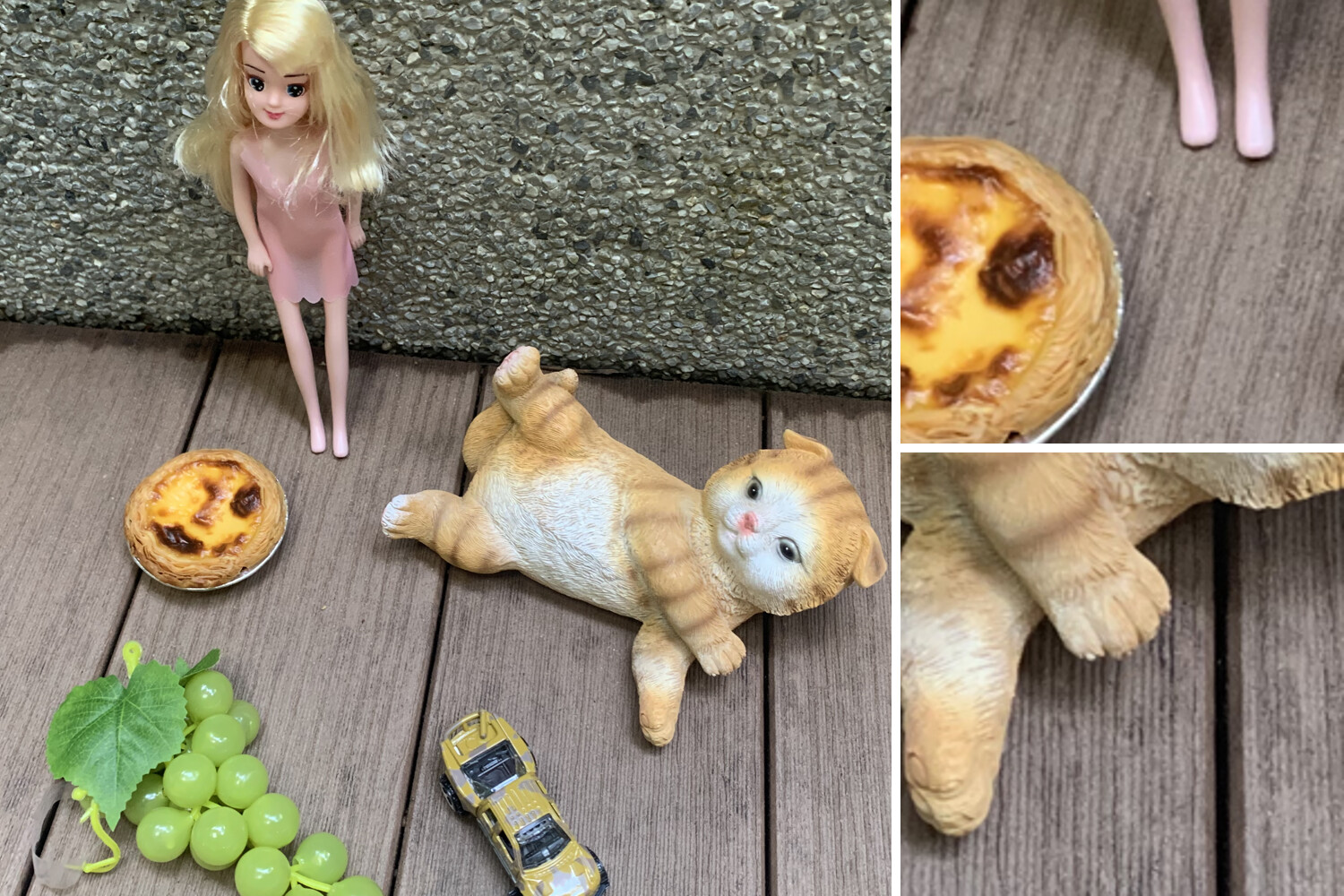} \\
    \vspace{-2pt}

    \raisebox{6mm}{\rotatebox{90}{\textbf{FlashSplat}}} &
    \includegraphics[width=0.237\linewidth]{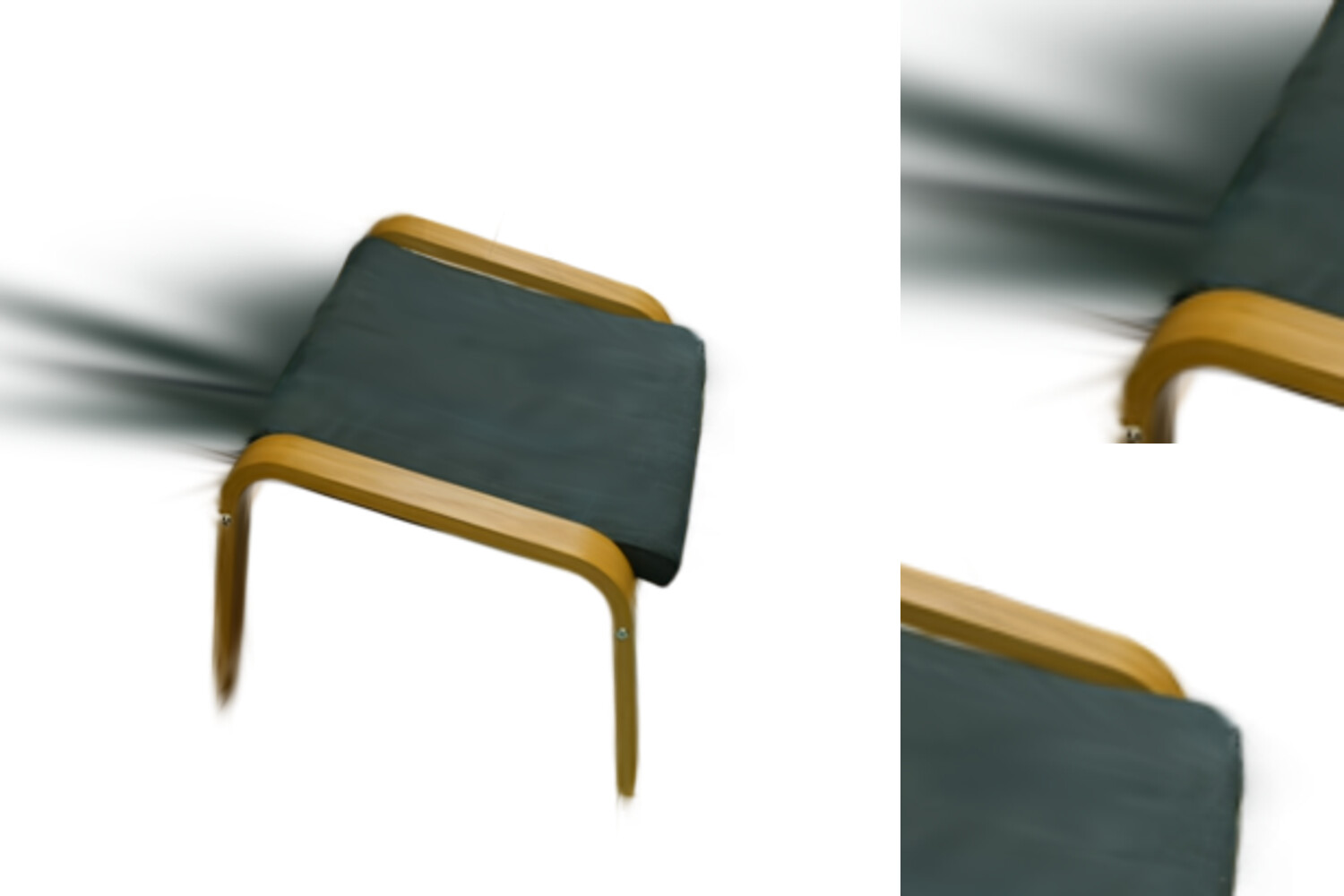} &
    \includegraphics[width=0.237\linewidth]{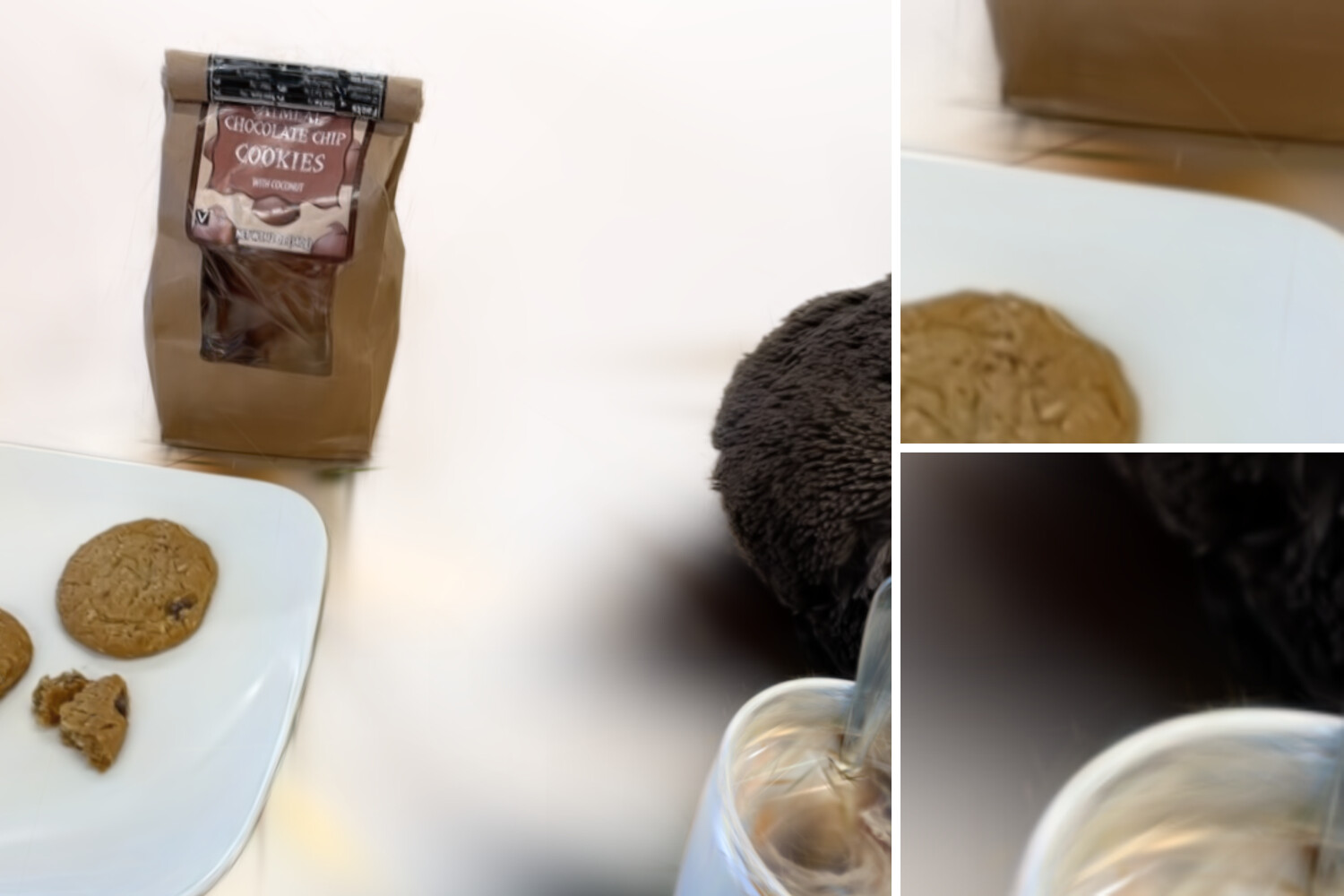} &
    \includegraphics[width=0.237\linewidth]{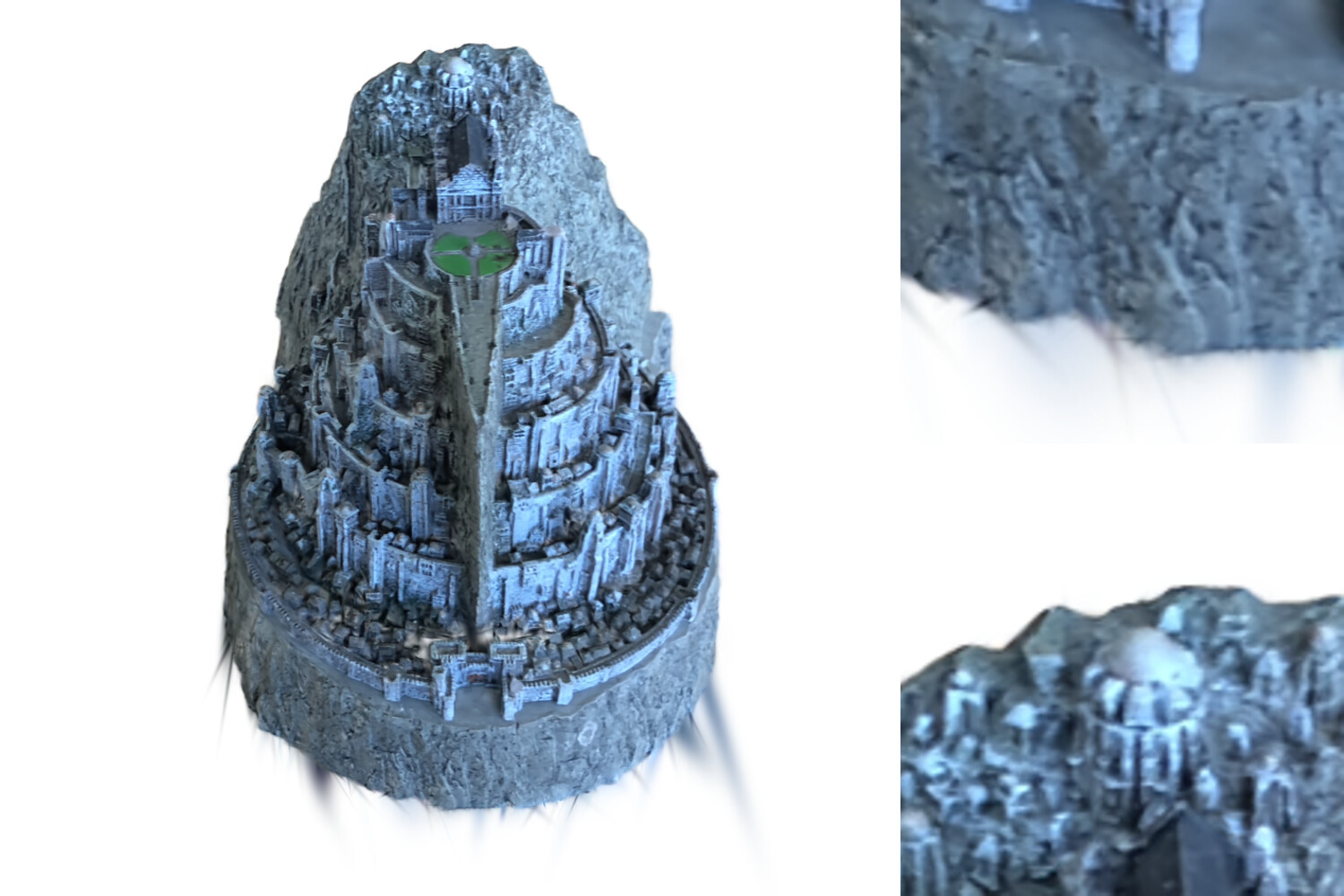} &
    \includegraphics[width=0.237\linewidth]{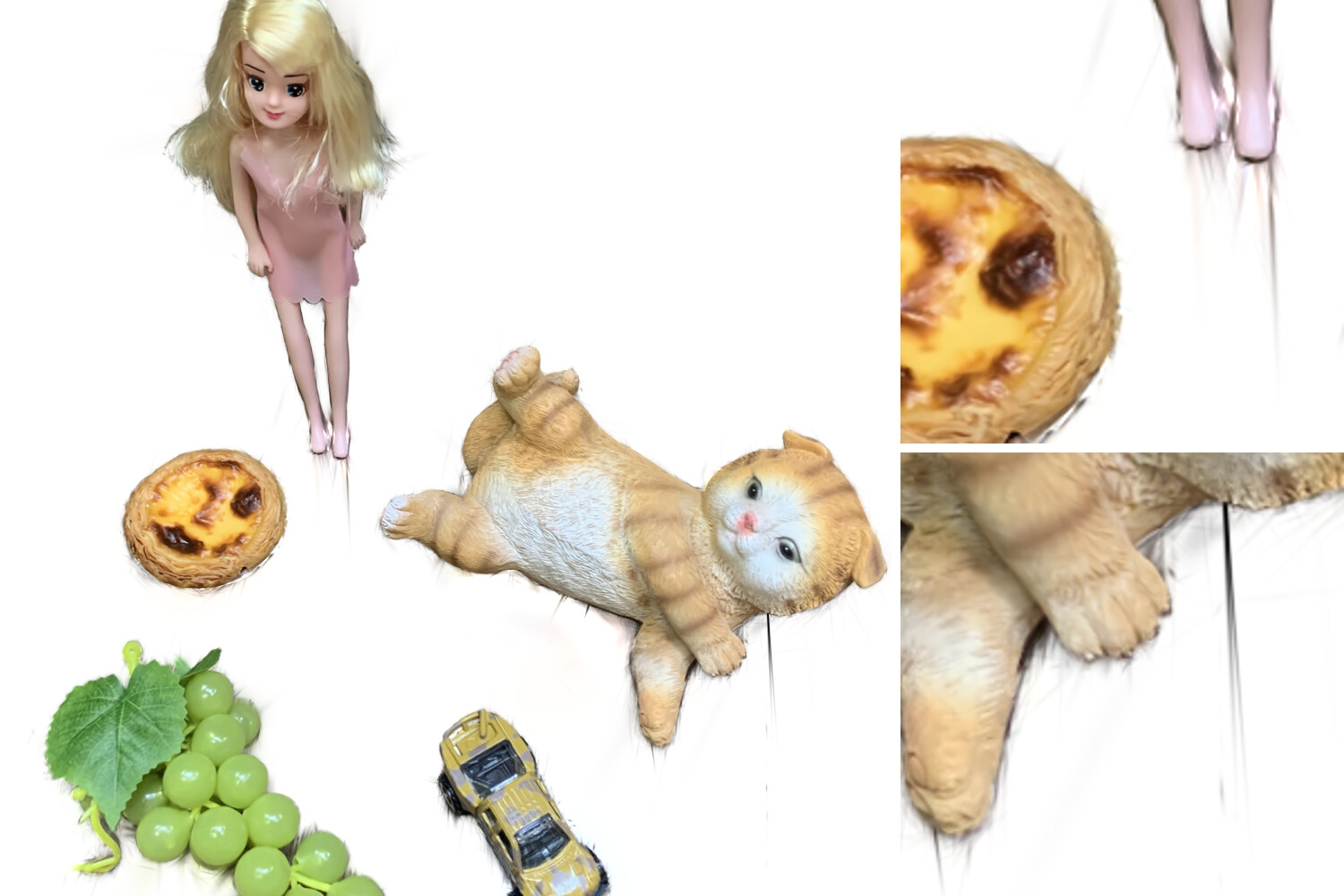} \\
    \vspace{-2pt}
    
    \raisebox{5mm}{\rotatebox{90}{\textbf{ObjectGS}}} &
    \includegraphics[width=0.237\linewidth]{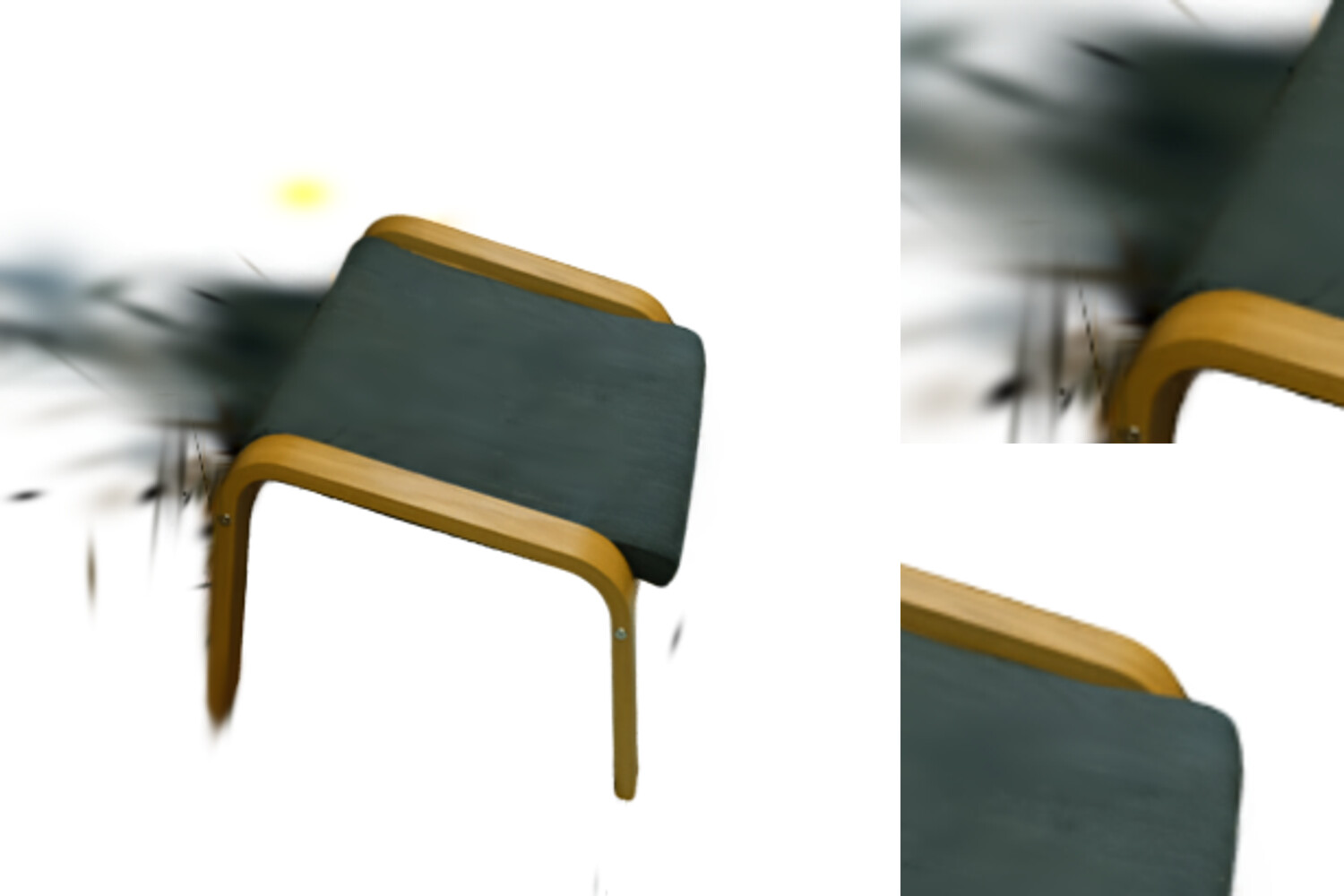} &
    \includegraphics[width=0.237\linewidth]{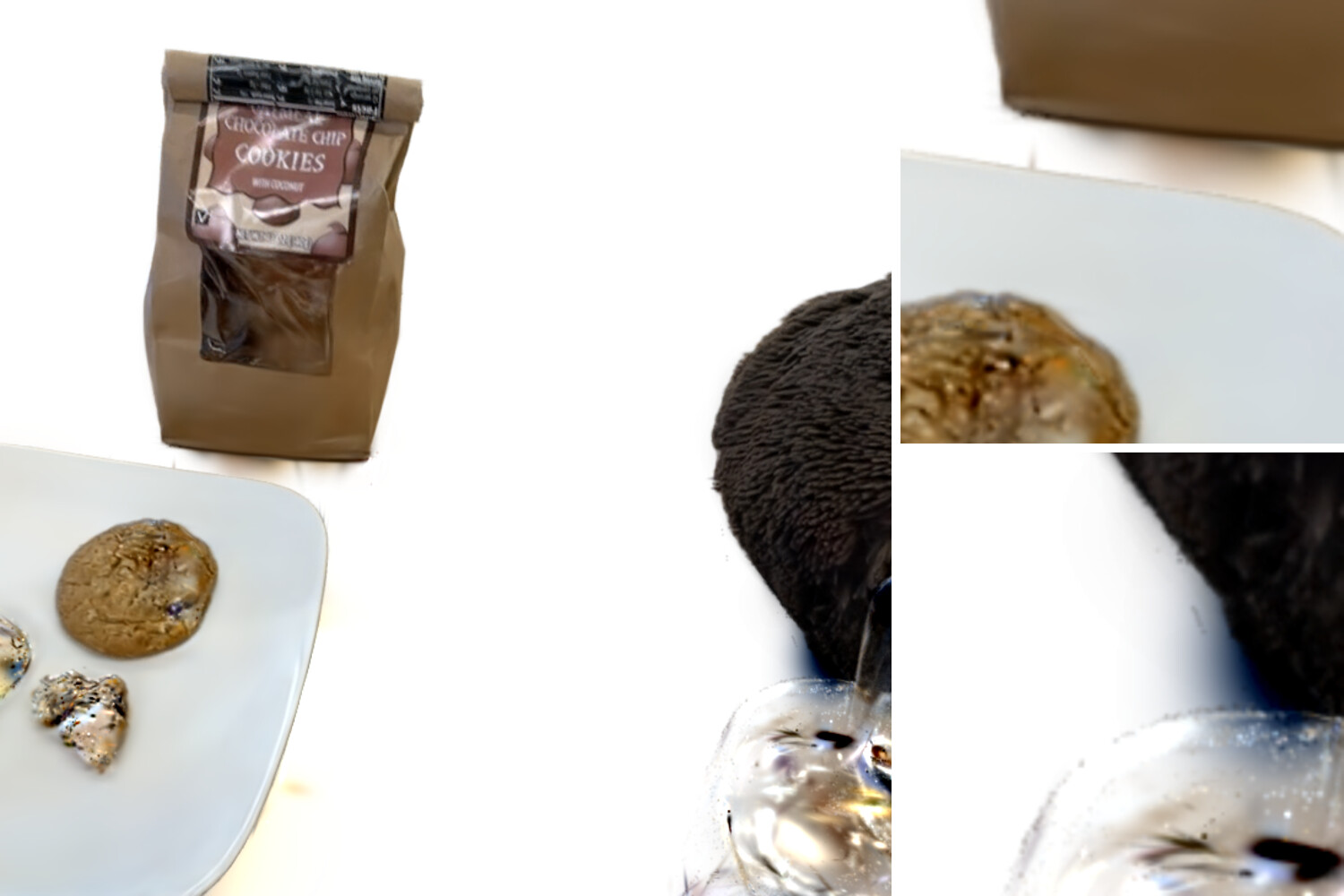} &
    \includegraphics[width=0.237\linewidth]{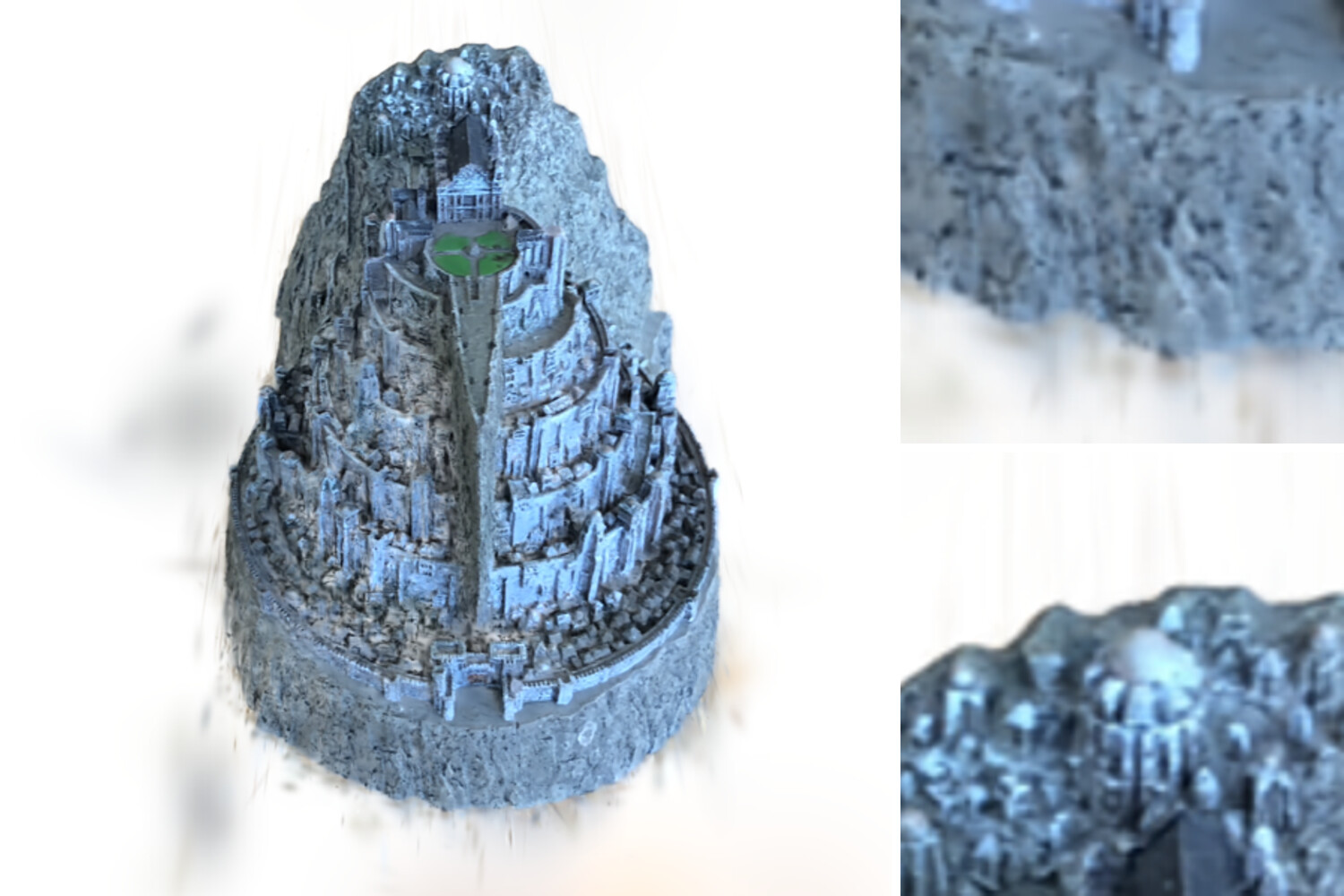} &
    \includegraphics[width=0.237\linewidth]{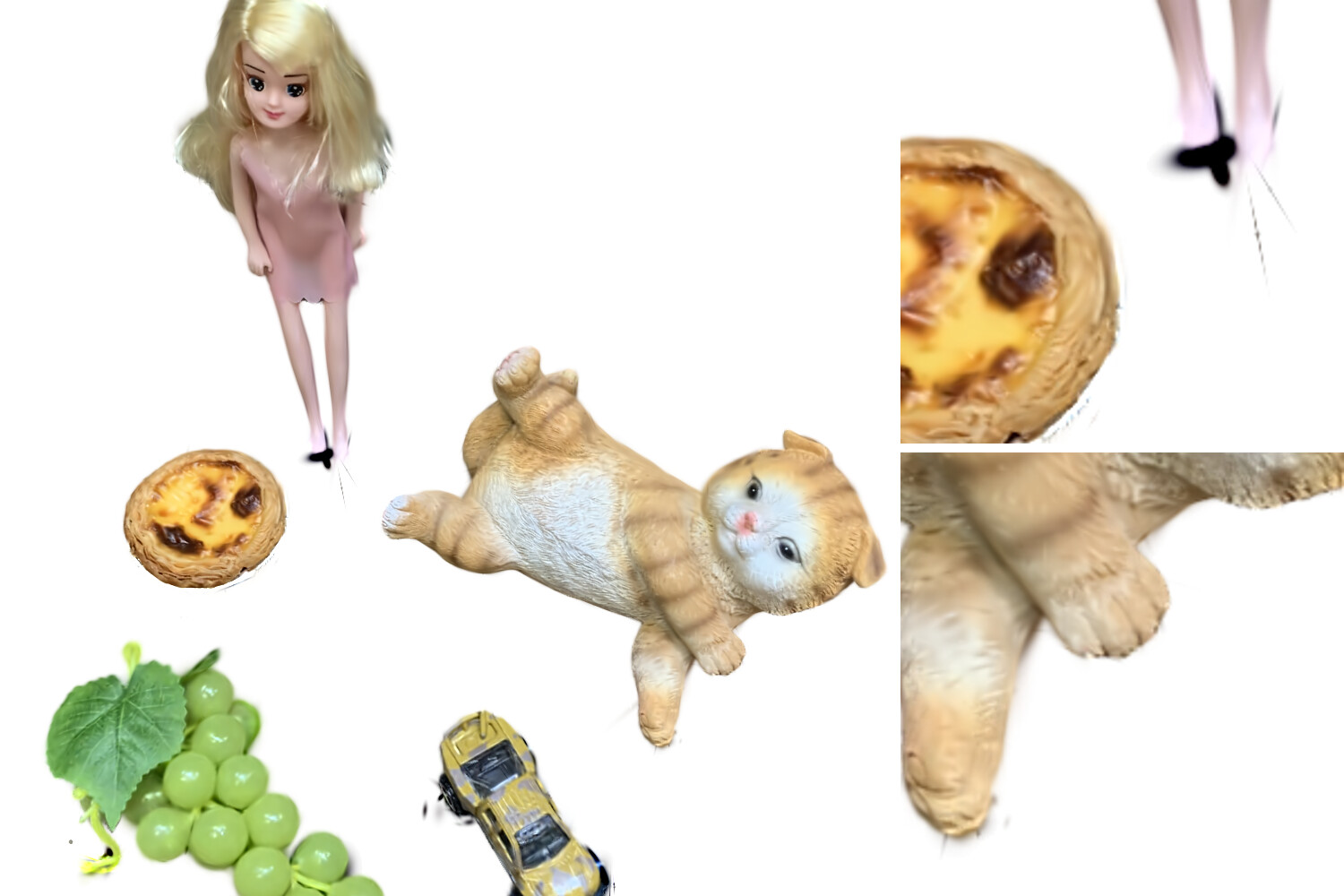} \\
    \vspace{-2pt}
    
    \raisebox{6mm}{\rotatebox{90}{\textbf{COB-GS}}} &
    \includegraphics[width=0.237\linewidth]{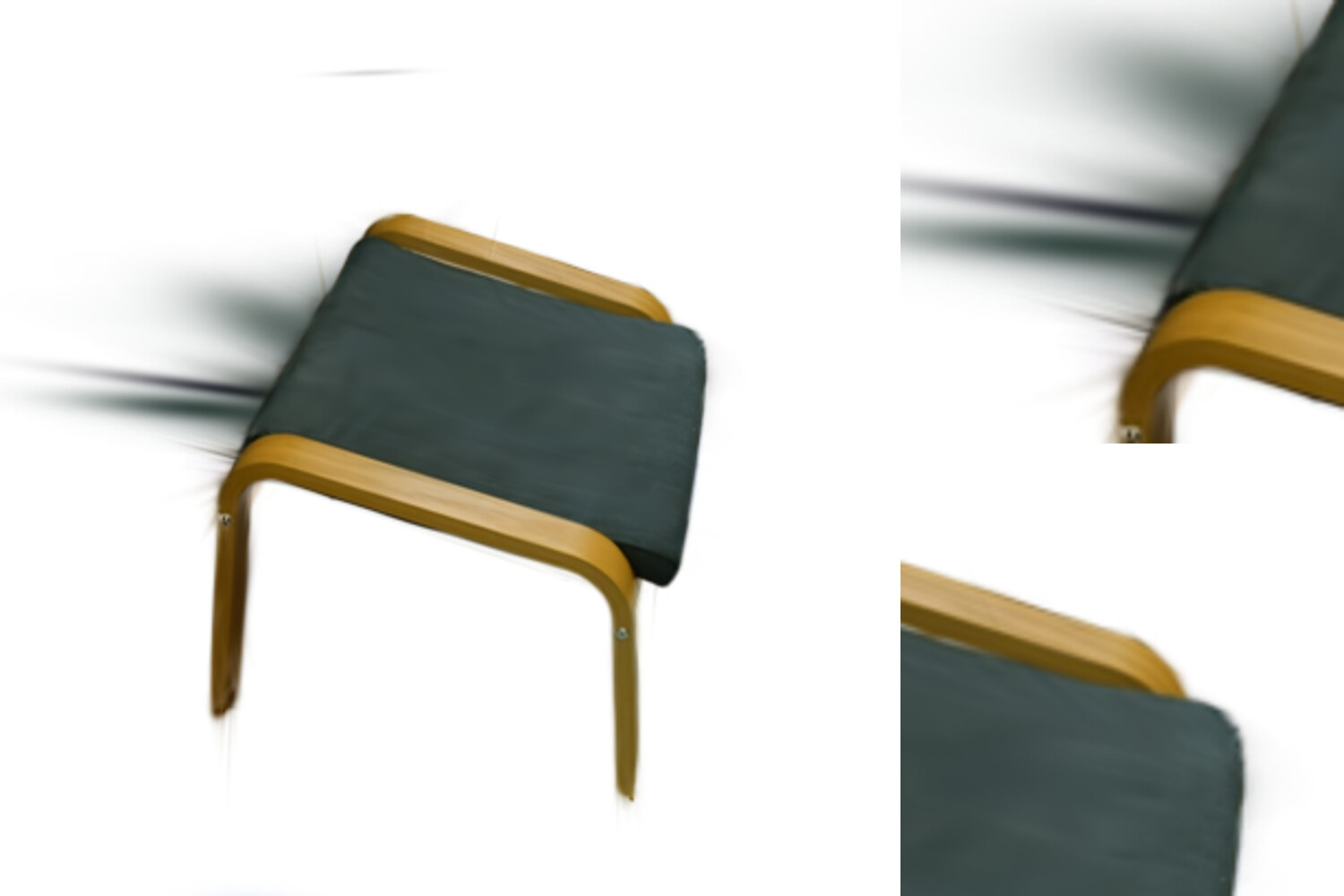} &
    \includegraphics[width=0.237\linewidth]{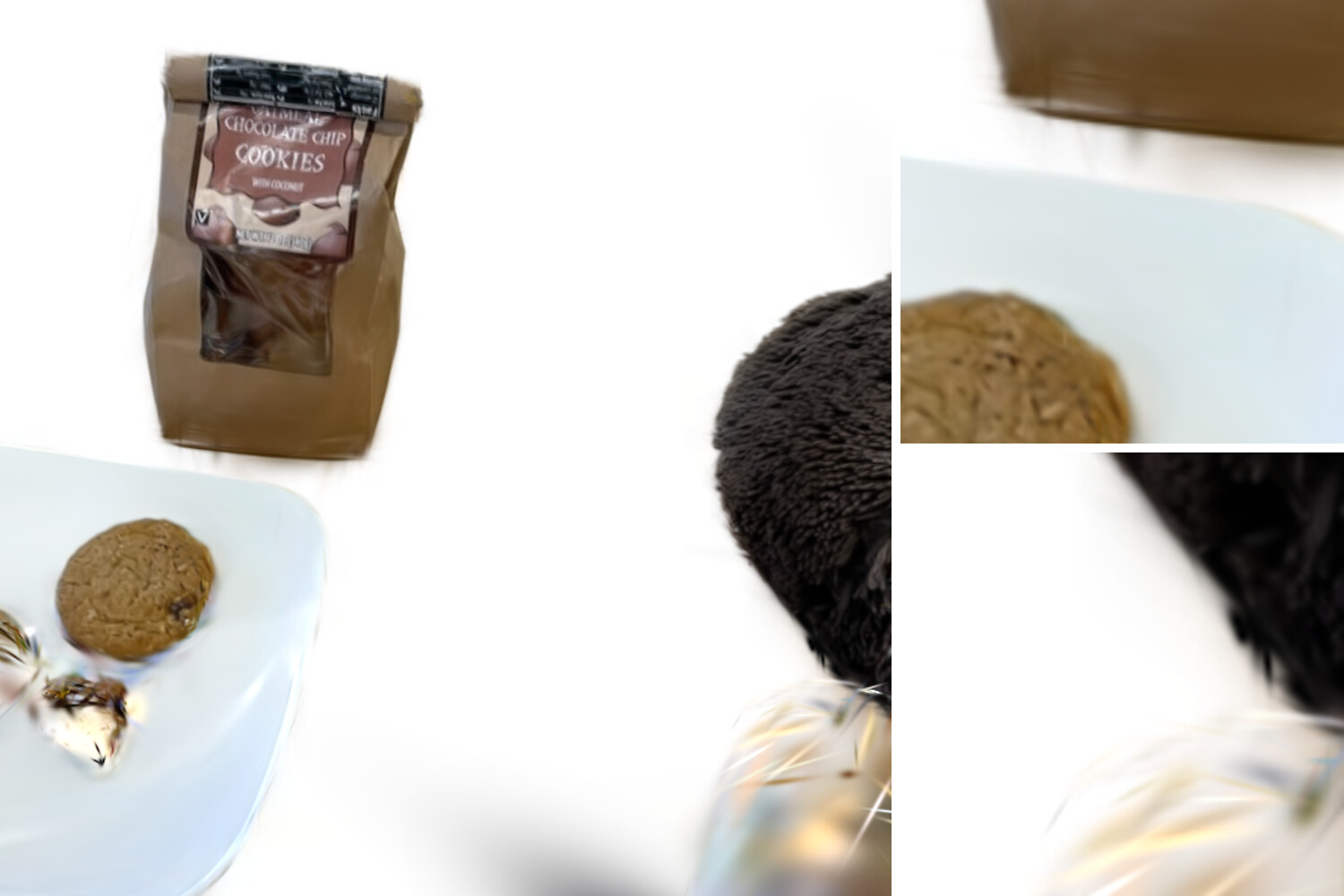} &
    \includegraphics[width=0.237\linewidth]{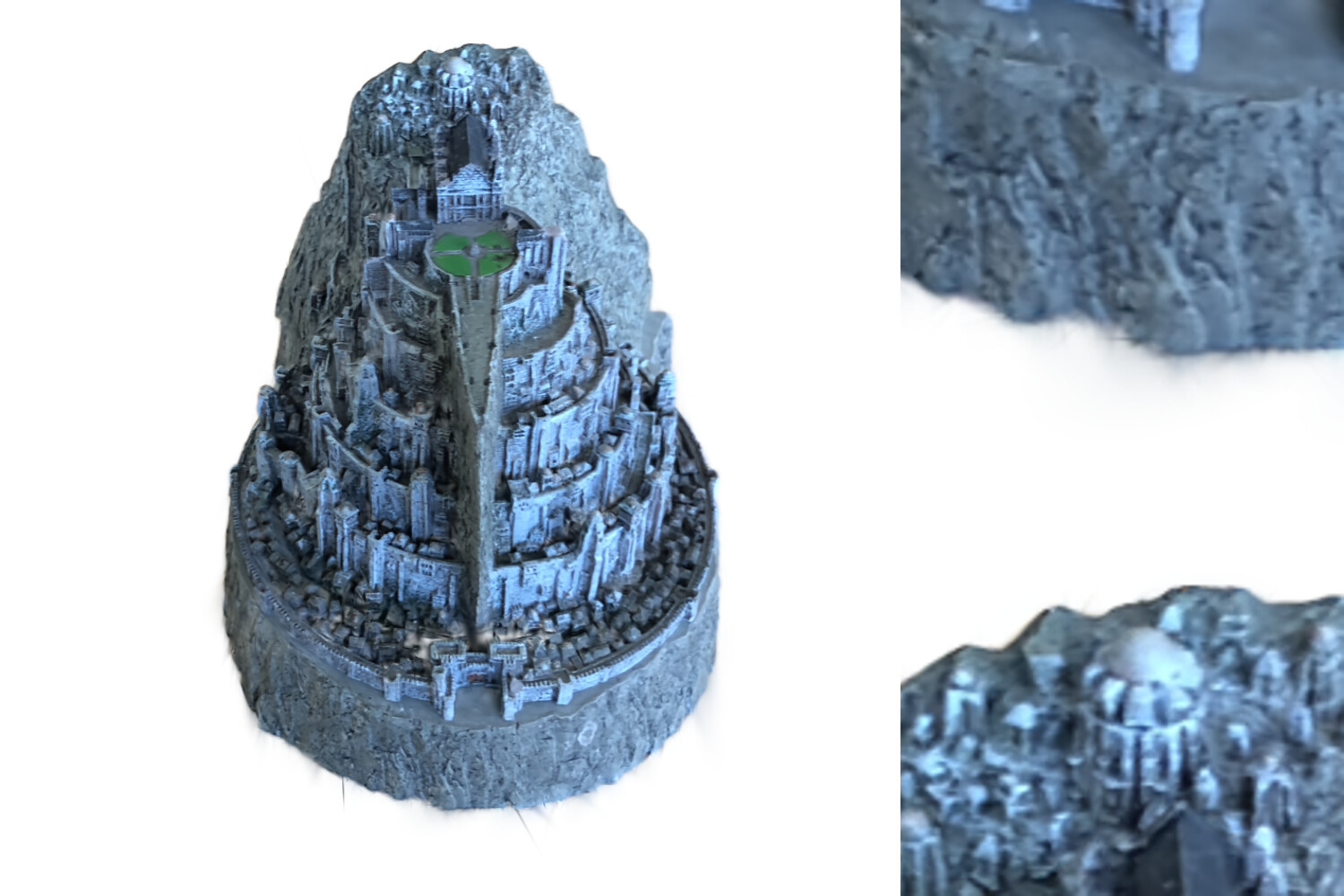} &
    \includegraphics[width=0.237\linewidth]{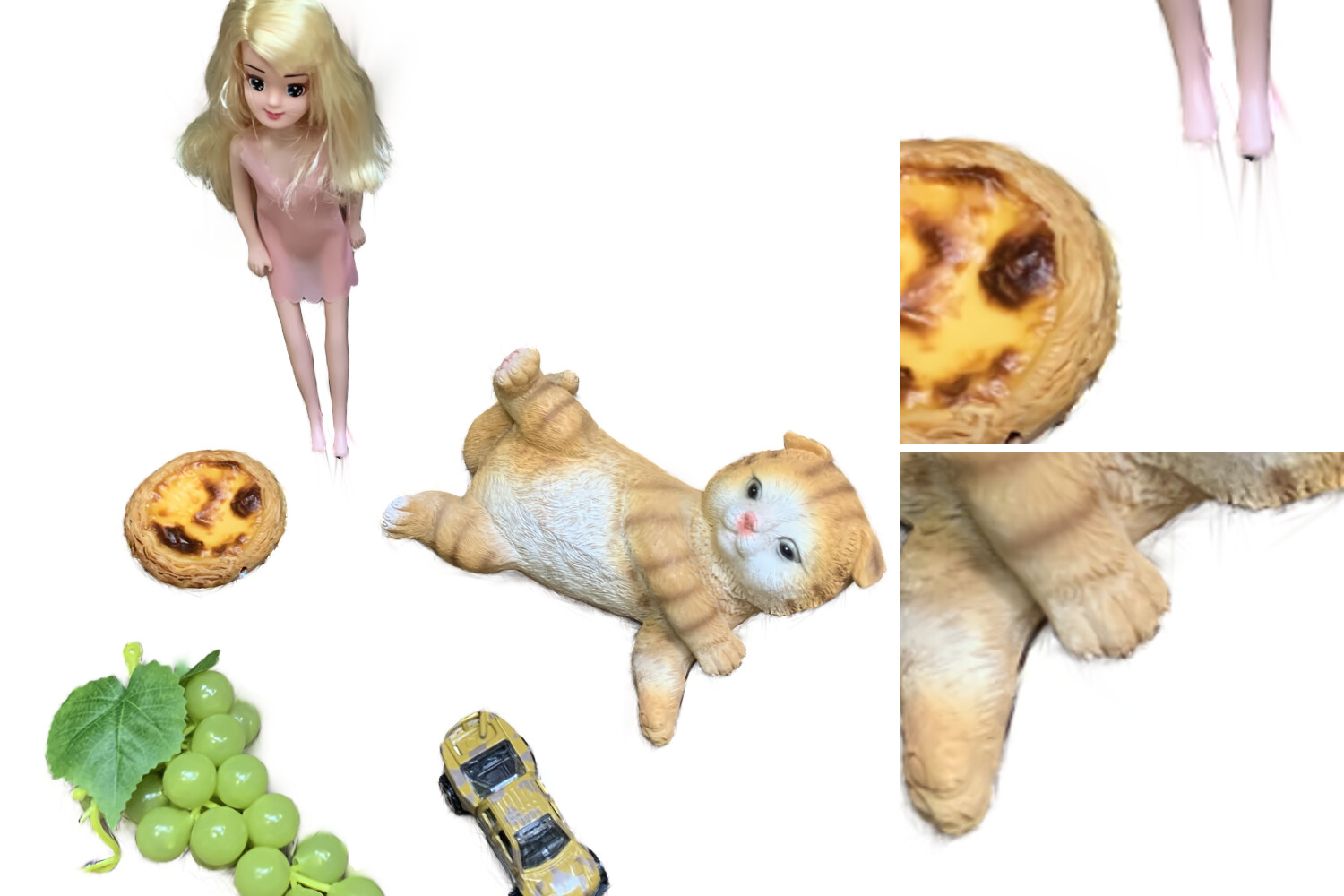} \\
    \vspace{-2pt}
    
    \raisebox{6mm}{\rotatebox{90}{\textbf{Trace3D}}} &
    \includegraphics[width=0.237\linewidth]{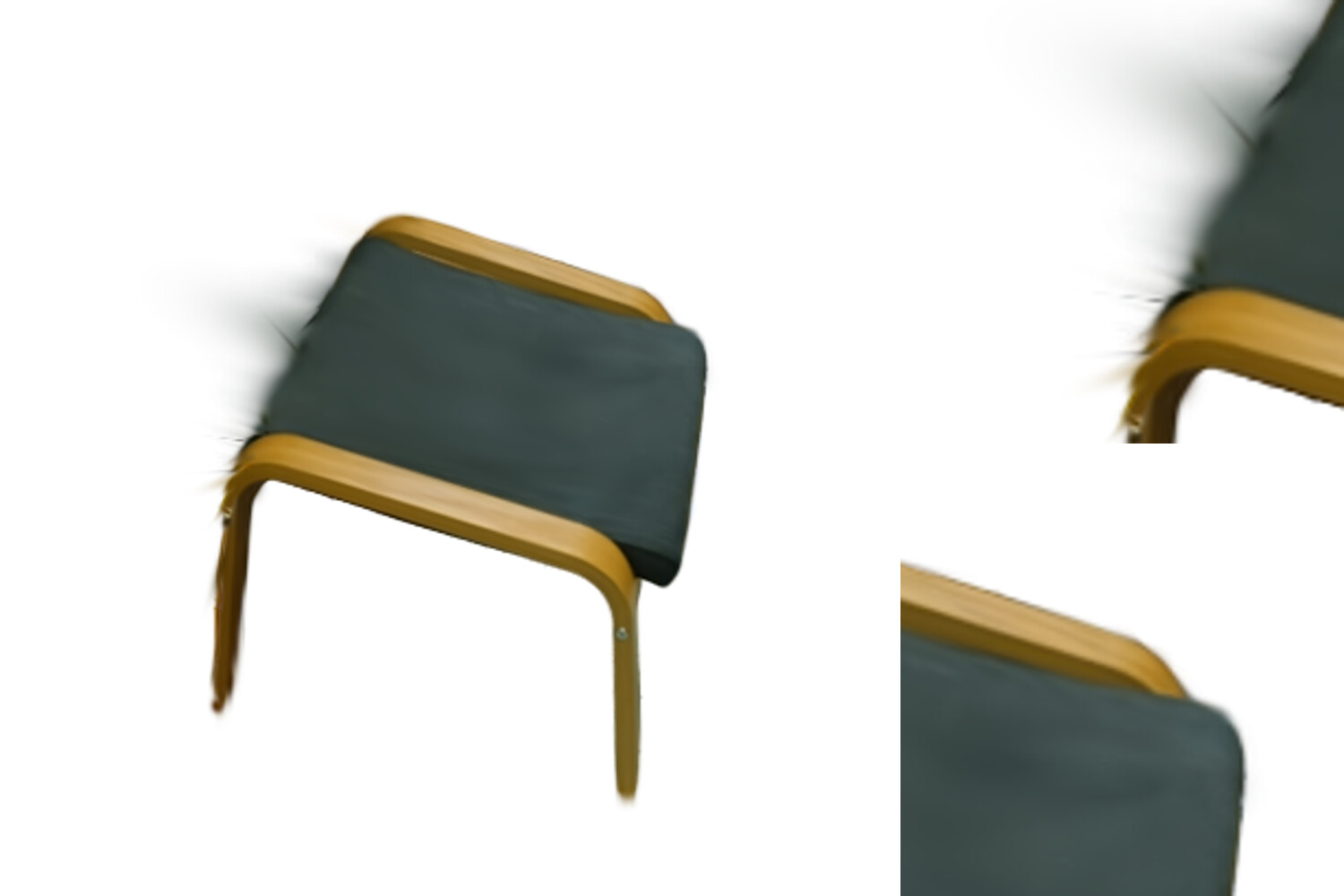} &
    \includegraphics[width=0.237\linewidth]{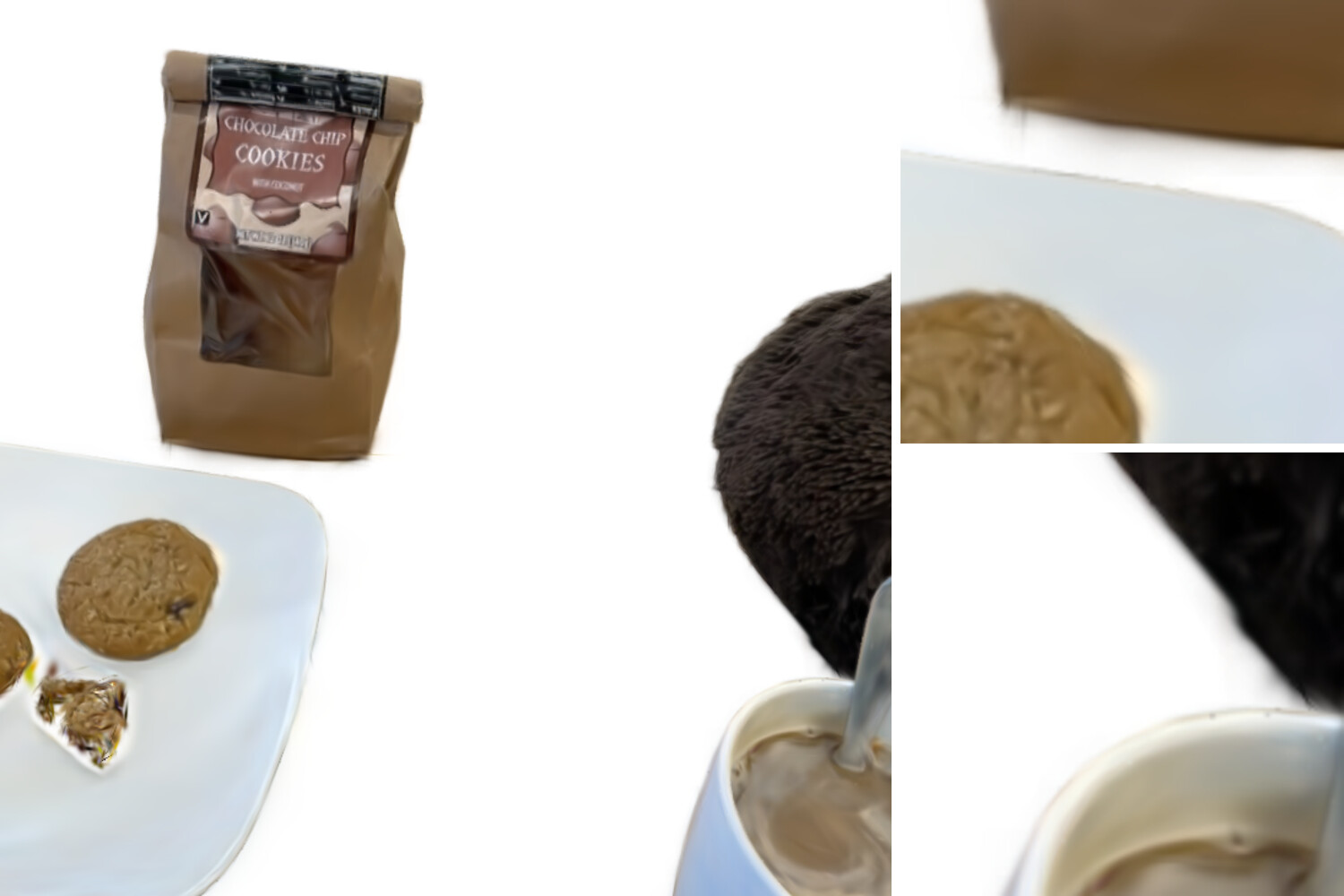} &
    \includegraphics[width=0.237\linewidth]{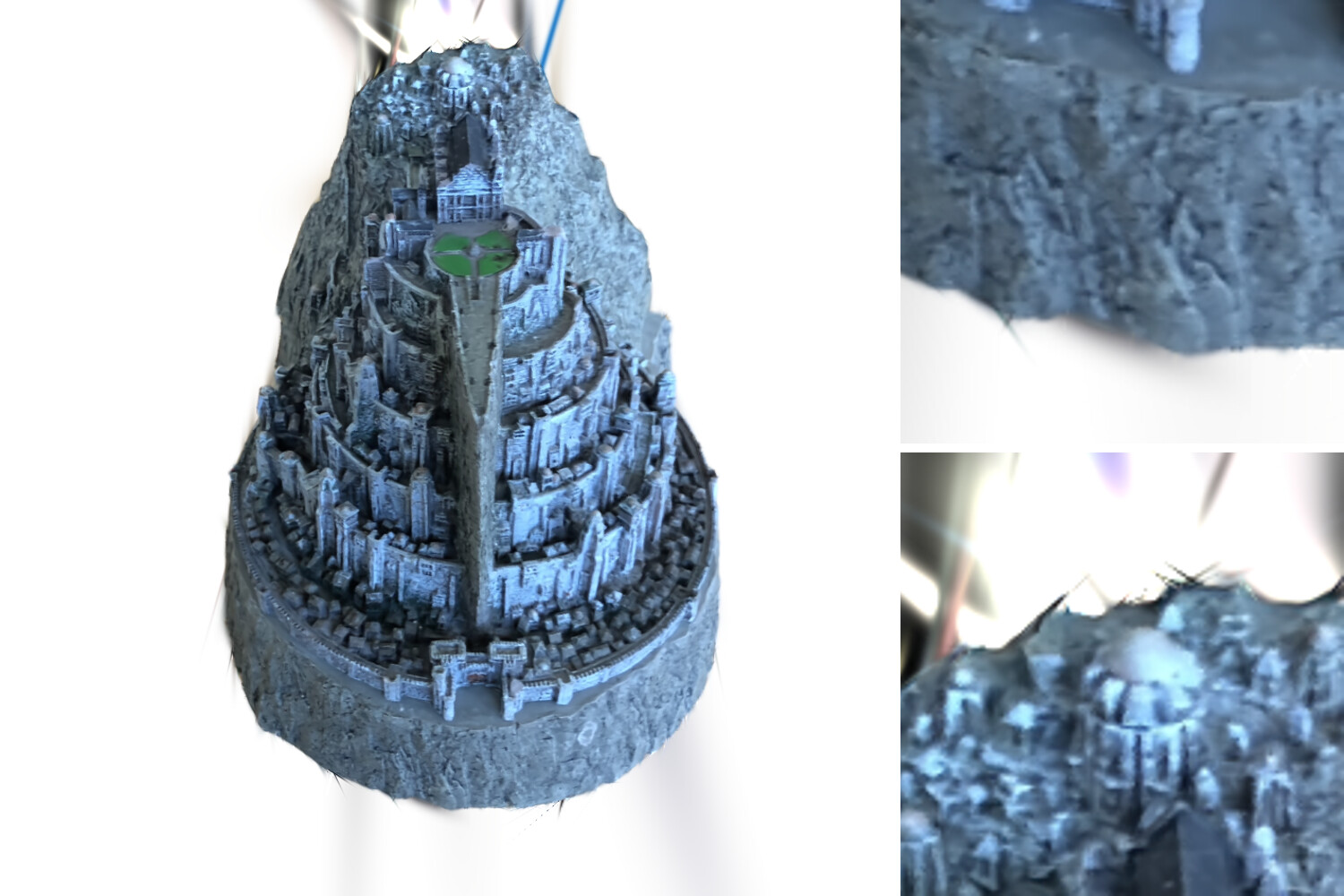} &
    \includegraphics[width=0.237\linewidth]{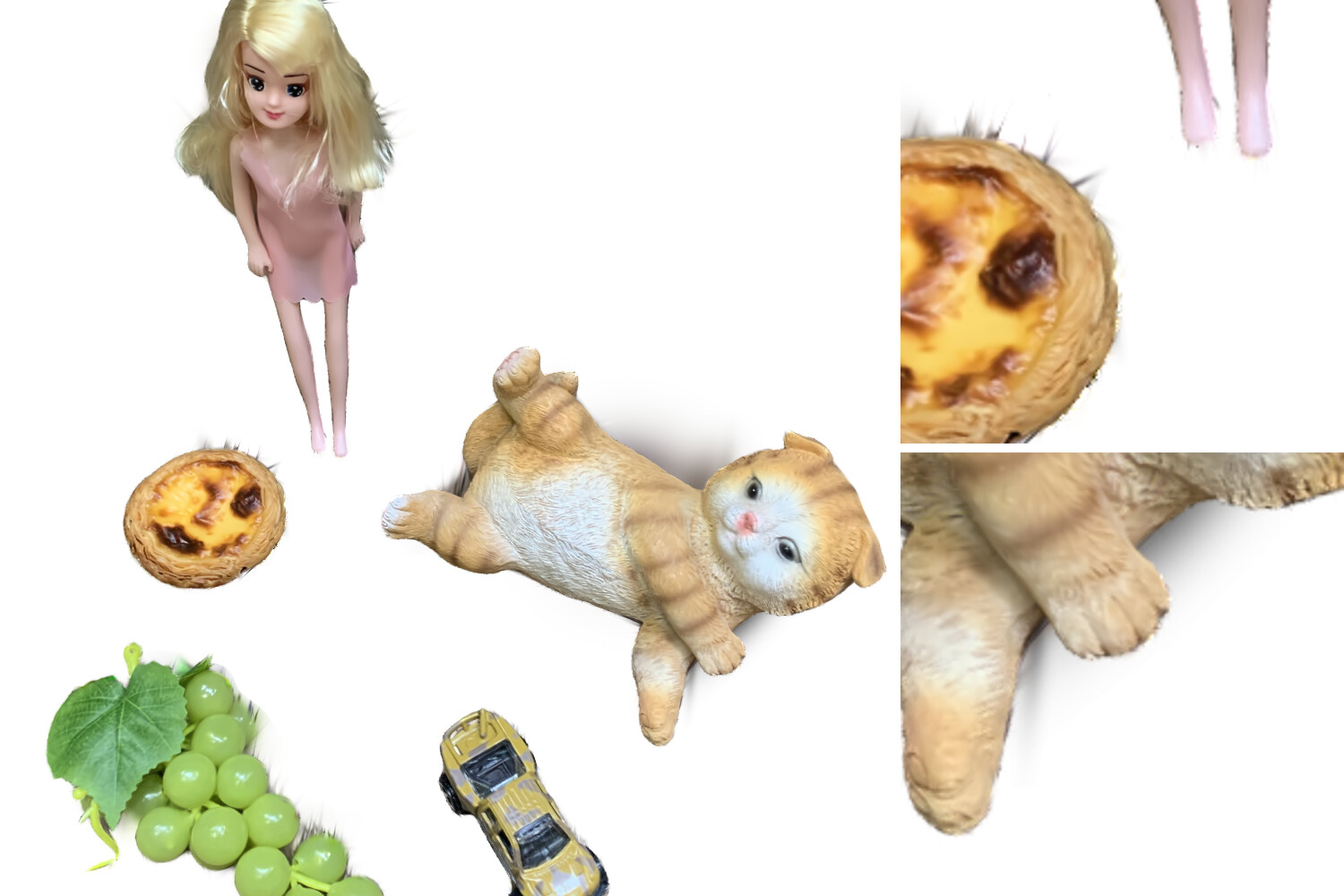} \\
    \vspace{-2pt}
    
    \raisebox{4mm}{\rotatebox{90}{\textbf{\methodname}}} &
    \includegraphics[width=0.237\linewidth]{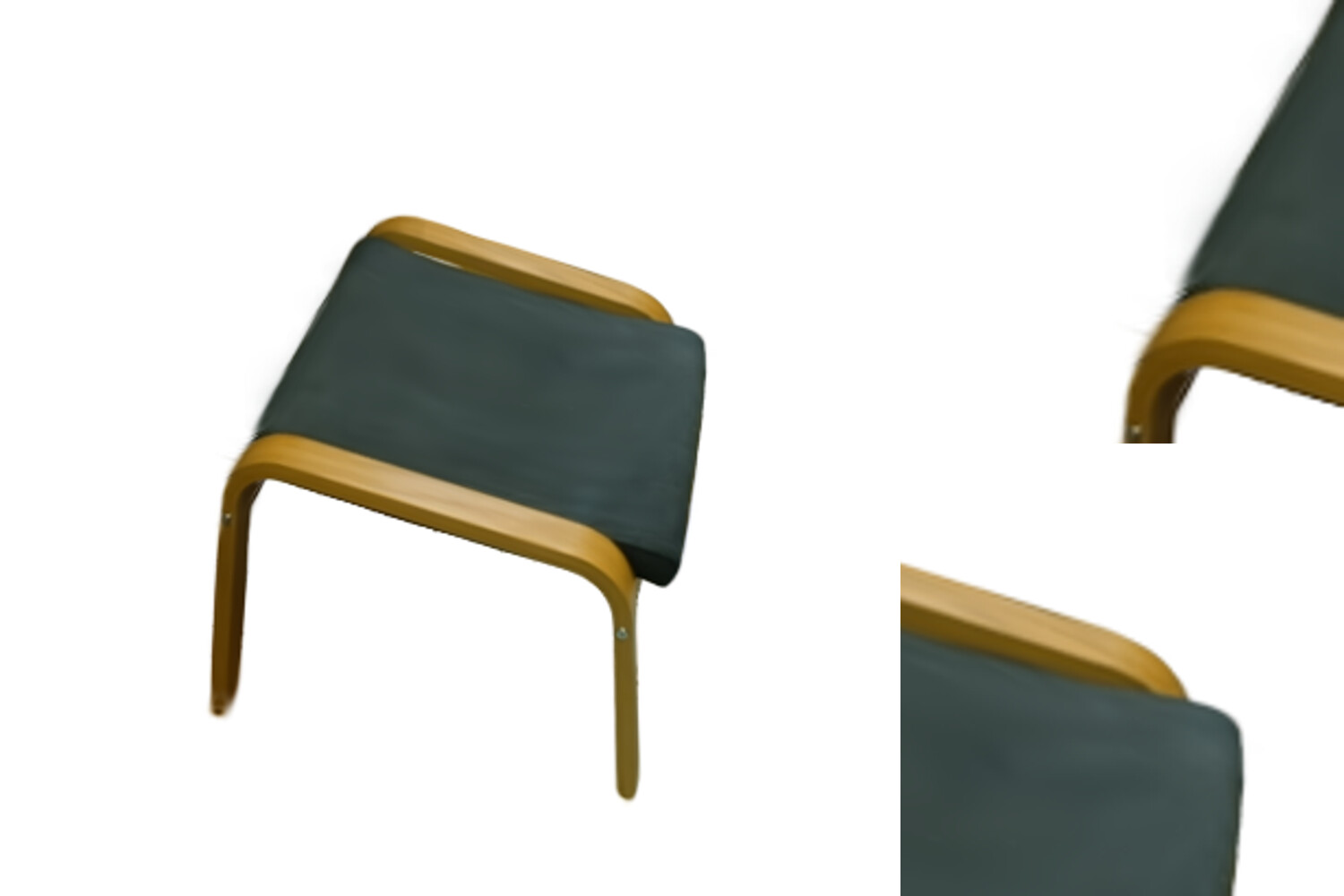} &
    \includegraphics[width=0.237\linewidth]{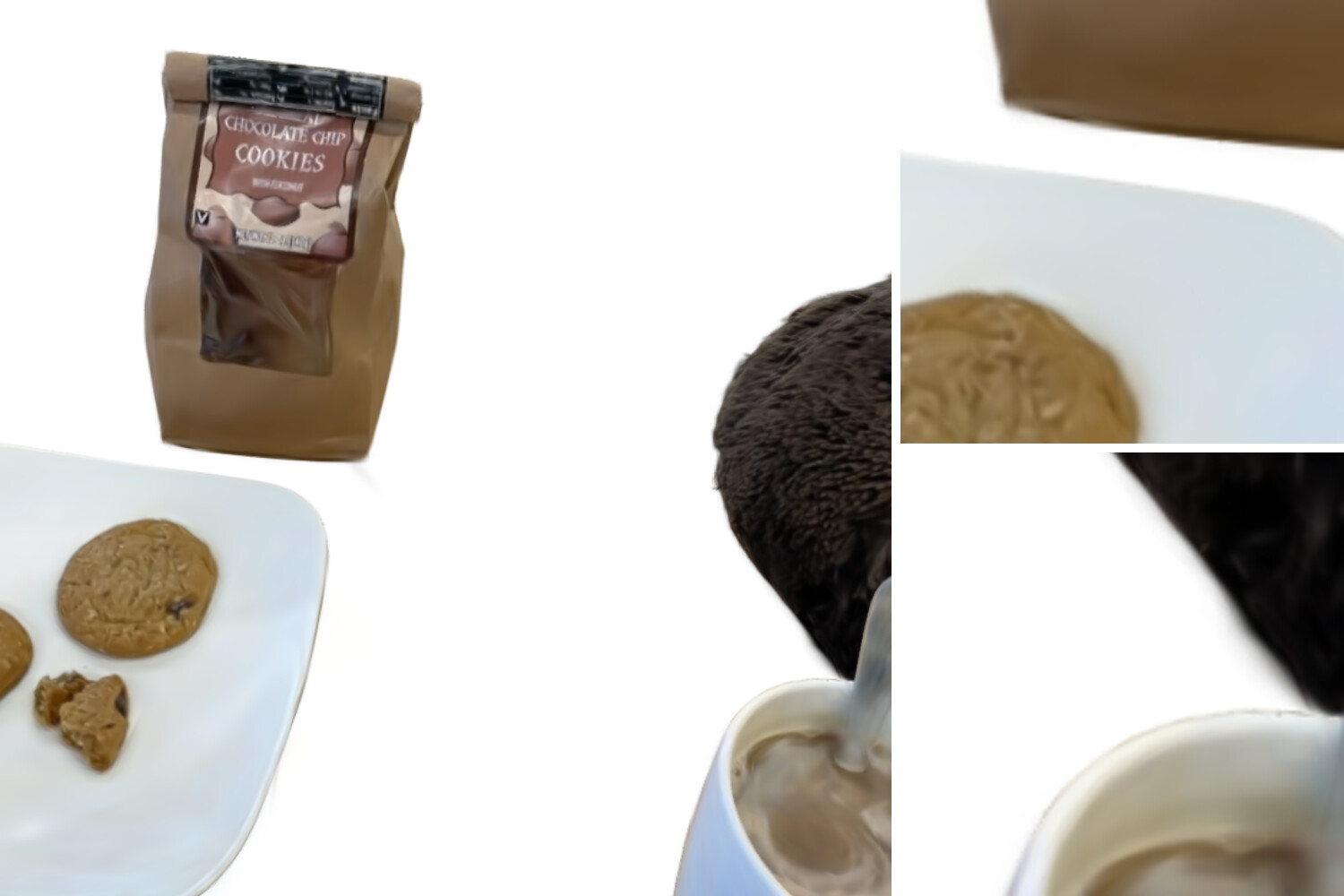} &
    \includegraphics[width=0.237\linewidth]{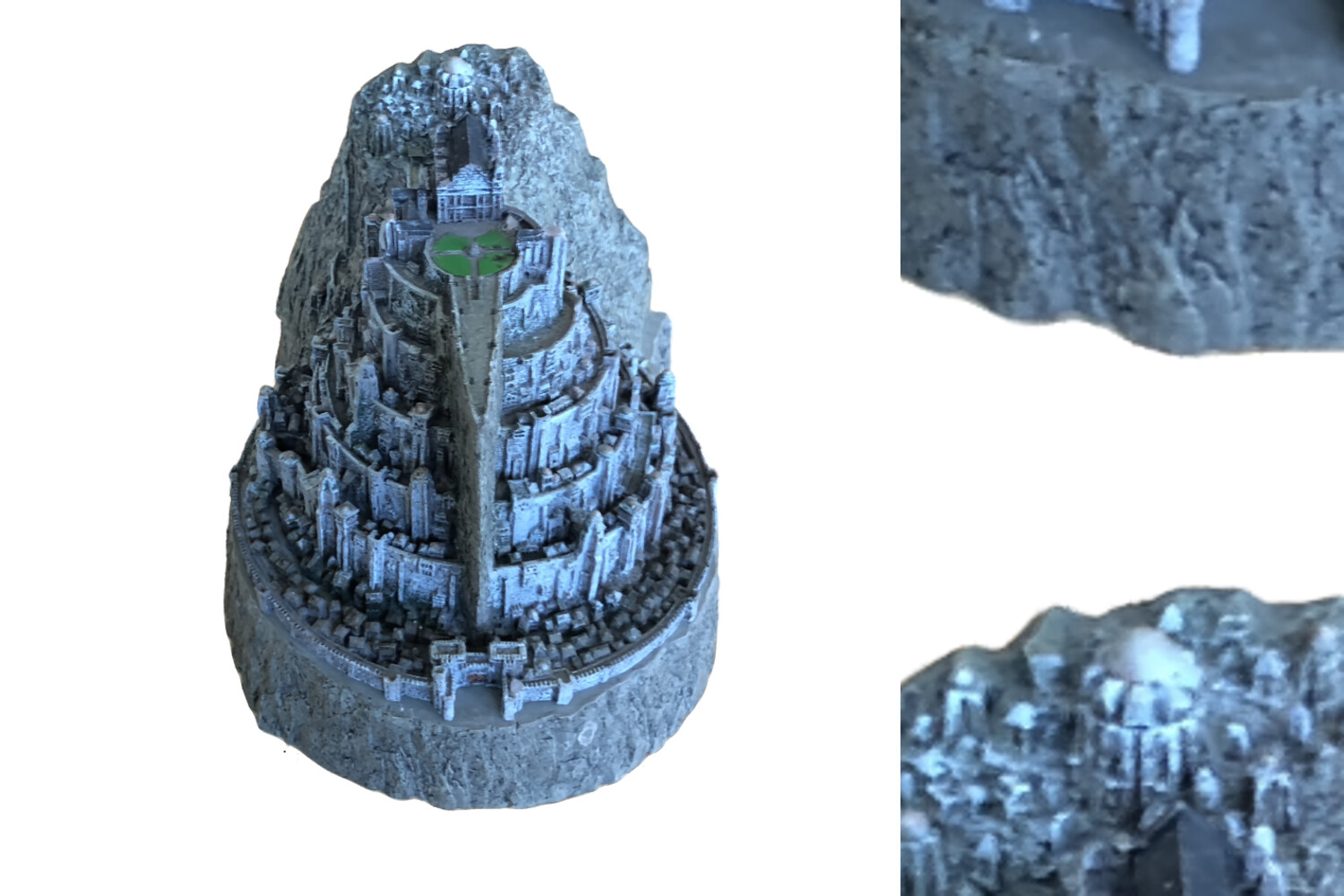} &
    \includegraphics[width=0.237\linewidth]{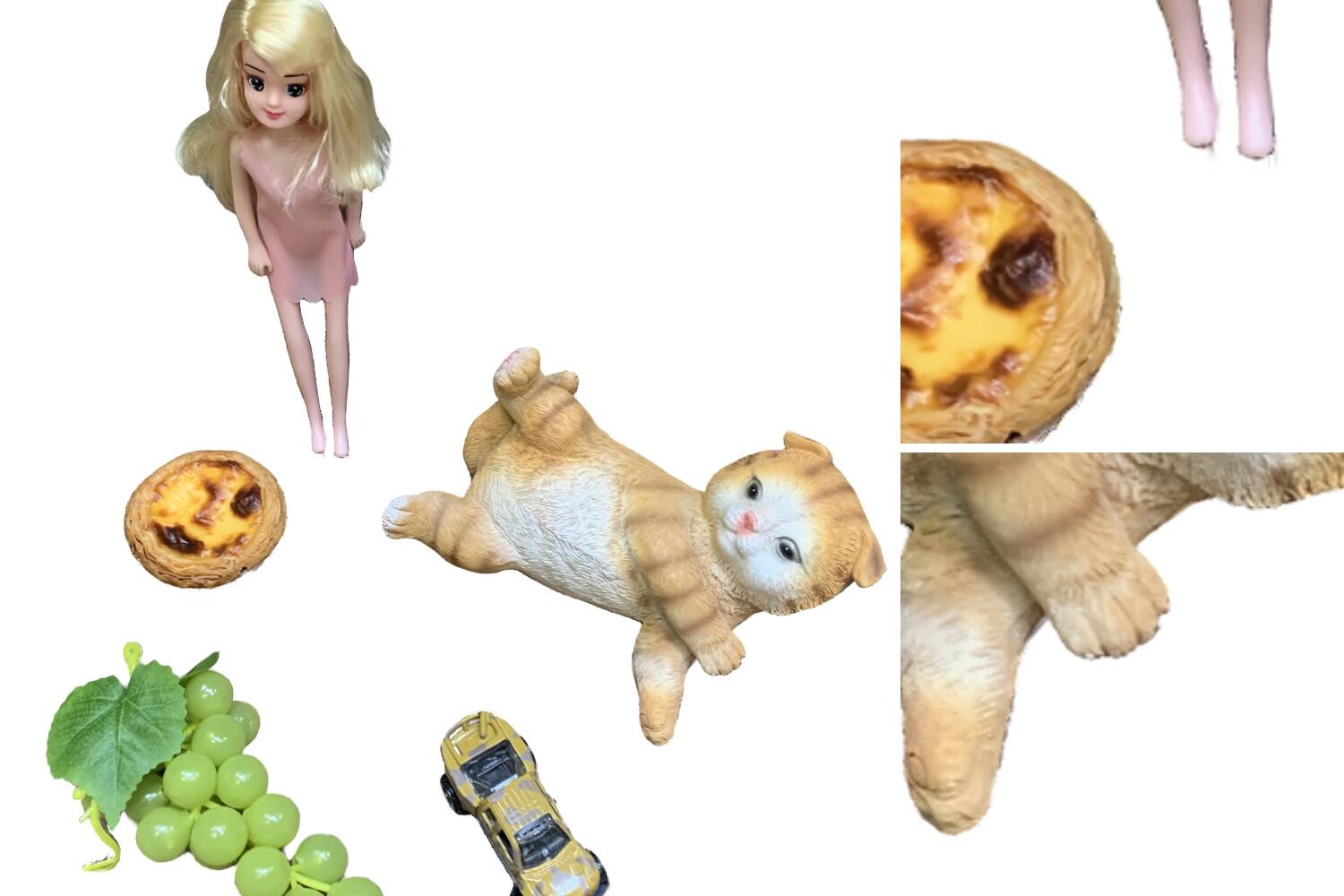} \\
    \vspace{-2pt}
        
    \end{tabular}
    \vspace{-4mm}
    \caption{Qualitative comparison across methods (rows) and scenes (columns): room (MipNeRF-360), teatime (LeRF), fortress (LLFF), and bench (3DOVS). Each cell shows the rendered result for that combination.}
    \label{fig:qualitative_results}
    \vspace{-1mm}
\end{figure*}

We show qualitative results in Fig.~\ref{fig:qualitative_results}. The boundaries obtained by our method are far more precise than those produced by other state of the art methods. We observe in the \textit{room} scene from the Mip-NeRF 360 dataset how our approach learns to disentangle the hidden Gaussians between the footstool and the sofa. This same capacity of disentangling Gaussians is visible in how our method manages to separate objects lying on top of a table or the ground, producing correct boundaries for the object.

\subsection{Ablation Study}
\label{sec:ablation_study}

To validate each of our contributions, we performed an ablation study, shown in Tab.~\ref{tab:ablation_study}. We report the detailed \textit{Extracted 3D Metrics} over the 4 datasets. The first row shows a vanilla baseline that just assigns each Gaussian the most frequent semantic class. The last row corresponds to our full method, where all proposed components are active. The intermediate rows illustrate the performance drop observed when deactivating each individual contribution. The results clearly highlight the importance of each component. The \textbf{Multiview Reprojection} module (\textbf{R}) improves 2D mask consistency in Mip-NeRF360 and LeRF, comprised of 360° scenes with occlusions and multiple training views, while offering no significant gain in LLFF and 3DOVS, which features mostly frontal views with minimal occlusion. The two loss terms are shown to be complementary: the \textbf{Boundary Loss} (\textbf{B}) refines fine details along visible boundaries, which the \textbf{Occupancy Loss} (\textbf{O}) cannot fully capture due to the discrete nature of its voxel-based formulation; conversely, \textbf{O} effectively penalizes unseen regions that \textbf{B} cannot address. 
Additional results, including an analysis of depth robustness and SAM2 failure cases, are provided in the supplementary material. We also include baselines 
training details, the intuition behind our loss weights, and visualizations of the voxel grids at object contact points used in our occupancy loss.

\section{Implementation Details}
\label{sec:implementation_details}

We follow the standard 2DGS training pipeline for 30K iterations using the original optimization and learning rate scheduling. At this point, we perform all preprocessing steps, including Multiview Reprojection, per-class Gaussian assignment, and the voxelization of the reference point clouds $V_c$. Afterwards, we train for additional 3K iterations adding our losses. The loss weights are set to $\lambda_{\text{bound}} = 0.5$ and $\lambda_{\text{occ}} = 10$. We sample $Z=20$ points per Gaussian, and the number of nearest neighbors $k$ is fixed to \textit{2000} across all scenes. Selected values for $Z$ and $k$ do not have a significant performance impact, we provide a detailed ablation in the supplemental material. We report the average training times for the standard 2DGS 30K iterations and the training time for our 3K iterations for all datasets (30K minutes/3K minutes): Mip-NeRF 360 (39/6), LeRF (40/4), LLFF (67/12), and 3DOVS (82/15). All experiments were conducted on an NVIDIA A100 GPU.

\section{Limitations}
\label{sec:limitations}

Our work relies on the inherently good 3D geometry provided by Gaussian Splatting; therefore, we can only perform as well as the quality of the initial reconstruction. Moreover, we do not infer or hallucinate the unseen parts of objects, which limits the completeness of extracted assets. Fortunately, our approach could be easily integrated with diffusion-based generation methods, such as \textit{InstaScene}, to address this limitation. We focus on a single objective and do not tackle complementary aspects often associated with semantic segmentation in Gaussian Splatting scenes, such as language grounding or mask ID correlation. We have concentrated solely on the 3D aspect of the problem, constructing Gaussian Splatting scenes that inherently represent the boundaries between objects. Our method does not ground classes with language embeddings, unlike approaches such as Dr. Splat~\cite{DrSplat}. Our approach also depends on 2D semantic segmentation models to provide reliable ID-consistent masks. While we can clean these masks and improve boundary consistency across views, we cannot address the problem of determining when two separate masks correspond to the same object, as Trace3D~\cite{Trace3D} does.

\section{Conclusions}
\label{sec:conclusion}

%We introduced \methodname, a novel framework for constructing 3D Gaussian Splatting scenes that produces geometrically consistent object boundaries. We achieve this by highlighting and addressing a major limitation of boundary optimization in Gaussian Splatting methods, which is its exclusive reliance on radiance. Through extensive experimental validation, our approach demonstrated superior performance over state of the art alternatives, laying a solid foundation for achieving high-quality object extraction in 3D Gaussian Splatting.

We introduced~\methodname, a novel framework for 3D Gaussian Splatting that significantly improves semantic object extraction by going beyond radiance-based supervision. Our method introduces two complementary losses: a 2D boundary loss that enforces semantic consistency at visible regions, and a 3D occupancy loss that regularizes non-visible or partially visible Gaussians. Together, these losses address a key limitation of prior work, hidden Gaussians causing boundary artifacts during object extraction. Through extensive experiments across multiple datasets and metrics,~\methodname~achieves state of the art results, laying a solid foundation for achieving high-quality object extraction in 3D Gaussian Splatting.

\section{Acknowledgements}
\label{acknowledgments}
This work was mainly supported by Arquimea Research Center
and Horizon Europe, Teaming for Excellence, under grant
agreement No 101059999, project QCircle. It was also partially supported by projects {\small PID2023-151351NB-I00,} and {\small PID2023-151184OB-I00,} funded by Ministerio de Ciencia e Innovación (MCIN)/ Agencia Estatal de Investigación (AEI) /10.13039/501100011033, by European Social Fund (ESF) Investing in your future and by \textit{ERDF, UE}.

\begin{comment}
MIP360 AVG 2DGS 30K TT 38.46 minutes -
MIP360 AVG 3K TT NO LOSSES 4.29 minutes -
MIP360 AVG 3K TT 4.29 minutes SAMP(0) -
MIP360 AVG 3K TT 5.3 minutes SAMP(1) -
MIP360 AVG 3K TT 5.57 minutes SAMP(5) -
MIP360 AVG 3K TT 6.42 minutes SAMP(20)

LERF AVG 2DGS 30K TT 40.3 minutes -
LERF AVG 3K TT NO LOSSES 2.5 minutes -
LERF AVG 3K TT 2.5 minutes SAMP(0) -
LERF AVG 3K TT 3.2 minutes SAMP(1) -
LERF AVG 3K TT 3.5 minutes SAMP(5) -
LERF AVG 3K TT 4.3 minutes SAMP(20)
\end{comment}

\appendix
\renewcommand{\thesection}{\Alph{section}}
\renewcommand{\thefigure}{A\arabic{figure}}
\renewcommand{\thetable}{A\arabic{table}}
\setcounter{figure}{0}
\setcounter{table}{0}

{
    \small
    \bibliographystyle{ieeenat_fullname}
    \bibliography{main}

@inproceedings{barron2022mipnerf360,
  title={Mip-nerf 360: Unbounded anti-aliased neural radiance fields},
  author={Barron, Jonathan T and Mildenhall, Ben and Verbin, Dor and Srinivasan, Pratul P and Hedman, Peter},
  booktitle={Proceedings of the IEEE/CVF conference on computer vision and pattern recognition},
  pages={5470--5479},
  year={2022}
}

@inproceedings{cen2023saga,
  title={Segment any 3d gaussians},
  author={Cen, Jiazhong and Fang, Jiemin and Yang, Chen and Xie, Lingxi and Zhang, Xiaopeng and Shen, Wei and Tian, Qi},
  booktitle={Proceedings of the AAAI conference on artificial intelligence},
  volume={39},
  number={2},
  pages={1971--1979},
  year={2025}
}

@inproceedings{chacko2025wacv,
  title={Lifting by gaussians: A simple, fast and flexible method for 3d instance segmentation},
  author={Chacko, Rohan and H{\"a}ni, Nicolai and Khaliullin, Eldar and Sun, Lin and Lee, Douglas},
  booktitle={2025 IEEE/CVF Winter Conference on Applications of Computer Vision (WACV)},
  pages={3497--3507},
  year={2025},
  organization={IEEE}
}

@inproceedings{Chao_2025_CVPR,
  title={Textured gaussians for enhanced 3d scene appearance modeling},
  author={Chao, Brian and Tseng, Hung-Yu and Porzi, Lorenzo and Gao, Chen and Li, Tuotuo and Li, Qinbo and Saraf, Ayush and Huang, Jia-Bin and Kopf, Johannes and Wetzstein, Gordon and others},
  booktitle={Proceedings of the Computer Vision and Pattern Recognition Conference},
  pages={8964--8974},
  year={2025}
}

@software{adobephotoshop,
    author = {{Adobe Inc.}},
    title = {Adobe Photoshop},
    url = {https://www.adobe.com/products/photoshop.html},
    version = {CC 2024}
}

@misc{paszke2019pytorchimperativestylehighperformance,
      title={PyTorch: An Imperative Style, High-Performance Deep Learning Library}, 
      author={Adam Paszke and Sam Gross and Francisco Massa and Adam Lerer and James Bradbury and Gregory Chanan and Trevor Killeen and Zeming Lin and Natalia Gimelshein and Luca Antiga and Alban Desmaison and Andreas Köpf and Edward Yang and Zach DeVito and Martin Raison and Alykhan Tejani and Sasank Chilamkurthy and Benoit Steiner and Lu Fang and Junjie Bai and Soumith Chintala},
      year={2019},
      eprint={1912.01703},
      archivePrefix={arXiv},
      primaryClass={cs.LG},
      url={https://arxiv.org/abs/1912.01703}, 
}

@inproceedings{chen2024dge,
  title={Dge: Direct gaussian 3d editing by consistent multi-view editing},
  author={Chen, Minghao and Laina, Iro and Vedaldi, Andrea},
  booktitle={European conference on computer vision},
  pages={74--92},
  year={2024},
  organization={Springer}
}

@inproceedings{gaussianeditor_cvpr_2024,
  title={Gaussianeditor: Swift and controllable 3d editing with gaussian splatting},
  author={Chen, Yiwen and Chen, Zilong and Zhang, Chi and Wang, Feng and Yang, Xiaofeng and Wang, Yikai and Cai, Zhongang and Yang, Lei and Liu, Huaping and Lin, Guosheng},
  booktitle={Proceedings of the IEEE/CVF conference on computer vision and pattern recognition},
  pages={21476--21485},
  year={2024}
}

@InProceedings{Cheng_2021_CVPR,
    author    = {Cheng, Bowen and Girshick, Ross and Dollar, Piotr and Berg, Alexander C. and Kirillov, Alexander},
    title     = {Boundary IoU: Improving Object-Centric Image Segmentation Evaluation},
    booktitle = {Proceedings of the IEEE/CVF Conference on Computer Vision and Pattern Recognition (CVPR)},
    month     = {June},
    year      = {2021},
    pages     = {15334-15342}
}

@inproceedings{ClickGaussian,
  title={Click-gaussian: Interactive segmentation to any 3d gaussians},
  author={Choi, Seokhun and Song, Hyeonseop and Kim, Jaechul and Kim, Taehyeong and Do, Hoseok},
  booktitle={European Conference on Computer Vision},
  pages={289--305},
  year={2024},
  organization={Springer}
}

@article{SplatFill,
  title={SplatFill: 3D Scene Inpainting via Depth-Guided Gaussian Splatting},
  author={Dahaghin, Mahtab and Padalkar, Milind G and Toso, Matteo and Del Bue, Alessio},
  journal={arXiv preprint arXiv:2509.07809},
  year={2025}
}

@inproceedings{NIPS2014_a5549f3f,
    author = {Dasgupta, Sanjoy and Kpotufe, Samory},
    booktitle = {Advances in Neural Information Processing Systems},
    pages = {},
    publisher = {Curran Associates, Inc.},
    title = {Optimal rates for k-NN density and mode estimation},
    volume = {27},
    year = {2014}
}

@inproceedings{guedon2024frosting,
  title={Gaussian frosting: Editable complex radiance fields with real-time rendering},
  author={Gu{\'e}don, Antoine and Lepetit, Vincent},
  booktitle={European conference on computer vision},
  pages={413--430},
  year={2024},
  organization={Springer}
}

@article{SAGD,
  title={SAGD: Boundary-enhanced segment anything in 3D Gaussian via Gaussian decomposition},
  author={Hu, Xu and Wang, Yuxi and Fan, Lue and Luo, Chuanchen and Fan, Junsong and Lei, Zhen and Li, Qing and Peng, Junran and Zhang, Zhaoxiang},
  journal={arXiv preprint arXiv:2401.17857},
  year={2024}
}

@inproceedings{Huang2DGS2024,
  title={2d gaussian splatting for geometrically accurate radiance fields},
  author={Huang, Binbin and Yu, Zehao and Chen, Anpei and Geiger, Andreas and Gao, Shenghua},
  booktitle={ACM SIGGRAPH 2024 conference papers},
  pages={1--11},
  year={2024}
}

@InProceedings{Huang_2025_CVPR,
    author    = {Huang, Sheng-Yu and Chou, Zi-Ting and Wang, Yu-Chiang Frank},
    title     = {3D Gaussian Inpainting with Depth-Guided Cross-View Consistency},
    booktitle = {Proceedings of the Computer Vision and Pattern Recognition Conference (CVPR)},
    month     = {June},
    year      = {2025},
    pages     = {26704-26713}
}

@article{jaganathan24iceg,
  title={Ice-g: Image conditional editing of 3d gaussian splats},
  author={Jaganathan, Vishnu and Huang, Hannah Hanyun and Irshad, Muhammad Zubair and Jampani, Varun and Raj, Amit and Kira, Zsolt},
  journal={arXiv preprint arXiv:2406.08488},
  year={2024}
}

@article{jain2024gaussiancut,
  title={Gaussiancut: Interactive segmentation via graph cut for 3d gaussian splatting},
  author={Jain, Umangi and Mirzaei, Ashkan and Gilitschenski, Igor},
  journal={Advances in Neural Information Processing Systems},
  volume={37},
  pages={89184--89212},
  year={2024}
}

@inproceedings{ILGS,
  title={Identity-aware language gaussian splatting for open-vocabulary 3d semantic segmentation},
  author={Jang, SungMin and Kim, Wonjun},
  booktitle={Proceedings of the IEEE/CVF International Conference on Computer Vision},
  pages={20467--20476},
  year={2025}
}

@article{kerbl3Dgaussians,
  title={3d gaussian splatting for real-time radiance field rendering.},
  author={Kerbl, Bernhard and Kopanas, Georgios and Leimk{\"u}hler, Thomas and Drettakis, George and others},
  journal={ACM Trans. Graph.},
  volume={42},
  number={4},
  pages={139--1},
  year={2023}
}

@inproceedings{lerf2023,
  title={Lerf: Language embedded radiance fields},
  author={Kerr, Justin and Kim, Chung Min and Goldberg, Ken and Kanazawa, Angjoo and Tancik, Matthew},
  booktitle={Proceedings of the IEEE/CVF international conference on computer vision},
  pages={19729--19739},
  year={2023}
}

@inproceedings{Kim_2025_ICCV,
  title={Robust 3d-masked part-level editing in 3d gaussian splatting with regularized score distillation sampling},
  author={Kim, Hayeon and Jang, Ji Ha and Chun, Se Young},
  booktitle={Proceedings of the IEEE/CVF International Conference on Computer Vision},
  pages={5501--5510},
  year={2025}
}

@article{MonteCarloGS,
  title={3d gaussian splatting as markov chain monte carlo},
  author={Kheradmand, Shakiba and Rebain, Daniel and Sharma, Gopal and Sun, Weiwei and Tseng, Yang-Che and Isack, Hossam and Kar, Abhishek and Tagliasacchi, Andrea and Yi, Kwang Moo},
  journal={Advances in Neural Information Processing Systems},
  volume={37},
  pages={80965--80986},
  year={2024}
}

@inproceedings{SAM,
  title={Segment anything},
  author={Kirillov, Alexander and Mintun, Eric and Ravi, Nikhila and Mao, Hanzi and Rolland, Chloe and Gustafson, Laura and Xiao, Tete and Whitehead, Spencer and Berg, Alexander C and Lo, Wan-Yen and others},
  booktitle={Proceedings of the IEEE/CVF international conference on computer vision},
  pages={4015--4026},
  year={2023}
}

@article{3DOVS_dataset_paper,
  title={Weakly supervised 3d open-vocabulary segmentation},
  author={Liu, Kunhao and Zhan, Fangneng and Zhang, Jiahui and Xu, Muyu and Yu, Yingchen and El Saddik, Abdulmotaleb and Theobalt, Christian and Xing, Eric and Lu, Shijian},
  journal={Advances in Neural Information Processing Systems},
  volume={36},
  pages={53433--53456},
  year={2023}
}

@article{Loftsgaarden1965ANE,
    title={A nonparametric estimate of a multivariate density function},
    author={Don O. Loftsgaarden and Charles P. Quesenberry},
    journal={Annals of Mathematical Statistics},
    year={1965},
    volume={36},
    pages={1049-1051}
}

@inproceedings{scaffoldgs,
    title={{Scaffold-GS}: Structured 3d gaussians for view-adaptive rendering},
    author={Lu, Tao and Yu, Mulin and Xu, Linning and Xiangli, Yuanbo and Wang, Limin and Lin, Dahua and Dai, Bo},
    booktitle={Proceedings of the IEEE/CVF Conference on Computer Vision and Pattern Recognition},
    pages={20654--20664},
    year={2024}
}

@article{lyu2024gaga,
  title={Gaga: Group any gaussians via 3d-aware memory bank},
  author={Lyu, Weijie and Li, Xueting and Kundu, Abhijit and Tsai, Yi-Hsuan and Yang, Ming-Hsuan},
  journal={arXiv preprint arXiv:2404.07977},
  year={2024}
}

@article{mazzucchelli2026virgi,
  title={VIRGi: View-dependent Instant Recoloring of 3D Gaussians Splats},
  author={Mazzucchelli, Alessio and Ojeda-Martin, Ivan and Rivas-Manzaneque, Fernando and Garces, Elena and Penate-Sanchez, Adrian and Moreno-Noguer, Francesc},
  journal={IEEE Transactions on Pattern Analysis and Machine Intelligence},
  year={2026},
  publisher={IEEE}
}

@article{mildenhall2019llff,
  title={Local light field fusion: Practical view synthesis with prescriptive sampling guidelines},
  author={Mildenhall, Ben and Srinivasan, Pratul P and Ortiz-Cayon, Rodrigo and Kalantari, Nima Khademi and Ramamoorthi, Ravi and Ng, Ren and Kar, Abhishek},
  journal={ACM Transactions on Graphics (ToG)},
  volume={38},
  number={4},
  pages={1--14},
  year={2019},
}

@inproceedings{mildenhall2020nerf,
  title={NeRF: Representing Scenes as Neural Radiance Fields for View Synthesis},
  author={Mildenhall, Ben and Srinivasan, Pratul P and Tancik, Matthew and Barron, Jonathan T and Ramamoorthi, Ravi and Ng, Ren},
  booktitle={European Conference on Computer Vision},
  pages={405--421},
  year={2020},
  organization={Springer}
}

@article{mirzaei2024reffusionreferenceadapteddiffusion,
  title={Reffusion: Reference adapted diffusion models for 3d scene inpainting},
  author={Mirzaei, Ashkan and De Lutio, Riccardo and Kim, Seung Wook and Acuna, David and Kelly, Jonathan and Fidler, Sanja and Gilitschenski, Igor and Gojcic, Zan},
  journal={arXiv preprint arXiv:2404.10765},
  year={2024}
}

@article{3dgrt2024,
  title={3d gaussian ray tracing: Fast tracing of particle scenes},
  author={Moenne-Loccoz, Nicolas and Mirzaei, Ashkan and Perel, Or and De Lutio, Riccardo and Martinez Esturo, Janick and State, Gavriel and Fidler, Sanja and Sharp, Nicholas and Gojcic, Zan},
  journal={ACM Transactions on Graphics (TOG)},
  volume={43},
  number={6},
  pages={1--19},
  year={2024},
}

@article{InstantNGP,
  title={Instant neural graphics primitives with a multiresolution hash encoding},
  author={M{\"u}ller, Thomas and Evans, Alex and Schied, Christoph and Keller, Alexander},
  journal={ACM transactions on graphics (TOG)},
  volume={41},
  number={4},
  pages={1--15},
  year={2022},
}

@inproceedings{qin2023langsplat,
  title={Langsplat: 3d language gaussian splatting},
  author={Qin, Minghan and Li, Wanhua and Zhou, Jiawei and Wang, Haoqian and Pfister, Hanspeter},
  booktitle={Proceedings of the IEEE/CVF Conference on Computer Vision and Pattern Recognition},
  pages={20051--20060},
  year={2024}
}

@inproceedings{clip2021,
  title={Learning transferable visual models from natural language supervision},
  author={Radford, Alec and Kim, Jong Wook and Hallacy, Chris and Ramesh, Aditya and Goh, Gabriel and Agarwal, Sandhini and Sastry, Girish and Askell, Amanda and Mishkin, Pamela and Clark, Jack and others},
  booktitle={International conference on machine learning},
  pages={8748--8763},
  year={2021},
  organization={PmLR}
}

@inproceedings{SAM2,
    title={{SAM} 2: Segment Anything in Images and Videos},
    author={Nikhila Ravi and Valentin Gabeur and Yuan-Ting Hu and Ronghang Hu and Chaitanya Ryali and Tengyu Ma and Haitham Khedr and Roman R{\"a}dle and Chloe Rolland and Laura Gustafson and Eric Mintun and Junting Pan and Kalyan Vasudev Alwala and Nicolas Carion and Chao-Yuan Wu and Ross Girshick and Piotr Dollar and Christoph Feichtenhofer},
    booktitle={The Thirteenth International Conference on Learning Representations},
    year={2025}
}

@inproceedings{nvos_ren_cvpr2022,
  title={Neural volumetric object selection},
  author={Ren, Zhongzheng and Agarwala, Aseem and Russell, Bryan and Schwing, Alexander G and Wang, Oliver},
  booktitle={Proceedings of the IEEE/CVF conference on computer vision and pattern recognition},
  pages={6133--6142},
  year={2022}
}

@InProceedings{Trace3D,
    author    = {Shen, Hongyu and Ni, Junfeng and Chen, Yixin and Li, Weishuo and Pei, Mingtao and Huang, Siyuan},
    title     = {Trace3D: Consistent Segmentation Lifting via Gaussian Instance Tracing},
    booktitle = {Proceedings of the IEEE/CVF International Conference on Computer Vision (ICCV)},
    month     = {October},
    year      = {2025},
    pages     = {6656-6666}
}

@inproceedings{flashsplat,
  title={Flashsplat: 2d to 3d gaussian splatting segmentation solved optimally},
  author={Shen, Qiuhong and Yang, Xingyi and Wang, Xinchao},
  booktitle={European Conference on Computer Vision},
  pages={456--472},
  year={2024},
  organization={Springer}
}

@inproceedings{shi2024language,
    title={Language embedded 3d gaussians for open-vocabulary scene understanding},
    author={Shi, Jin-Chuan and Wang, Miao and Duan, Hao-Bin and Guan, Shao-Hua},
    booktitle={Proceedings of the IEEE/CVF Conference on Computer Vision and Pattern Recognition},
    pages={5333--5343},
    year={2024}
}

@inproceedings{wang2024gscream,
  title={Learning 3d geometry and feature consistent gaussian splatting for object removal},
  author={Wang, Yuxin and Wu, Qianyi and Zhang, Guofeng and Xu, Dan},
  booktitle={European conference on computer vision},
  pages={1--17},
  year={2024},
  organization={Springer}
}

@inproceedings{wang2024view,
  title={View-consistent 3d editing with gaussian splatting},
  author={Wang, Yuxuan and Yi, Xuanyu and Wu, Zike and Zhao, Na and Chen, Long and Zhang, Hanwang},
  booktitle={European conference on computer vision},
  pages={404--420},
  year={2024},
  organization={Springer}
}

@inproceedings{gaussctrl2024,
  title={Gaussctrl: Multi-view consistent text-driven 3d gaussian splatting editing},
  author={Wu, Jing and Bian, Jia-Wang and Li, Xinghui and Wang, Guangrun and Reid, Ian and Torr, Philip and Prisacariu, Victor Adrian},
  booktitle={European conference on computer vision},
  pages={55--71},
  year={2024},
  organization={Springer}
}

@inproceedings{wu20253dgut,
  title={3dgut: Enabling distorted cameras and secondary rays in gaussian splatting},
  author={Wu, Qi and Esturo, Janick Martinez and Mirzaei, Ashkan and Moenne-Loccoz, Nicolas and Gojcic, Zan},
  booktitle={Proceedings of the Computer Vision and Pattern Recognition Conference},
  pages={26036--26046},
  year={2025}
}

@article{wu2024opengaussian,
  title={Opengaussian: Towards point-level 3d gaussian-based open vocabulary understanding},
  author={Wu, Yanmin and Meng, Jiarui and Li, Haijie and Wu, Chenming and Shi, Yahao and Cheng, Xinhua and Zhao, Chen and Feng, Haocheng and Ding, Errui and Wang, Jingdong and others},
  journal={Advances in Neural Information Processing Systems},
  volume={37},
  pages={19114--19138},
  year={2024}
}

@InProceedings{InstaScene,
    author    = {Yang, Zesong and Yang, Bangbang and Dong, Wenqi and Cao, Chenxuan and Cui, Liyuan and Ma, Yuewen and Cui, Zhaopeng and Bao, Hujun},
    title     = {InstaScene: Towards Complete 3D Instance Decomposition and Reconstruction from Cluttered Scenes},
    booktitle = {Proceedings of the IEEE/CVF International Conference on Computer Vision (ICCV)},
    month     = {October},
    year      = {2025},
    pages     = {7771-7781}
}

@inproceedings{gaussian_grouping,
  title={Gaussian grouping: Segment and edit anything in 3d scenes},
  author={Ye, Mingqiao and Danelljan, Martin and Yu, Fisher and Ke, Lei},
  booktitle={European conference on computer vision},
  pages={162--179},
  year={2024},
  organization={Springer}
}

@InProceedings{PanoGS,
    author    = {Zhai, Hongjia and Li, Hai and Li, Zhenzhe and Pan, Xiaokun and He, Yijia and Zhang, Guofeng},
    title     = {PanoGS: Gaussian-based Panoptic Segmentation for 3D Open Vocabulary Scene Understanding},
    booktitle = {Proceedings of the Computer Vision and Pattern Recognition Conference (CVPR)},
    month     = {June},
    year      = {2025},
    pages     = {14114-14124}
}

@InProceedings{COBGS,
    author    = {Zhang, Jiaxin and Jiang, Junjun and Chen, Youyu and Jiang, Kui and Liu, Xianming},
    title     = {COB-GS: Clear Object Boundaries in 3DGS Segmentation Based on Boundary-Adaptive Gaussian Splitting},
    booktitle = {Proceedings of the Computer Vision and Pattern Recognition Conference (CVPR)},
    month     = {June},
    year      = {2025},
    pages     = {19335-19344}
}

@inproceedings{zhang20243DitScene,
  title={3DitScene: Editing Any Scene via Language-guided Disentangled Gaussian Splatting},
  author={Zhang, Qihang and Xu, Yinghao and Wang, Chaoyang and Lee, Hsin-Ying and Wetzstein, Gordon and Zhou, Bolei and Yang, Ceyuan},
  booktitle={The Thirteenth International Conference on Learning Representations},
  year = {2025}
}

@article{ZHOU2025104362,
  title={High-fidelity 3D Gaussian inpainting: Preserving multi-view consistency and photorealistic details},
  author={Zhou, Jun and Li, Dinghao and Li, Nannan and Wang, Mingjie},
  journal={Computers \& Graphics},
  pages={104362},
  year={2025},
  publisher={Elsevier}
}

@inproceedings{zhou2024feature,
  title={Feature 3dgs: Supercharging 3d gaussian splatting to enable distilled feature fields},
  author={Zhou, Shijie and Chang, Haoran and Jiang, Sicheng and Fan, Zhiwen and Zhu, Zehao and Xu, Dejia and Chari, Pradyumna and You, Suya and Wang, Zhangyang and Kadambi, Achuta},
  booktitle={Proceedings of the IEEE/CVF Conference on Computer Vision and Pattern Recognition},
  pages={21676--21685},
  year={2024}
}

@InProceedings{Unilift,
    author    = {Zhu, Runsong and Qiu, Shi and Liu, Zhengzhe and Hui, Ka-Hei and Wu, Qianyi and Heng, Pheng-Ann and Fu, Chi-Wing},
    title     = {Rethinking End-to-End 2D to 3D Scene Segmentation in Gaussian Splatting},
    booktitle = {Proceedings of the Computer Vision and Pattern Recognition Conference (CVPR)},
    month     = {June},
    year      = {2025},
    pages     = {3656-3665}
}

@InProceedings{ObjectGS,
    author    = {Zhu, Ruijie and Yu, Mulin and Xu, Linning and Jiang, Lihan and Li, Yixuan and Zhang, Tianzhu and Pang, Jiangmiao and Dai, Bo},
    title     = {ObjectGS: Object-aware Scene Reconstruction and Scene Understanding via Gaussian Splatting},
    booktitle = {Proceedings of the IEEE/CVF International Conference on Computer Vision (ICCV)},
    month     = {October},
    year      = {2025},
    pages     = {8350-8360}
}

@article{FMGS_Zuo_IJCV2024,
  title={Fmgs: Foundation model embedded 3d gaussian splatting for holistic 3d scene understanding},
  author={Zuo, Xingxing and Samangouei, Pouya and Zhou, Yunwen and Di, Yan and Li, Mingyang},
  journal={International Journal of Computer Vision},
  volume={133},
  number={2},
  pages={611--627},
  year={2025},
  publisher={Springer}
}

@InProceedings{AG2aussian,
    author    = {Wang, Zhaonan and Li, Manyi and Tu, Changhe},
    title     = {AG2aussian: Anchor-Graph Structured Gaussian Splatting for Instance-Level 3D Scene Understanding and Editing},
    booktitle = {Proceedings of the IEEE/CVF International Conference on Computer Vision (ICCV)},
    month     = {October},
    year      = {2025},
    pages     = {26806-26816}
}

@InProceedings{DrSplat,
    author    = {Jun-Seong, Kim and Kim, GeonU and Yu-Ji, Kim and Wang, Yu-Chiang Frank and Choe, Jaesung and Oh, Tae-Hyun},
    title     = {Dr. Splat: Directly Referring 3D Gaussian Splatting via Direct Language Embedding Registration},
    booktitle = {Proceedings of the Computer Vision and Pattern Recognition Conference (CVPR)},
    month     = {June},
    year      = {2025},
    pages     = {14137-14146}
}
}

\clearpage

\section{Hyperparameter Sensitivity Analysis}

We have performed a sensitivity analysis for $k$, the number of sampled points to obtain the density of the point cloud, and for $Z$, the number of points that are sampled from each Gaussian when computing the occupancy loss $\mathcal{L}_{occ}$. Our detailed analysis for $k$ is presented in Tab.~\ref{tab:ablation_voxel_size}, and for $Z$ in Tab.~\ref{tab:ablation_sample_points_all}. The sensitivity of $k$ in our approach is very small, it can be seen that there is nearly no change in rendering quality ($PSNR$). When analyzing the boundary metrics ($IoU, BIoU$), the differences are also minor with small oscillations between datasets. We chose a value that slightly balances these changes. When assessing the sensitivity of $Z$ we see the same behavior, no real difference in rendering quality, and an even smaller oscillation on boundary metrics. The only limitation comes from sampling a number of points bigger or equal than 50 as we run out of memory in the GPU in some scenes. We chose $Z=20$, but any value that does not overload the VRAM of the GPU is fine as can be observed in Tab.~\ref{tab:ablation_sample_points_all}. It can be seen that the selection of both $k$ and $Z$ has a very small impact on performance, which shows that our method is robust.

\begin{table}[t]
\centering
\caption{Sensitivity analysis on the number of points ($k$) used to compute the voxel density. Extracted metrics (PSNR, IoU, BIoU) across Mip-NeRF 360, LeRF, LLFF, and 3DOVS. }
\label{tab:ablation_voxel_size}
\setlength{\tabcolsep}{2pt}
\resizebox{0.475\textwidth}{!}{%
\begin{tabular}{c ccc ccc ccc ccc}
\toprule
$k$ & \multicolumn{3}{c}{Mip-NeRF 360} & \multicolumn{3}{c}{LeRF} & \multicolumn{3}{c}{LLFF} & \multicolumn{3}{c}{3DOVS} \\
\cmidrule(lr){2-4} \cmidrule(lr){5-7} \cmidrule(lr){8-10} \cmidrule(lr){11-13}
& PSNR & IoU & BIoU & PSNR & IoU & BIoU & PSNR & IoU & BIoU & PSNR & IoU & BIoU \\
\midrule
250  & 29.07 & 92.0 & 85.3 & 24.98 & 90.4 & 85.3 & 24.83 & 93.0 & 79.9 & 26.47 & 93.3 & 87.6 \\
500  & 29.08 & 92.0 & 85.6 & 25.00 & 90.1 & 84.9 & 24.85 & 93.0 & 80.2 & 26.48 & 93.3 & 87.5 \\
1000 & 29.09 & 92.0 & 85.7 & 25.02 & 90.0 & 84.5 & 24.86 & 93.0 & 80.7 & 26.48 & 93.3 & 87.5 \\
2000 & 29.10 & 92.0 & 85.8 & 25.03 & 89.4 & 83.6 & 24.87 & 93.0 & 80.7 & 26.49 & 93.2 & 87.3 \\
3000 & 29.10 & 92.0 & 85.8 & 25.04 & 88.9 & 82.9 & 24.87 & 92.9 & 80.7 & 26.49 & 93.2 & 87.2 \\
4000 & 29.10 & 92.0 & 85.8 & 25.04 & 88.8 & 82.6 & 24.87 & 92.9 & 80.6 & 26.49 & 93.1 & 87.2 \\
5000 & 29.10 & 91.9 & 85.8 & 25.04 & 88.3 & 82.0 & 24.88 & 92.9 & 80.5 & 26.49 & 93.0 & 87.0 \\
\bottomrule
\end{tabular}
}
\end{table}

\begin{table}[t]
\centering
\caption{Sensitivity analysis of the number of sampled points $Z$. Extracted metrics (PSNR, IoU, BIoU) across Mip-NeRF 360, LeRF, LLFF, and 3DOVS. $OOM$ stands for \textit{Out Of Memory}, and thus, the method could not be run. }
\label{tab:ablation_sample_points_all}
\setlength{\tabcolsep}{2pt}
\resizebox{0.475\textwidth}{!}{%
\begin{tabular}{c ccc ccc ccc ccc}
\toprule
$Z$ & \multicolumn{3}{c}{Mip-NeRF 360} & \multicolumn{3}{c}{LeRF} & \multicolumn{3}{c}{LLFF} & \multicolumn{3}{c}{3DOVS} \\
\cmidrule(lr){2-4} \cmidrule(lr){5-7} \cmidrule(lr){8-10} \cmidrule(lr){11-13}
& PSNR & IoU & BIoU & PSNR & IoU & BIoU & PSNR & IoU & BIoU & PSNR & EIoU & BIoU \\
\midrule
1  & 29.10 & 92.0 & 85.7 & 25.03 & 89.4 & 83.6 & 24.87 & 93.0 & 80.6 & 26.50 & 93.2 & 87.2 \\
2  & 29.10 & 92.0 & 85.7 & 25.04 & 89.3 & 83.5 & 24.87 & 93.0 & 80.6 & 26.49 & 93.3 & 87.3 \\
5  & 29.10 & 92.0 & 85.8 & 25.04 & 89.4 & 83.5 & 24.87 & 93.0 & 80.6 & 26.49 & 93.1 & 87.4 \\
10 & 29.10 & 92.0 & 85.8 & 25.04 & 89.4 & 83.6 & 24.88 & 93.0 & 80.6 & 26.49 & 93.3 & 87.3 \\
20 & 29.10 & 92.0 & 85.8 & 25.03 & 89.4 & 83.6 & 24.87 & 93.0 & 80.7 & 26.49 & 93.2 & 87.3 \\
50 & 29.10 & 92.0 & 85.8 & 25.03 & 89.4 & 83.6 & \multicolumn{3}{c}{$OOM$} & \multicolumn{3}{c}{$OOM$} \\
\bottomrule
\end{tabular}
}
\end{table}

 \begin{figure}[t]
    \centering
        \includegraphics[trim=0 0 0 13, clip, width=1\linewidth]{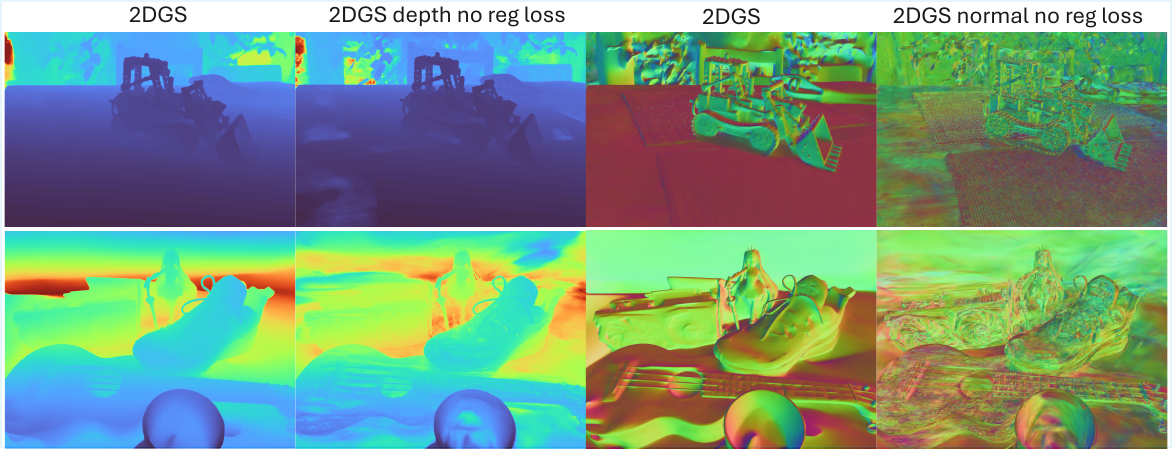}
    \vspace{-7mm}
    \caption{2DGS Depth (regularized vs. unregularized), 2DGS Normals (regularized vs. unregularized)
}
     \label{fig:depth_noise}
\end{figure}

\section{Depth Robustness Analysis}
 To test robustness under degraded geometric supervision, we evaluated performance under noisy depth and surface estimation obtained by retraining the 2DGS backbone without depth and normal regularization losses (Fig.~\ref{fig:depth_noise}). We then applied BEA-GS using these degraded depth maps. Tab~\ref{tab:extracted_results} reports extracted IoU and BIoU for the original setting (BEA-GS) and the degraded-geometry setting (BEA-Noisy). On datasets with stable reconstructions (Mip-NeRF360, LLFF), performance remains consistent. On more challenging datasets with weaker capture conditions (LeRF, 3DOVS), we observe a moderate performance drop, indicating that reconstruction quality affects performance, but not to the extent of causing optimization failure.

\begin{table}[t]
\centering
\caption{Depth robustness analysis. Extracted metrics (IoU, BIoU) across Mip-NeRF 360, LeRF, LLFF, and 3DOVS.}
\setlength{\tabcolsep}{3pt}
\resizebox{\columnwidth}{!}{%
\begin{tabular}{l cc cc cc cc}

 & \multicolumn{2}{c}{MipNeRF360}
 & \multicolumn{2}{c}{LeRF}
 & \multicolumn{2}{c}{LLFF}
 & \multicolumn{2}{c}{3DOVS} \\

\midrule

Method
& IoU & BIoU
& IoU & BIoU
& IoU & BIoU
& IoU & BIoU \\

\midrule

BEA-GS
& 92.0 & 85.8
& 89.4 & 83.6
& 93.0 & 80.7
& 93.2 & 87.3 \\

BEA-Noisy
& 92.3 & 86.1
& 88.4 & 81.6
& 92.8 & 80.7
& 89.8 & 83.9 \\

\bottomrule
\end{tabular}
}
\label{tab:extracted_results}
\end{table}

\section{SAM2  Reprojection Analysis}

\begin{figure}[h]
    \vspace{-2mm}
    \centering
    \includegraphics[width=1\linewidth]{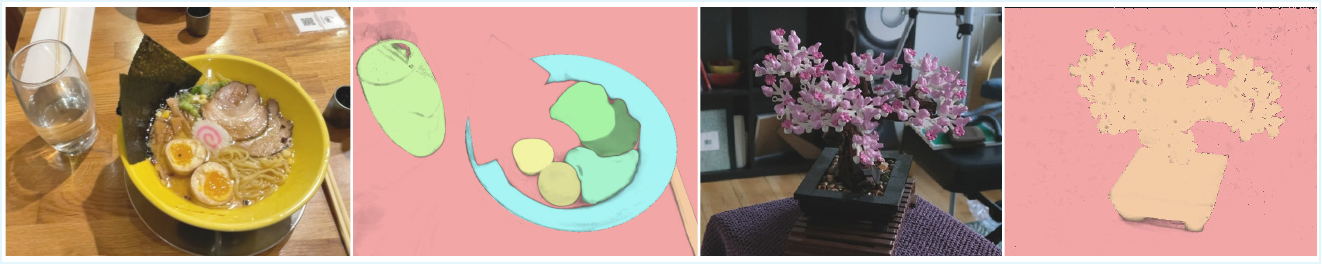}
    \vspace{-8mm}
    \caption{SAM2 probability masks after reprojection}
    \label{fig:reprojection_anal}
    \vspace{-4mm}
\end{figure}

We visualize reprojection by coloring each 2D pixel with its class assignment probability (Fig.~\ref{fig:reprojection_anal}), where brighter colors indicate higher consensus. These probabilities correspond to the per-view reprojection before computing $M'=\textrm{argmax}(M_\phi)$ (Fig.~3, main paper). In the left example, thin or occluded parts (e.g., eggs, pork) are missed by SAM2 in some views, yielding low-confidence regions, but reprojection recovers coherent boundaries when the object is correctly segmented in at least 50\% of frames in which it is visible. Transparent elements (e.g., glass) mainly cause noise due to depth ambiguity. In the bonsai scene (Fig.~\ref{fig:reprojection_anal}, right), dark low-texture areas near the pot base show consistent SAM2 failures across views, leading to regions that reprojection cannot correct. For the  baseline without reprojection  (Tab~3, row~4, main paper), BEA-GS remains competitive and reprojection yields the smallest performance change.

\begin{figure}[t]
    \centering
    \includegraphics[width=1\linewidth]{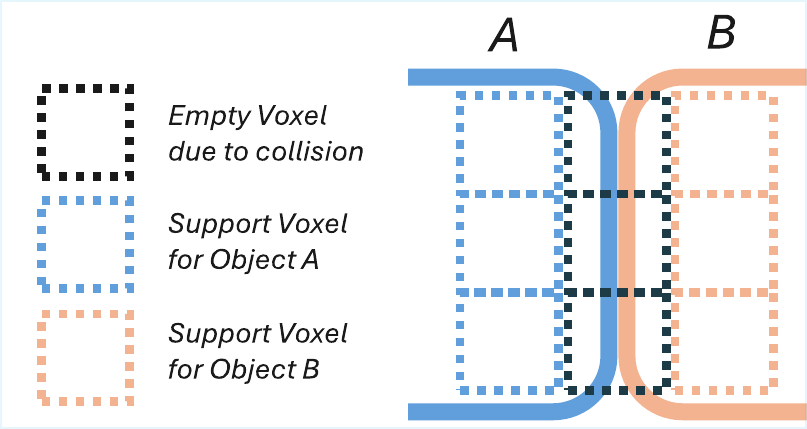}
    \caption{Visualization of edge cases where two objects are in contact. Because voxel validity is determined via neighborhood density checks, erosion effects are mitigated at contact interfaces, ensuring sample points that happens in empty voxels (indicated by the black box) remain valid.}
    \label{fig:voxel_visual} % Moved to AFTER the caption
\end{figure}

\begin{figure*}[t]
    \centering
    \setlength{\tabcolsep}{1pt}
    \renewcommand{\arraystretch}{0.99}

    \begin{tabular}{ccc}
        \includegraphics[width=0.33\linewidth]{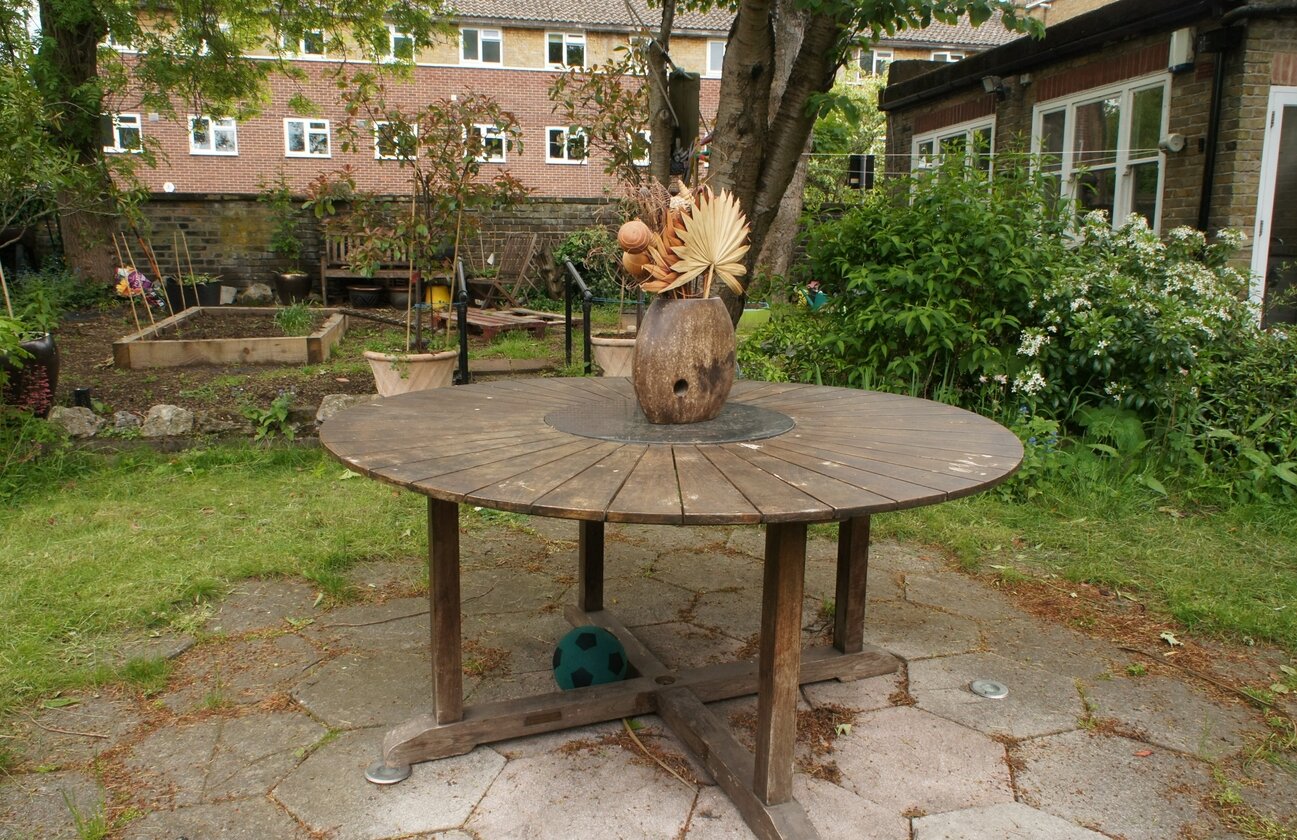} &
        \includegraphics[width=0.33\linewidth]{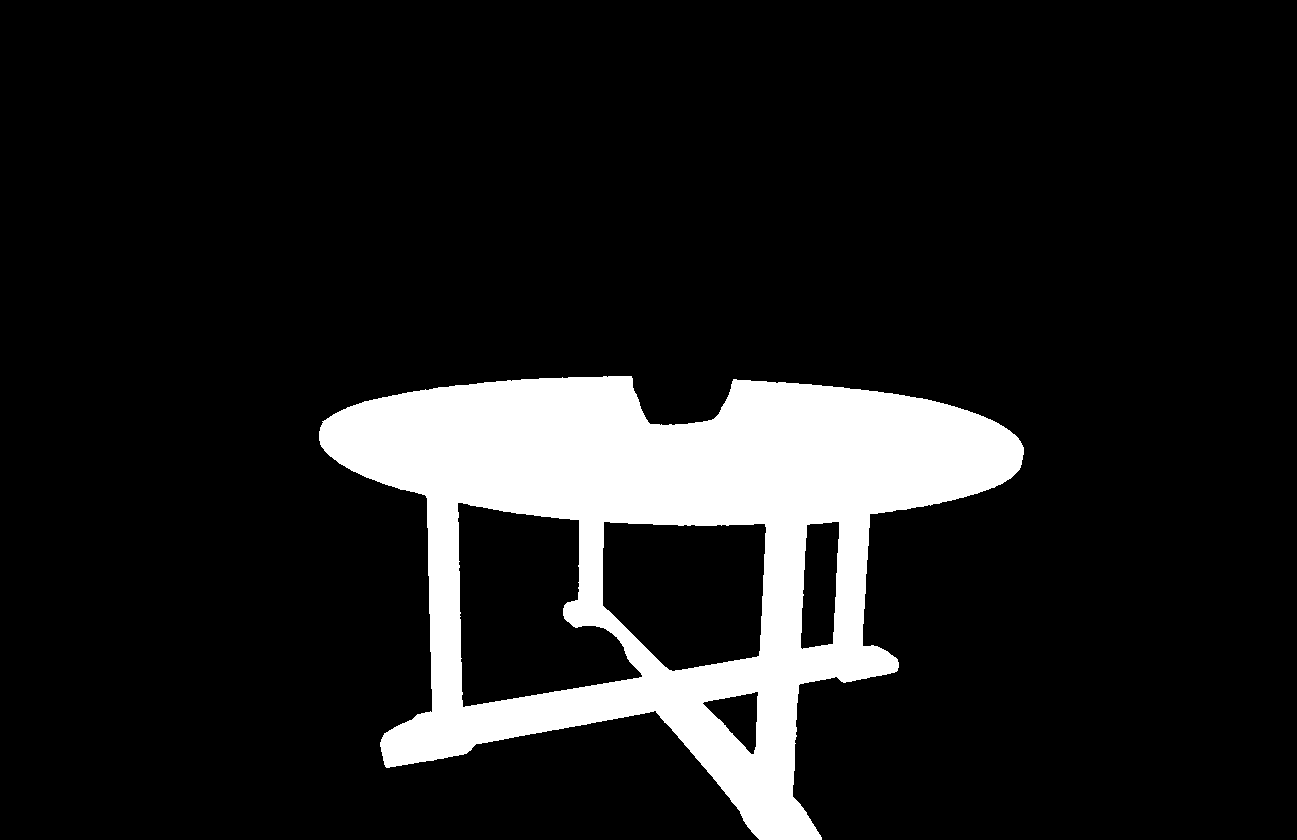} &
        \includegraphics[width=0.33\linewidth]{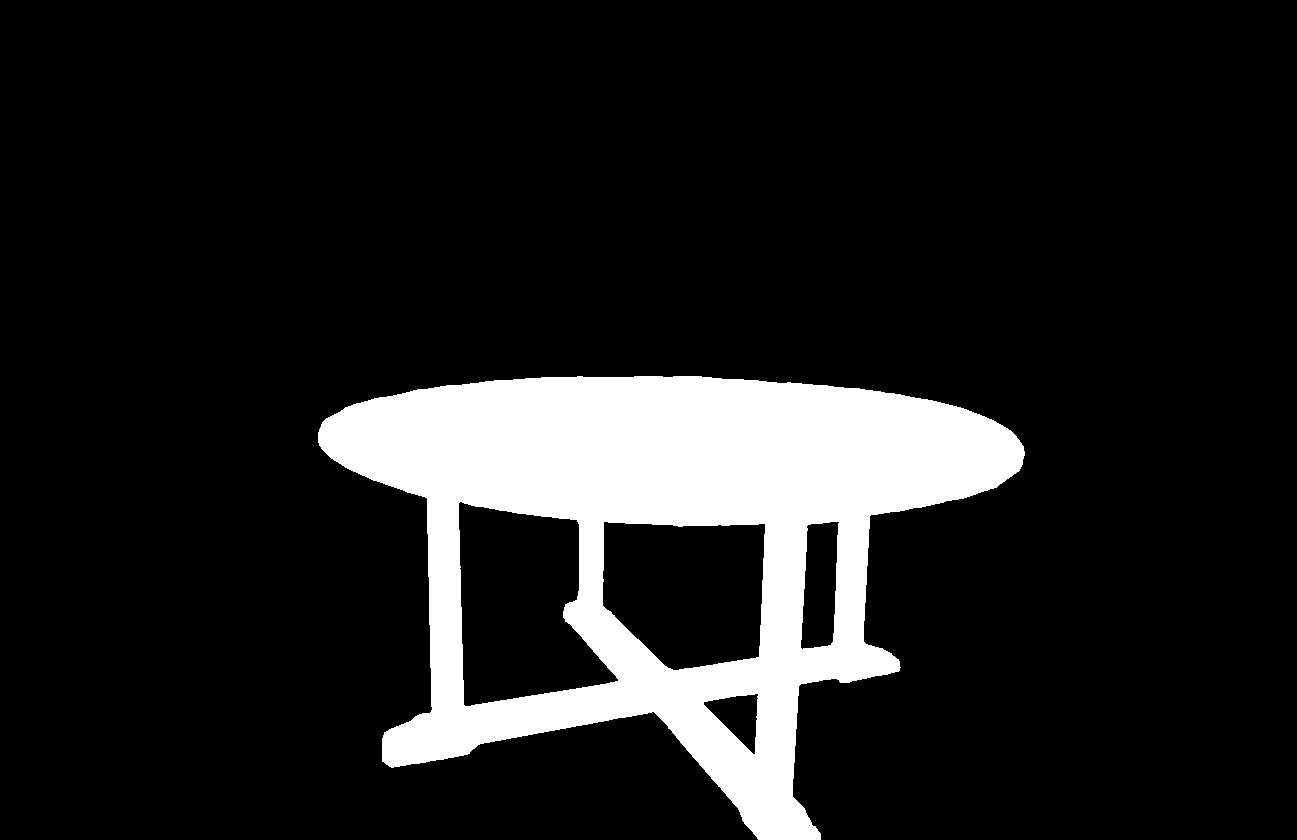} \\

        \includegraphics[width=0.33\linewidth]{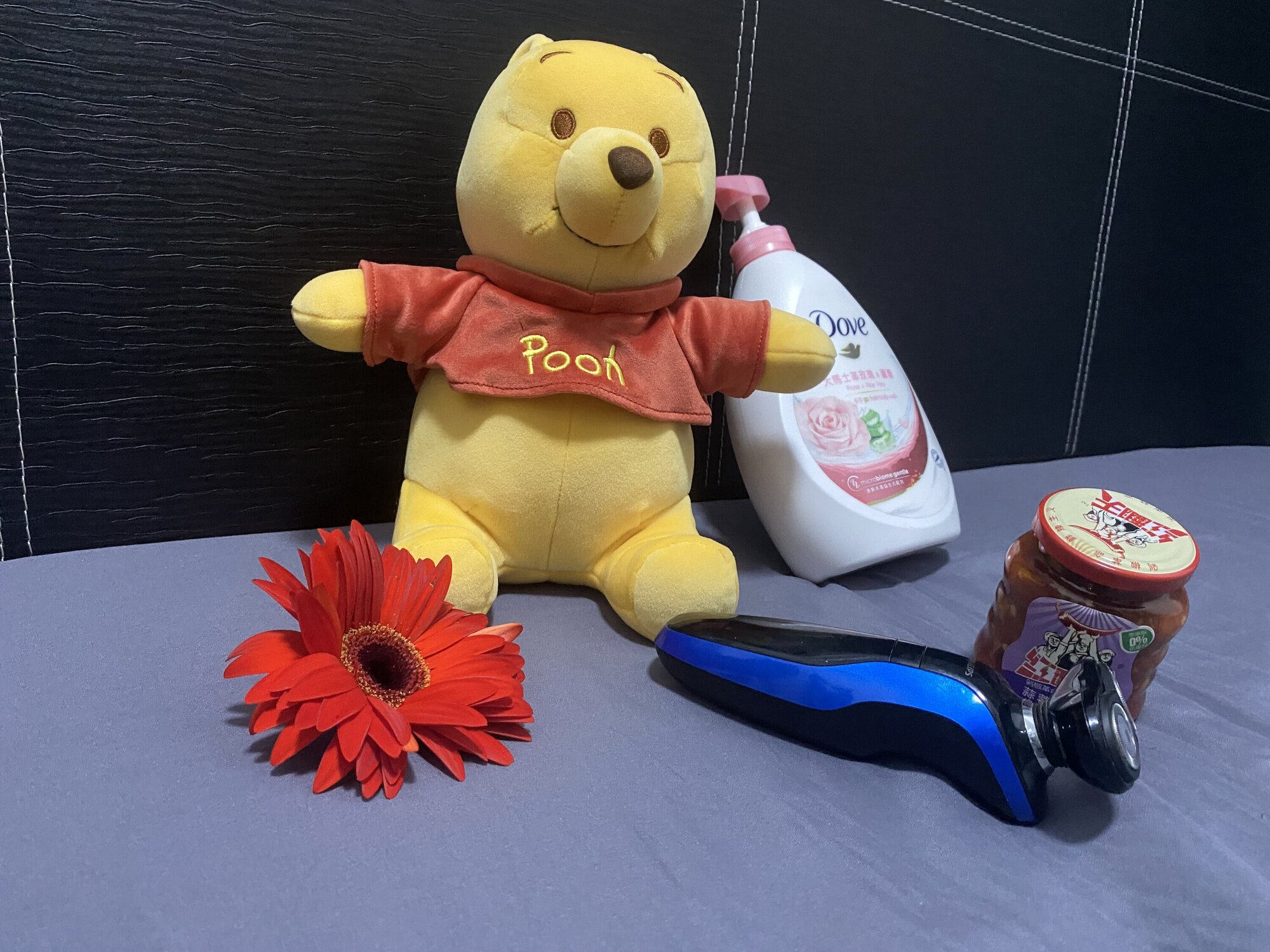} &
        \includegraphics[width=0.33\linewidth]{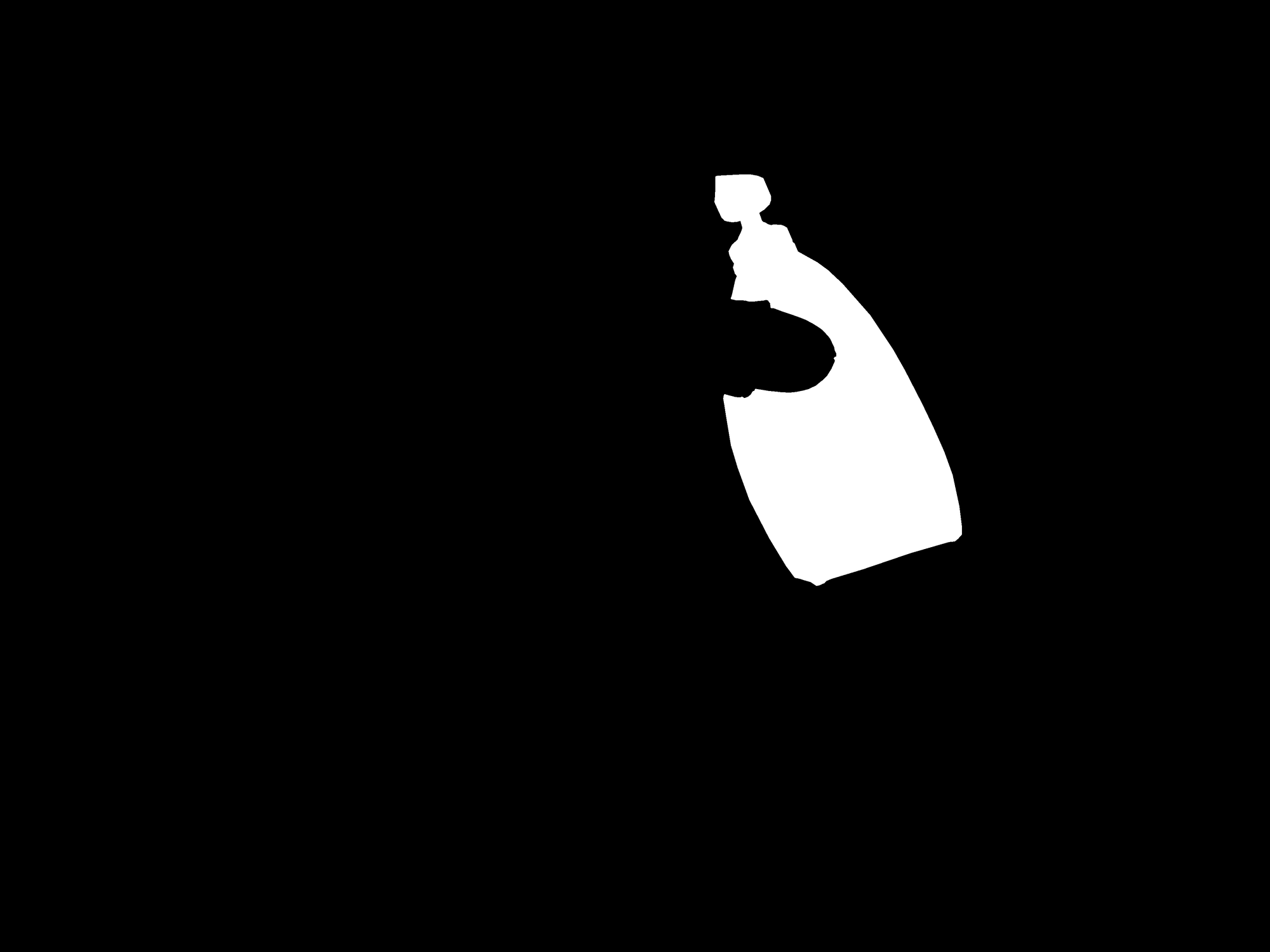} &
        \includegraphics[width=0.33\linewidth]{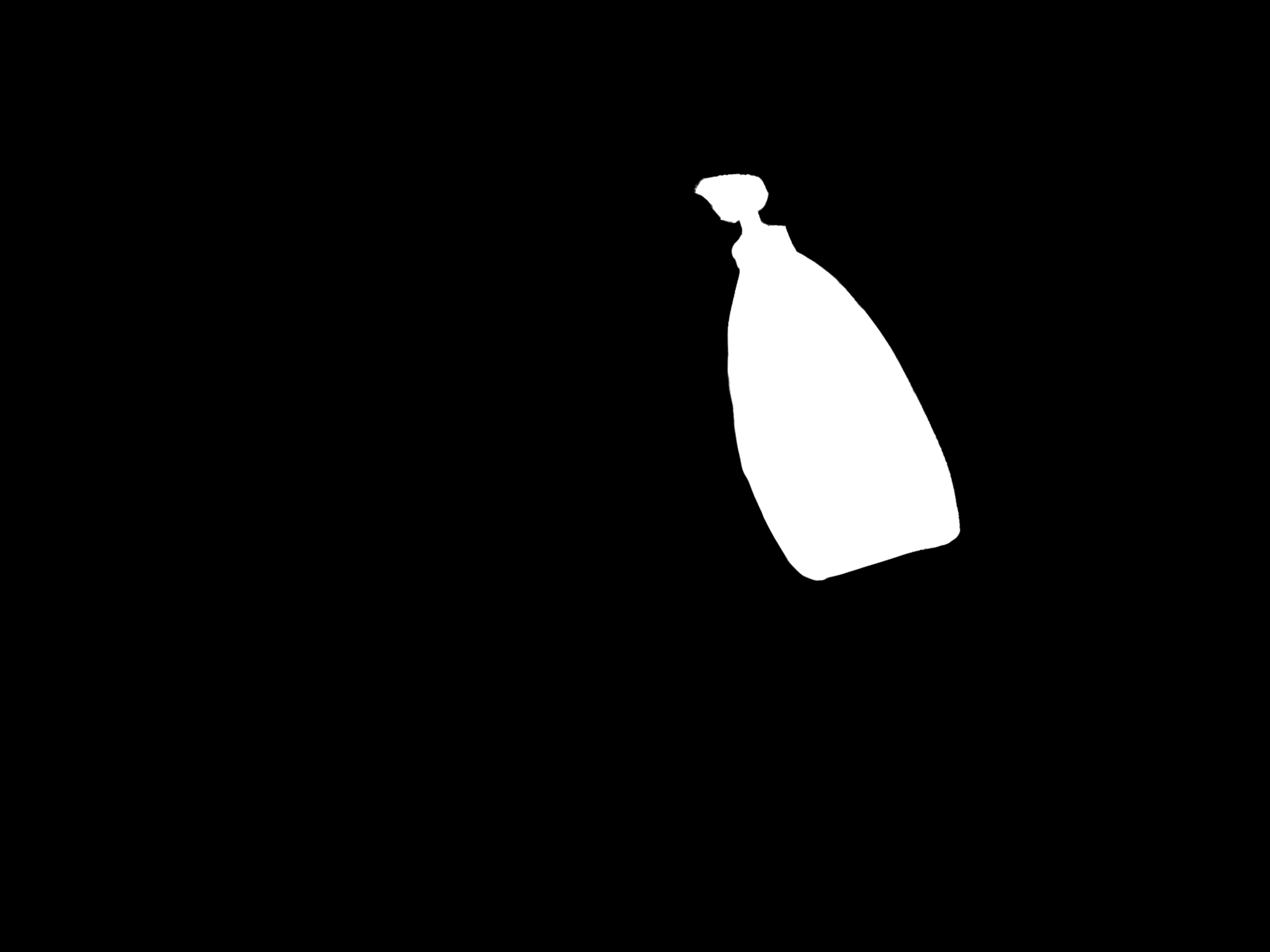} \\

        \includegraphics[width=0.33\linewidth]{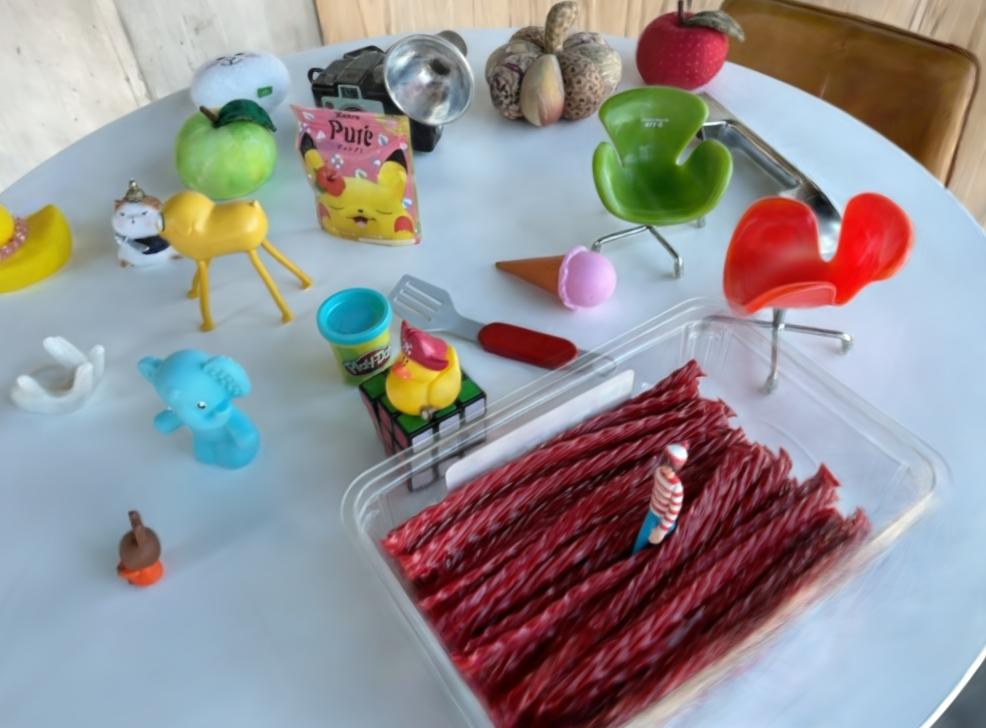} &
        \includegraphics[width=0.33\linewidth]{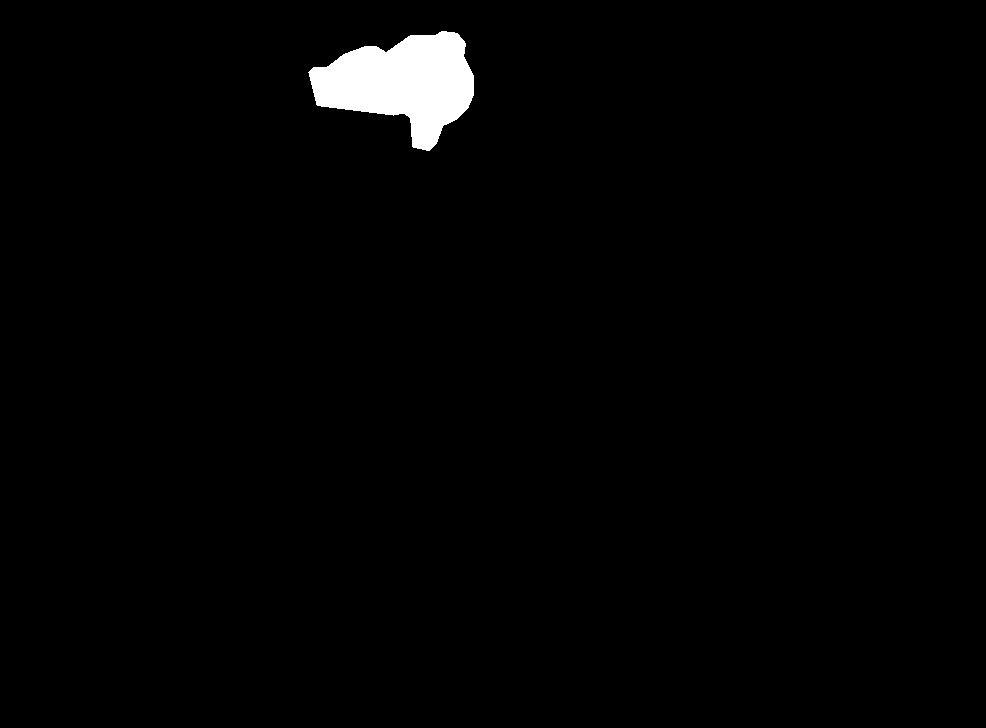} &
        \includegraphics[width=0.33\linewidth]{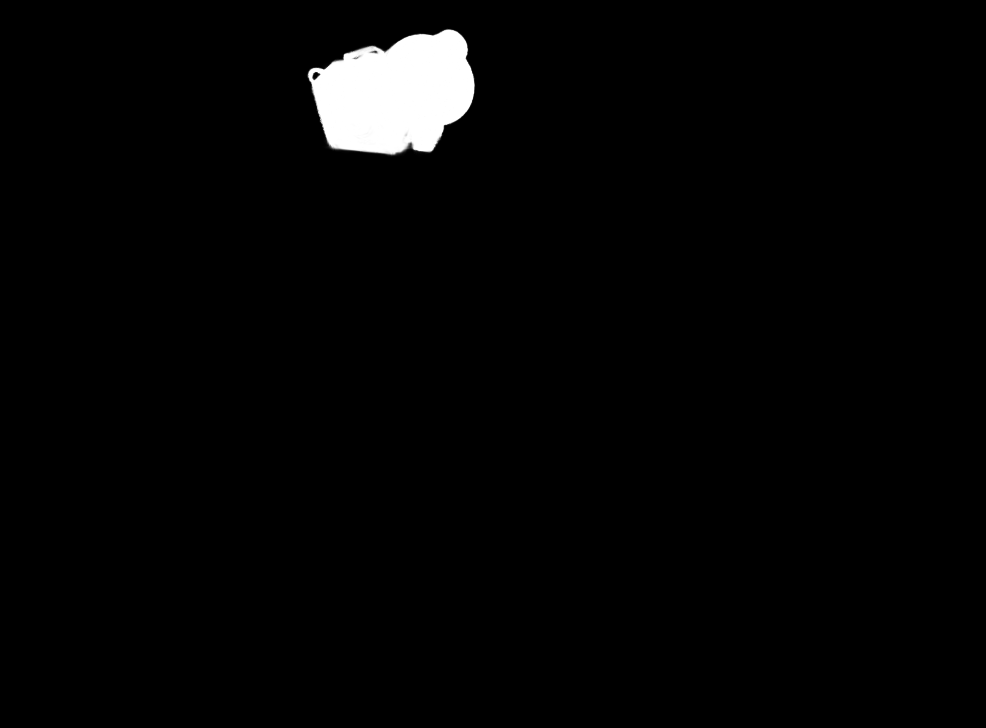} \\        
        
    \end{tabular}

    \vspace{-2mm}
    \caption{Ground Truth (GT) segmentation masks for both rendered and extracted metrics. \textit{First column:} Original RGB image of the scene. \textit{Middle column:} GT segmentation mask used for rendered metrics, \textit{Last column:} GT segmentation mask used for extracted metrics. The first row shows an example of our GT generated masks for the Mip-NeRF 360 dataset. The second row shows the rendered GT segmentation mask provided by~\cite{3DOVS_dataset_paper}, the GT extracted mask was created by us as described in Sec~\ref{supp_sec:object_masks}. The third row shows the GT rendered segmentation mask provided by~\cite{gaussian_grouping} for the LeRF dataset, the GT extracted mask was created by us as described in Sec~\ref{supp_sec:object_masks}.}
    \label{fig:masks}
\end{figure*}

\section{Occupancy and Boundary Loss Weights}
Although selected empirically, the loss weights can be explained by different gradient scaling behaviors. RGB and boundary losses accumulate gradients over many pixels per Gaussian, yielding larger magnitudes, whereas the occupancy loss is normalized by both Gaussians and sampled points, producing much smaller gradients. Hence, a larger weight ($\lambda_{\text{occ}}=10$) is needed for balance.

\section{Empty Voxels}
Our occupancy query checks not only the queried voxel but also its spatial neighborhood. Therefore, even if a voxel is marked as empty due to collisions between multiple semantic classes, sampled points are not penalized as long as valid supporting voxels exist in the local neighborhood, a visualization is shown in Fig.~\ref{fig:voxel_visual}. This tries to mitigate undesired erosion effects at object contact regions and provides robustness to artifacts introduced by voxelization. 

\section{Object Masks Annotation}
\label{supp_sec:object_masks}

In order to perform a quantitative evaluation on all datasets (Mip-NeRF 360~\cite{barron2022mipnerf360}, LLFF~\cite{mildenhall2019llff}, LeRF~\cite{lerf2023}, and 3DOVS~\cite{3DOVS_dataset_paper}) across the two metrics used in our paper, we created the necessary ground-truth segmentation masks. The rendered-view masks were kept exactly as provided by the original datasets as they already segmented the visible regions of the objects and were sufficient for the rendered metrics. The only exception is Mip-NeRF 360, for which no object masks existed; in this case, we generated both the rendered and extracted masks from scratch.
Additional masks were required only for the extracted metrics, which need occlusion free silhouettes capturing the full geometry of each object. To construct these, we first extracted the target objects using Trace3D~\cite{Trace3D} and used the resulting renders as a visual guide in Adobe Photoshop~\cite{adobephotoshop} to create the mask. The Trace3D outputs helped us determine how the masks should be filled, and all final mask boundaries were traced and refined manually. LLFF~\cite{mildenhall2019llff} contains no significant occlusions, so the NVOS~\cite{nvos_ren_cvpr2022} original masks were used for both rendered and extracted metrics. For LeRF~\cite{lerf2023}, we used the rendered masks provided by Gaussian Grouping~\cite{gaussian_grouping} and manually created only the additional extracted masks for occluded test views, guided by Trace3D when necessary. The same procedure was followed for 3DOVS~\cite{3DOVS_dataset_paper}, reusing the authors’ rendered masks and manually generating the extracted ones where needed. For Mip-NeRF 360, both rendered and extracted masks were manually created in Photoshop, again using Trace3D extractions as a reference. Examples of both rendered and extracted masks are shown in Fig.~\ref{fig:masks}.

\section{Implementation Details}

The 2D Boundary Loss is implemented by modifying both the \textit{forward.cu} and \textit{backward.cu} CUDA kernels in the original rasterizer. In contrast, the 3D Boundary Loss is implemented entirely in PyTorch~\cite{paszke2019pytorchimperativestylehighperformance}.
The 3K training follows the same learning-rate schedule used in the original 2DGS~\cite{Huang2DGS2024} between iterations 27K and 30K. For training, LeRF~\cite{lerf2023} was used at its original resolution, Mip-NeRF 360~\cite{barron2022mipnerf360} was downsampled by a factor of 4, and both 3DOVS~\cite{3DOVS_dataset_paper} and LLFF~\cite{mildenhall2019llff} were resized to 1.6k pixels width following the standard 3DGS~\cite{kerbl3Dgaussians} preprocessing procedure. All baselines were optimized using the same input data and the same initial masks for all scenes. Each baseline was optimized using its original training schedule. All methods already included their own mechanism for selecting Gaussians in 3D, so we used their implementations directly without introducing any additional components or modifications specific to our evaluation. We computed all metric means by first averaging the scores over all masks within each scene, and then averaging these scene-level means across the entire dataset.

\section{Computational Complexity Discussion}

Given the heterogeneous backbones used across the methods we compare to, a direct runtime comparison is not straightforward. We therefore evaluate boundary refinement methods in terms of the additional iterations they require. ObjectGS~\cite{ObjectGS} is the most efficient, performing boundary refinement within the original $30000$ training iterations. Trace3D~\cite{Trace3D} introduces $9000$ additional refinement iterations. COB-GS~\cite{COBGS}, however, trains a separate model for each object instance, causing the number of iterations to scale with both the number of images and the number of classes, making it inefficient for multi-class scenes. For example, in the Figurines scene from LeRF (300 images, 7 classes), COB-GS requires $300 \times 14 \times 7 = 29400$ iterations, compared to $9000$ for Trace3D and $3000$ for our method, which remains independent of the number of classes.

%Given the heterogeneous backbones of all methods, a direct runtime comparison is not straightforward. 
%We therefore analyze for the boundary modifying methods the \textit{extra iterations} required for boundary refinement. ObjectGS~\cite{ObjectGS} is the most efficient, refining boundaries within the original 30K training iterations. Trace3D~\cite{Trace3D} adds 9K refinement steps. COB-GS~\cite{COBGS}, instead, trains per object instance, with iterations scaling with image and class counts, inefficient for multi-class scenes. 
%In the \textit{Figurines} scene of LeRF (300 images, 7 classes), it requires $300 \times 14 \times 7 = 29{,}400$ iterations, compared to 9K for Trace3D and 3K for ours, independent of class count.

\section{Additional Qualitative Results}

%Due to limited space in the original paper we could not include as many result images as we would have liked. We now show several extracted objects obtained by our method in Fig.~\ref{fig:additional_qualitative1}, Fig.~\ref{fig:additional_qualitative2}, and in Fig.~\ref{fig:additional_qualitative3}. We show for each scene 3 different rendered viewpoints. It can be seen that our approach manages to produce high quality boundary segmentations. We show examples from all four datasets (Mip-NeRF 360~\cite{barron2022mipnerf360}, LLFF~\cite{mildenhall2019llff}, LeRF~\cite{lerf2023}, and 3DOVS~\cite{3DOVS_dataset_paper}). We remove the background for all scenes and show the remaining object classes, several for LeRF and 3DOVS, and a single object on Mip-NeRF 360 and LLFF. The obtained results show that our approach manages to obtain fine grained details, like the T-Rex ribcage or the lego bonsai leafs, that are complex to understand from 2D segmentations. The results also showcase that non-visible Gaussians are no longer an issue when performing object extraction. We show additional qualitative comparisons and results in the video that is provided alongside this supplemental material.

\begin{figure*}[t]
    \centering
    \setlength{\tabcolsep}{1pt}
    \renewcommand{\arraystretch}{0.95}

    \begin{tabular}{ccc}
        \includegraphics[width=0.32\linewidth]{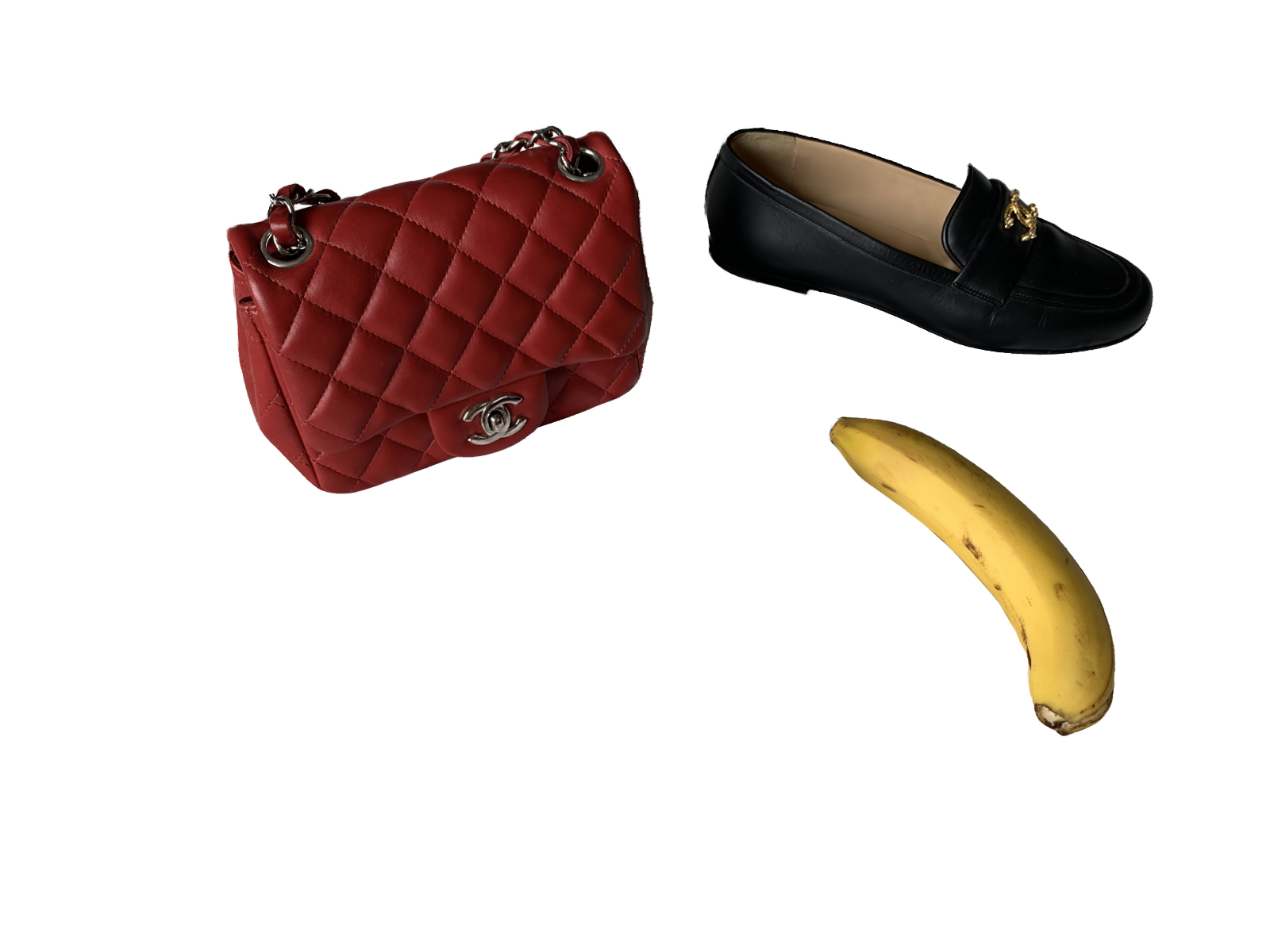} &
        \includegraphics[width=0.32\linewidth]{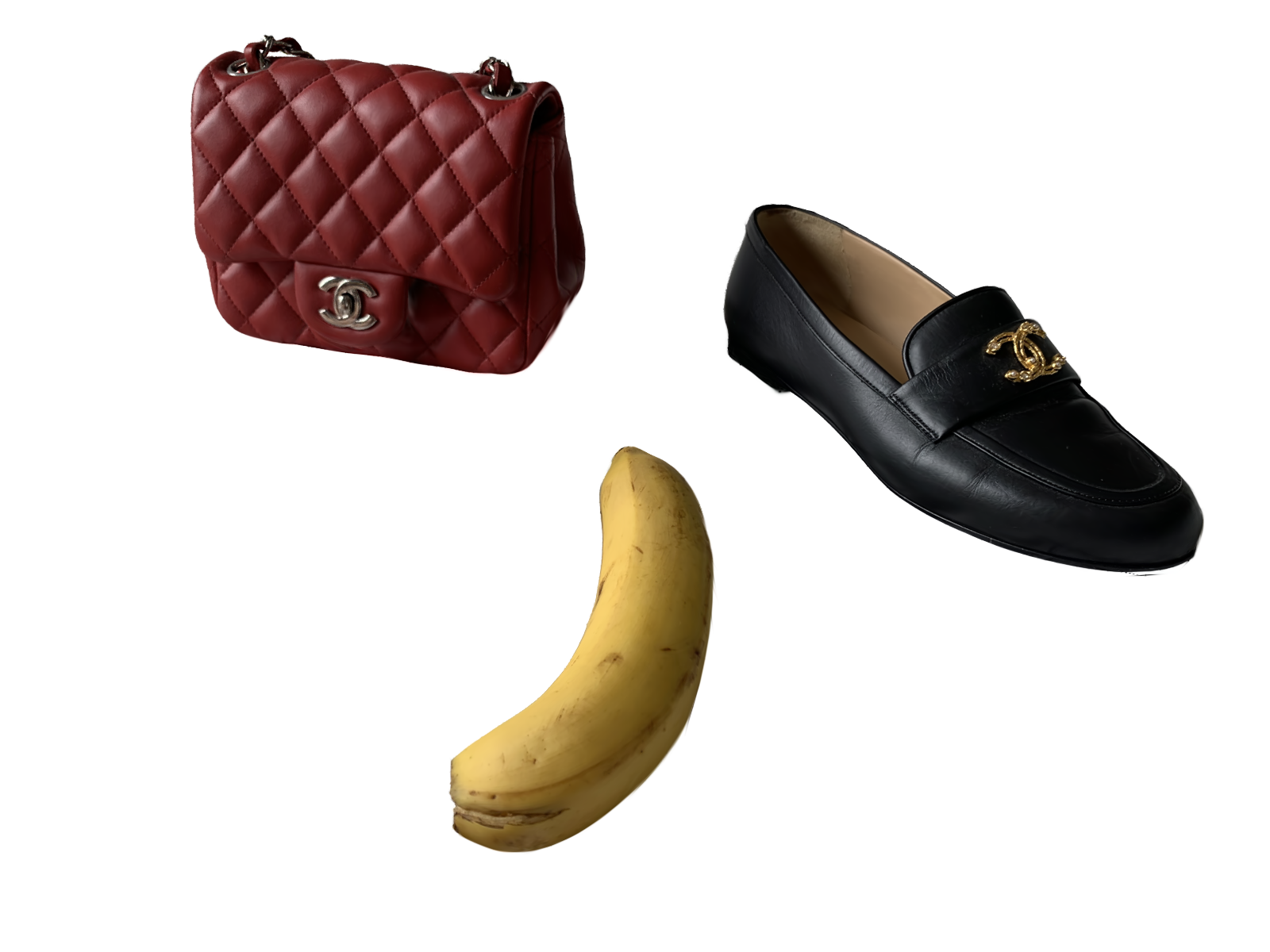} &
        \includegraphics[width=0.32\linewidth]{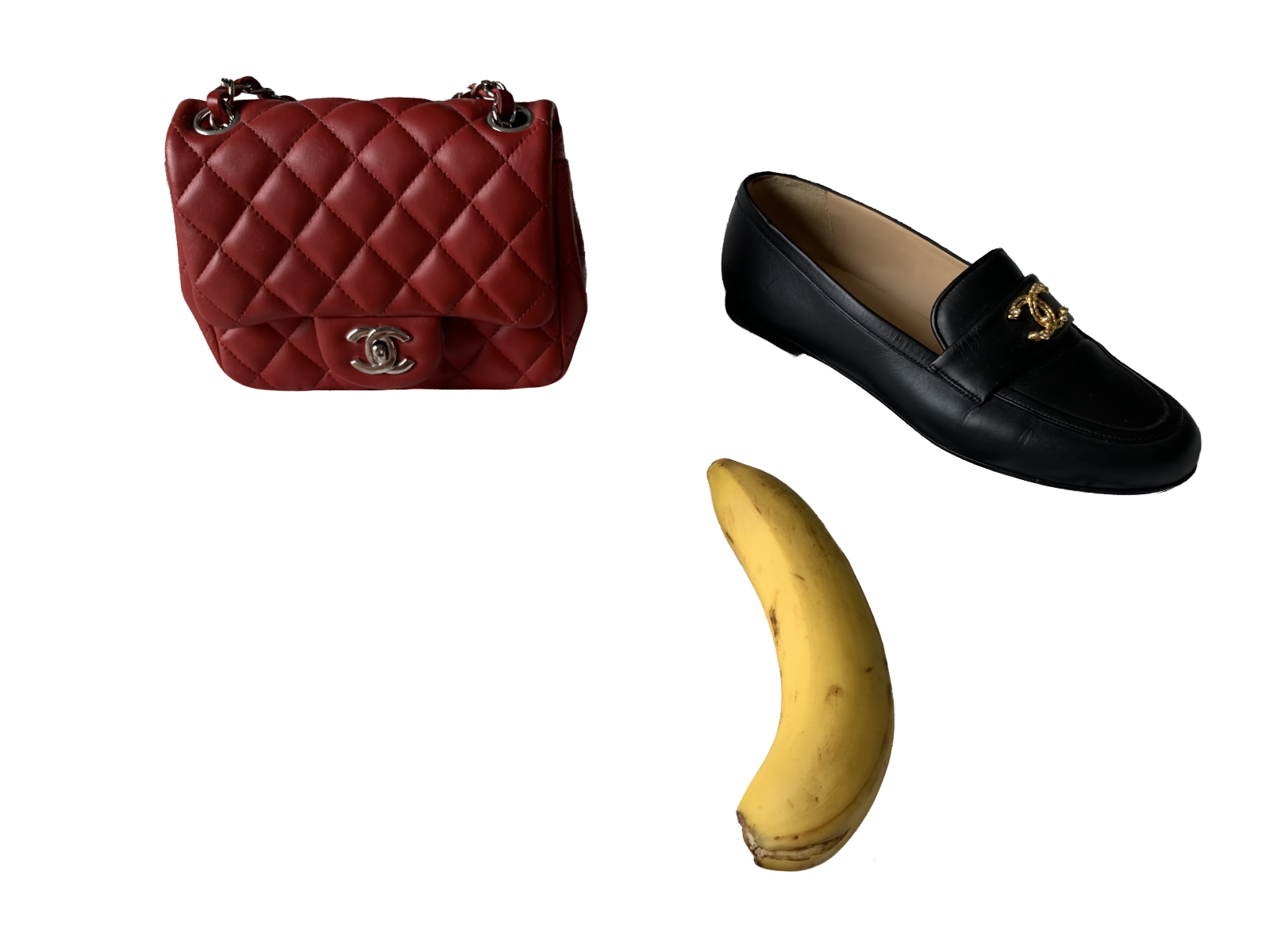} \\

        \includegraphics[width=0.32\linewidth]{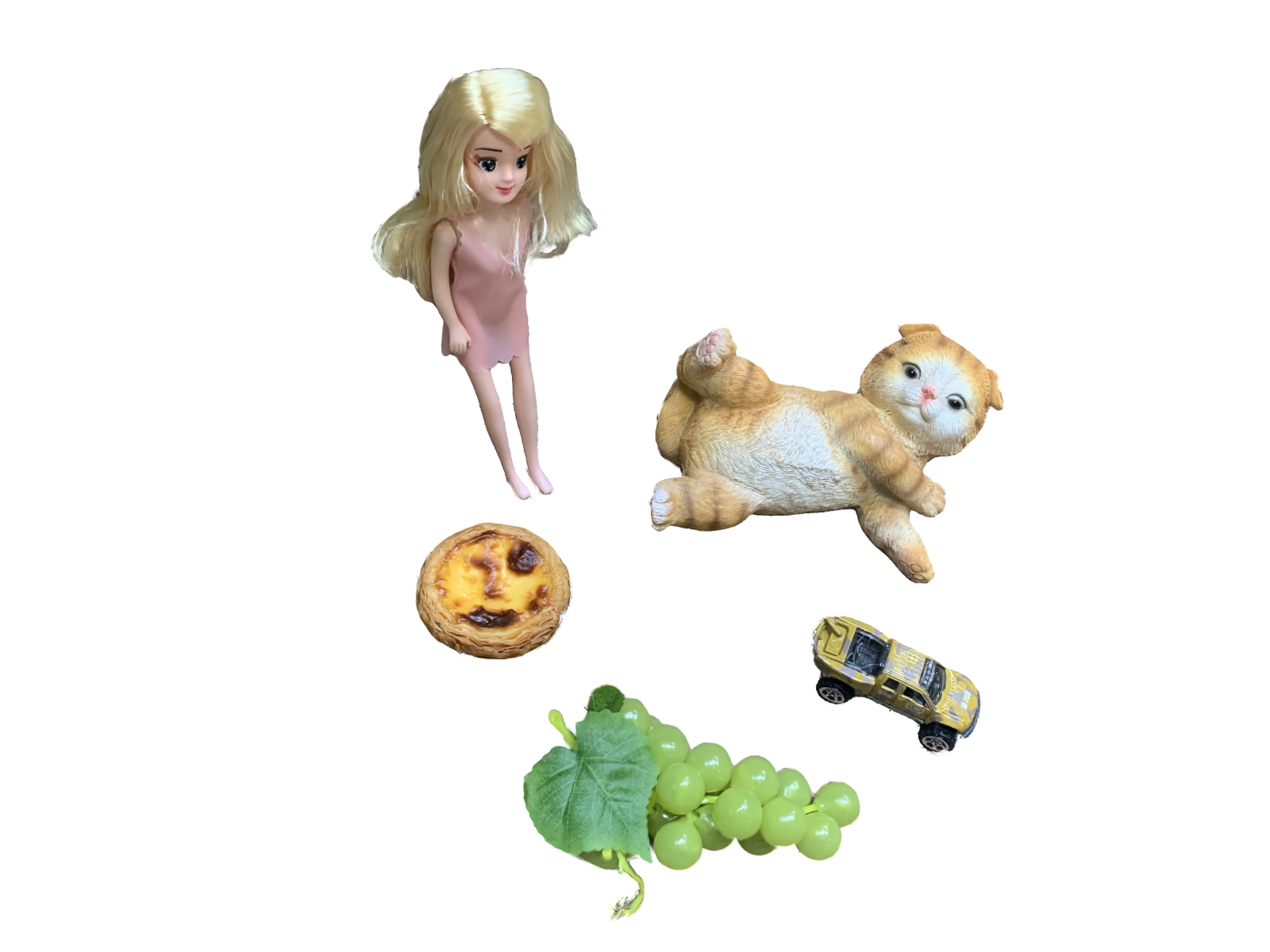} &
        \includegraphics[width=0.32\linewidth]{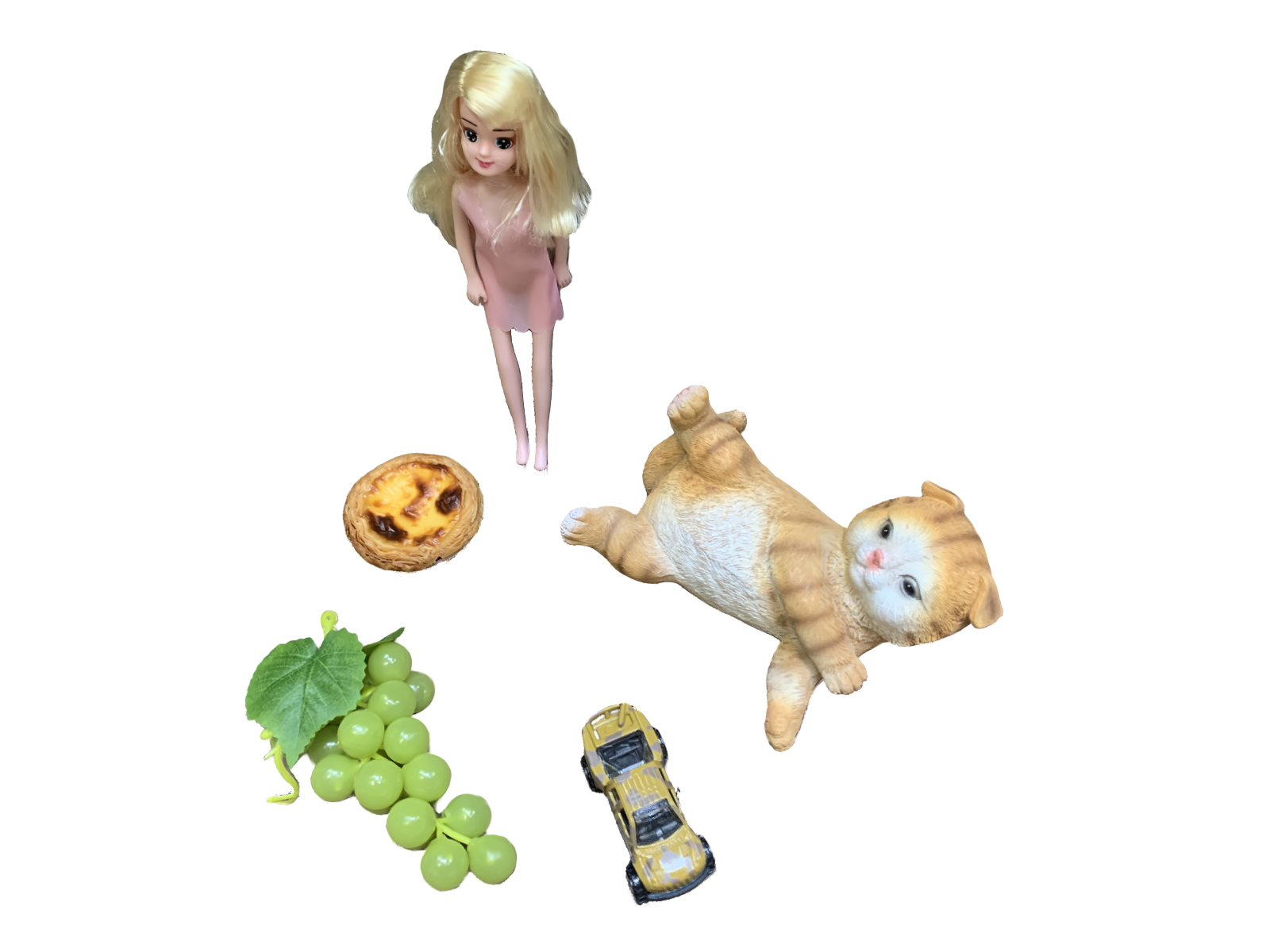} &
        \includegraphics[width=0.32\linewidth]{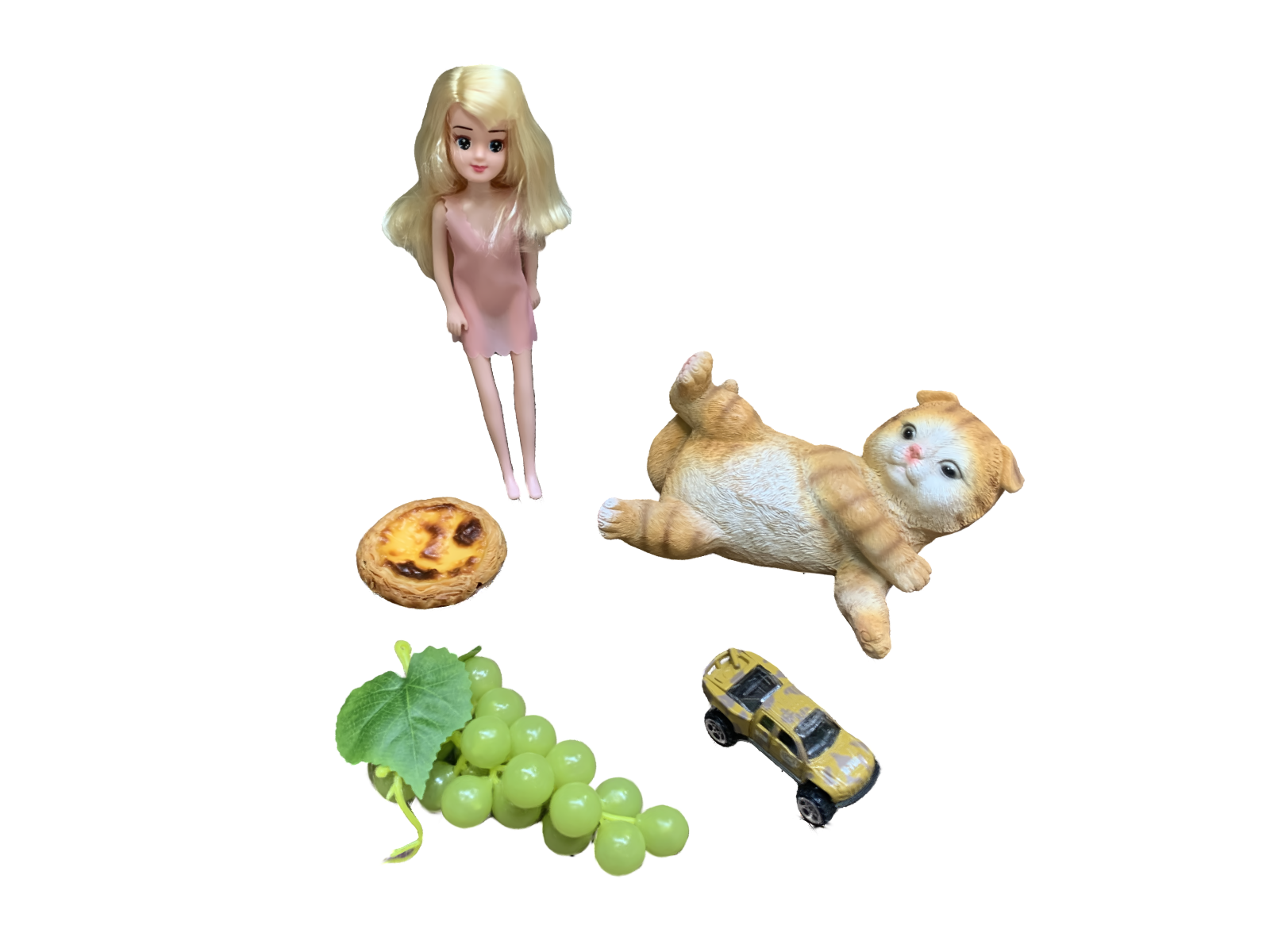} \\

        \includegraphics[width=0.32\linewidth]{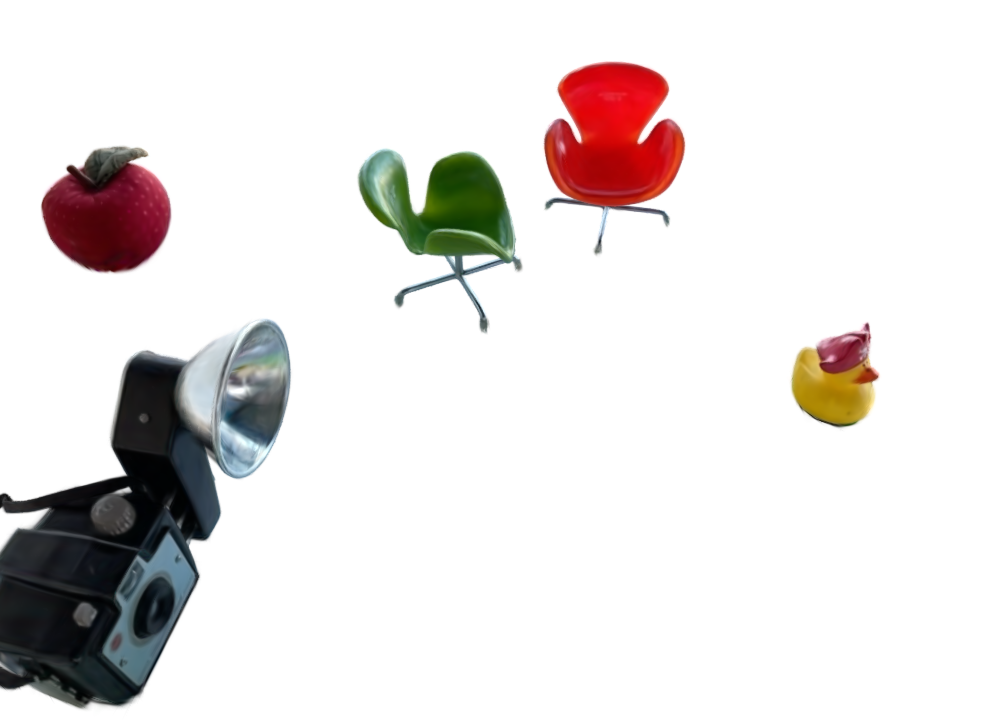} &
        \includegraphics[width=0.32\linewidth]{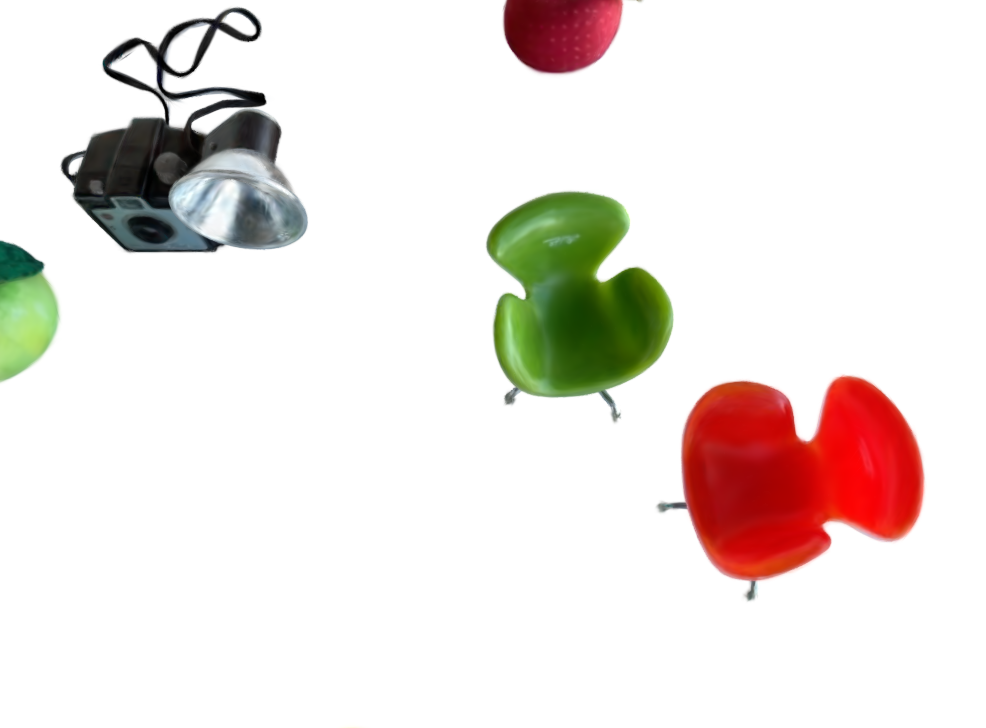} &
        \includegraphics[width=0.32\linewidth]{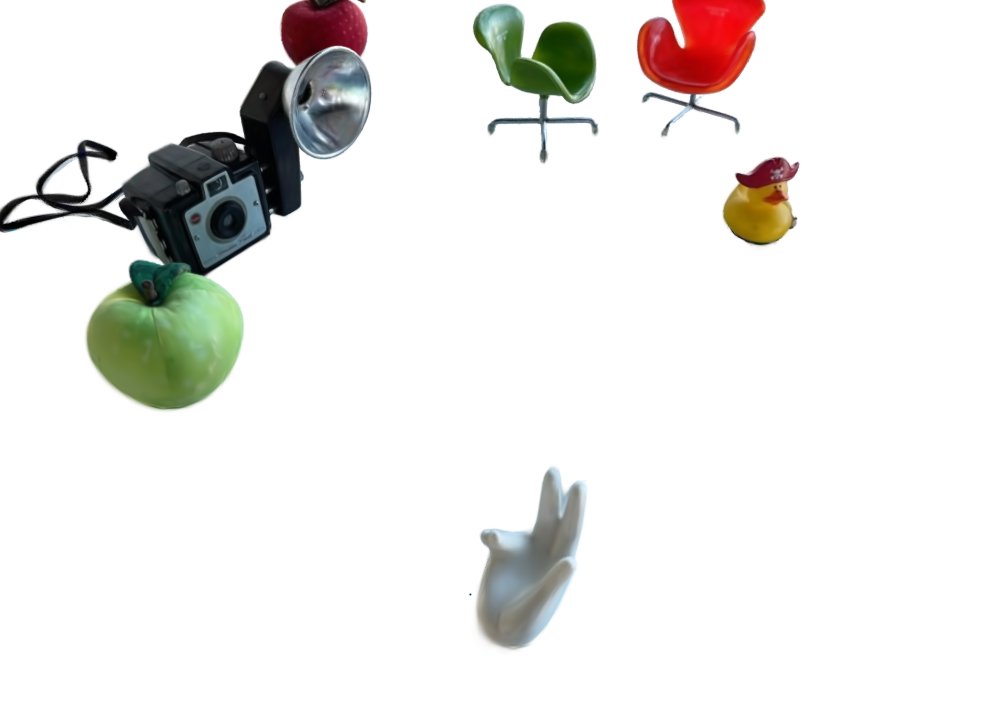} \\        
        
    \end{tabular}

    \vspace{-3mm}
    \caption{Additional qualitative results obtained using our proposed approach}
    \label{fig:additional_qualitative1}
\end{figure*}

\begin{figure*}[t]
    \centering
    \setlength{\tabcolsep}{1pt}
    \renewcommand{\arraystretch}{0.95}

    \begin{tabular}{ccc}

        \includegraphics[width=0.32\linewidth]{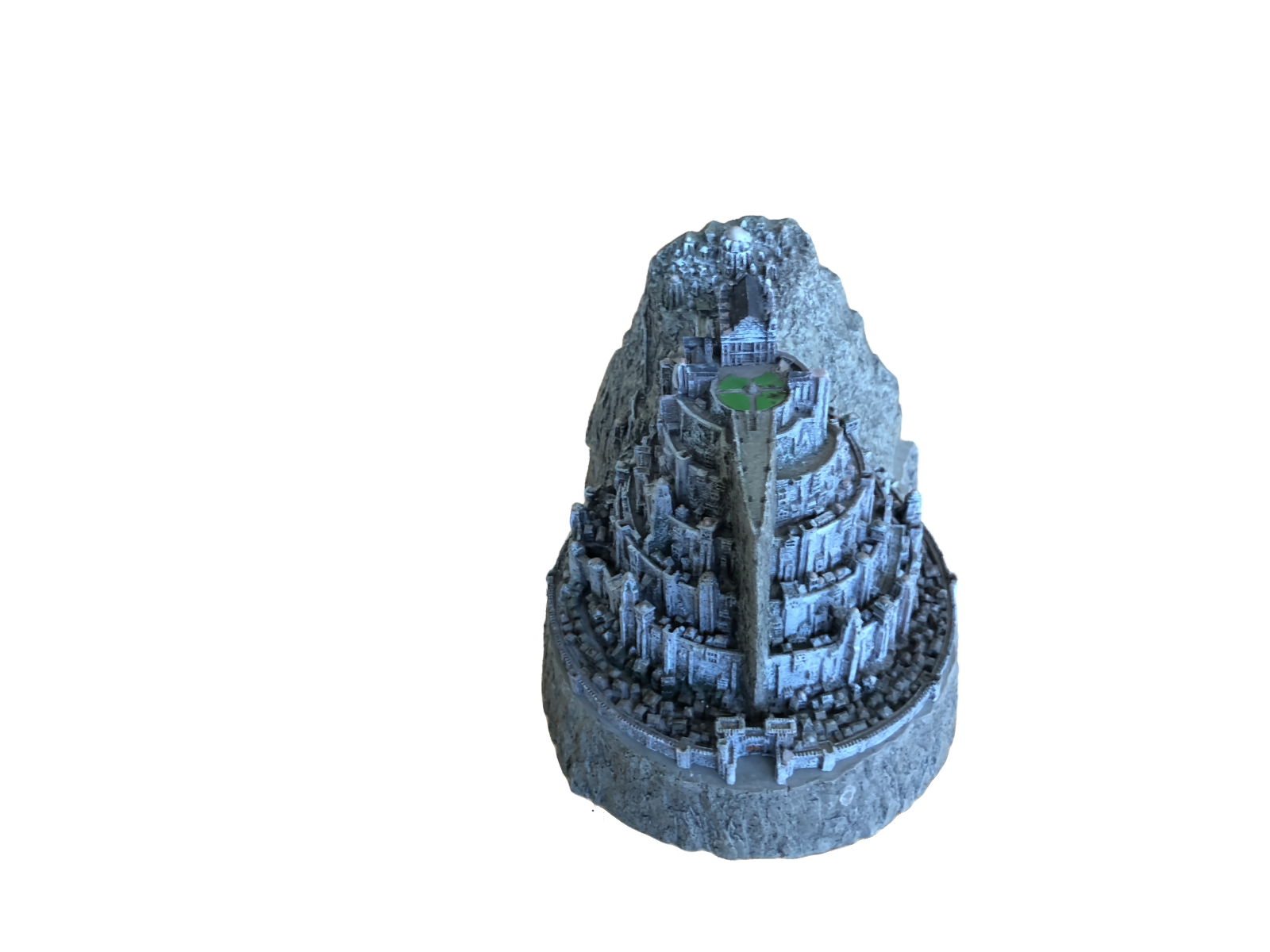} &
        \includegraphics[width=0.32\linewidth]{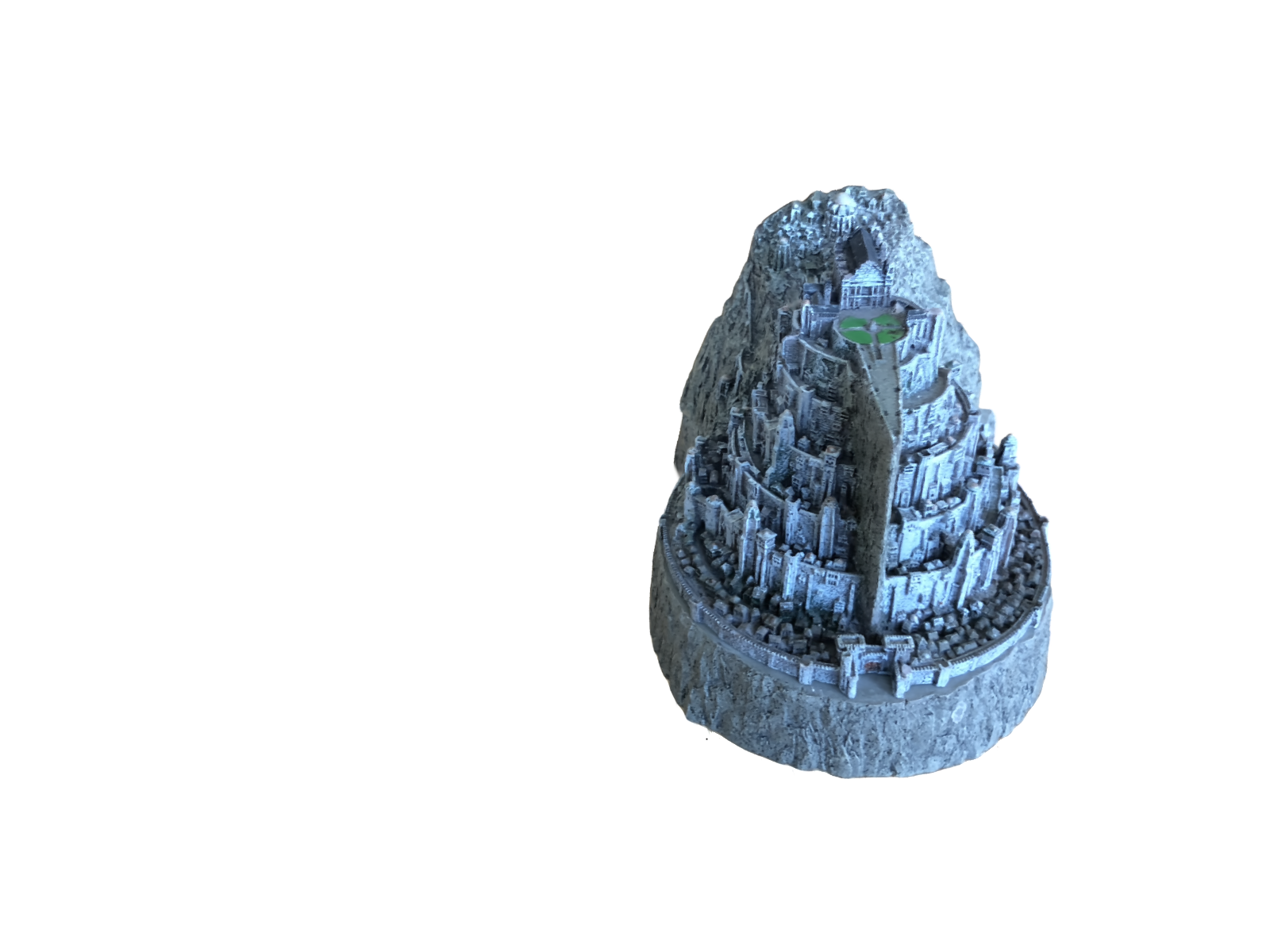} &
        \includegraphics[width=0.32\linewidth]{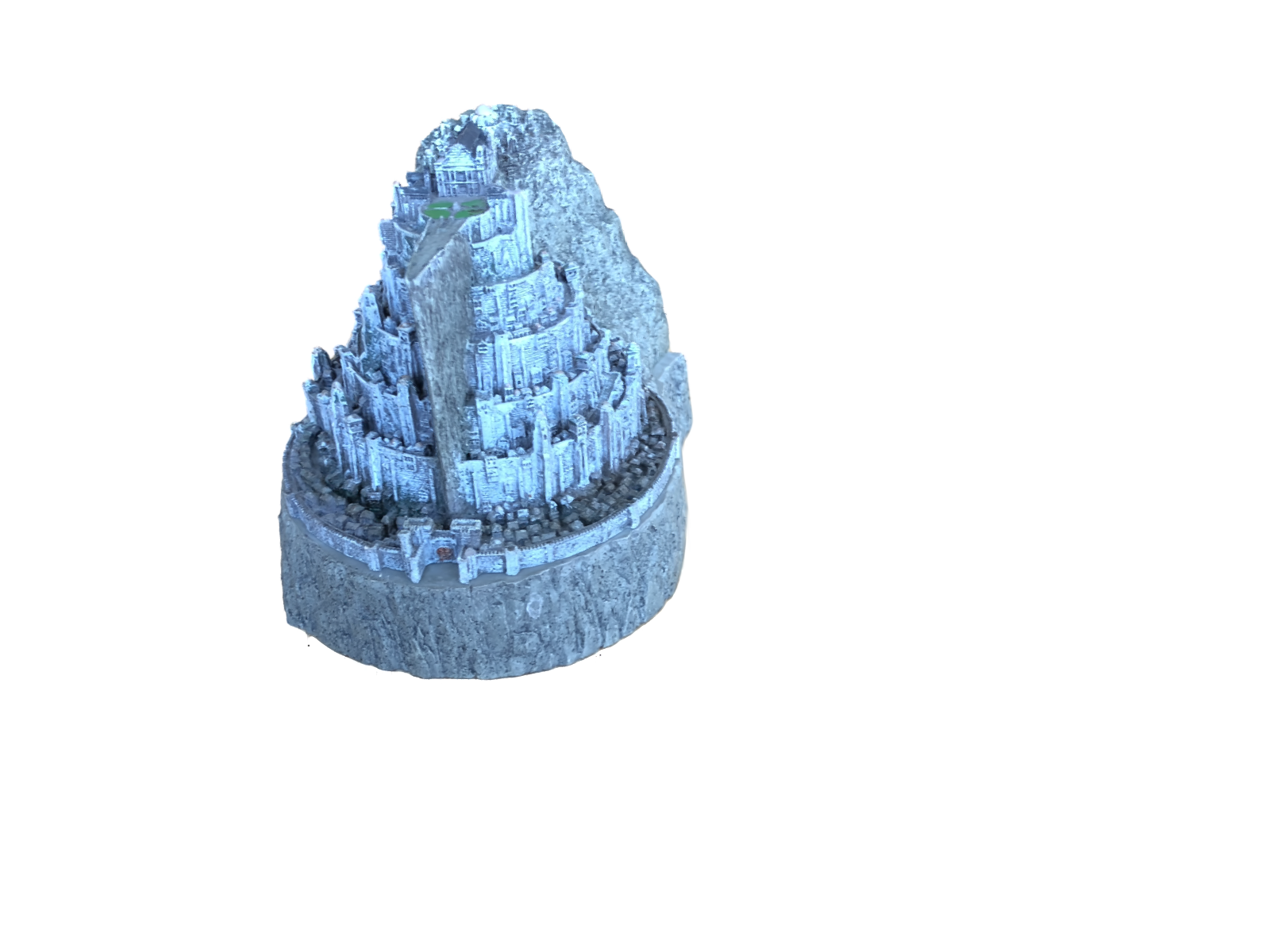} \\

        \includegraphics[width=0.32\linewidth]{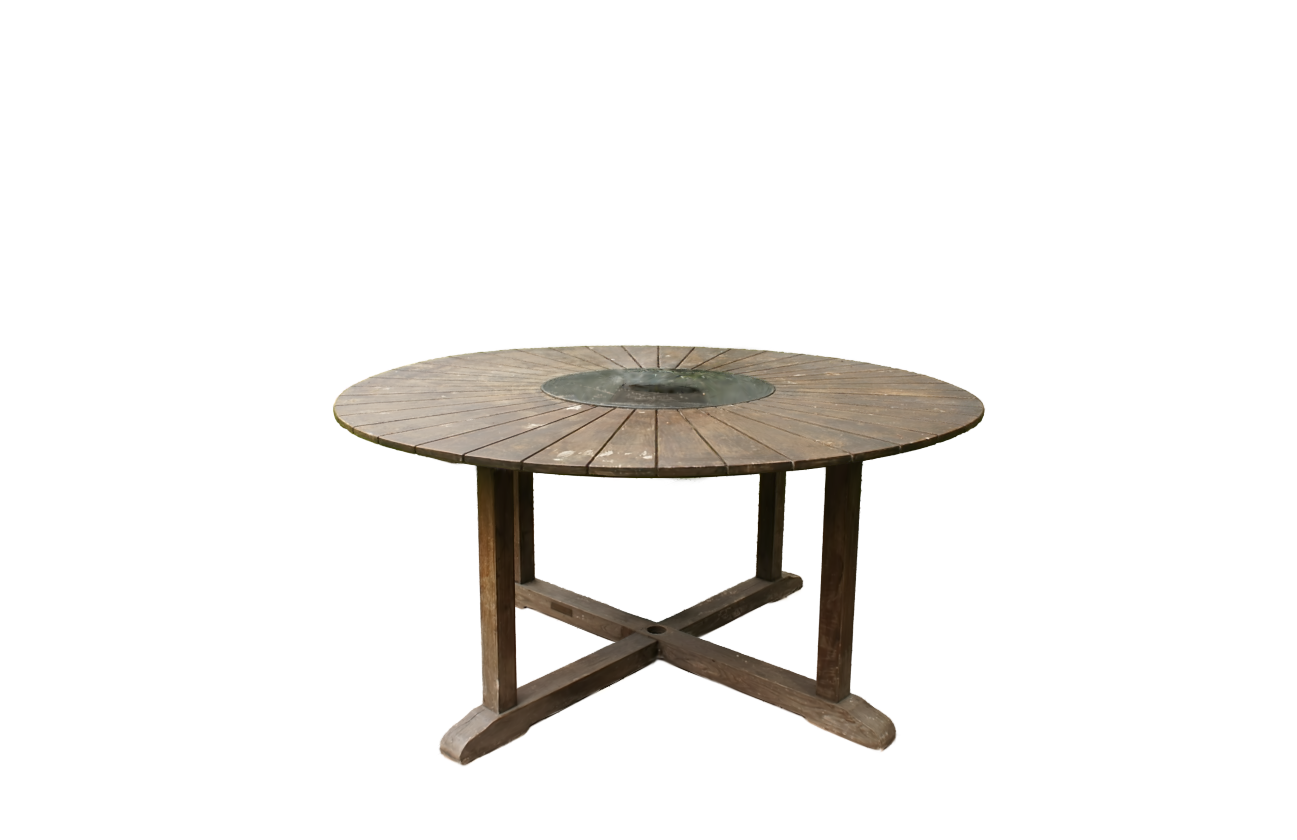} &
        \includegraphics[width=0.32\linewidth]{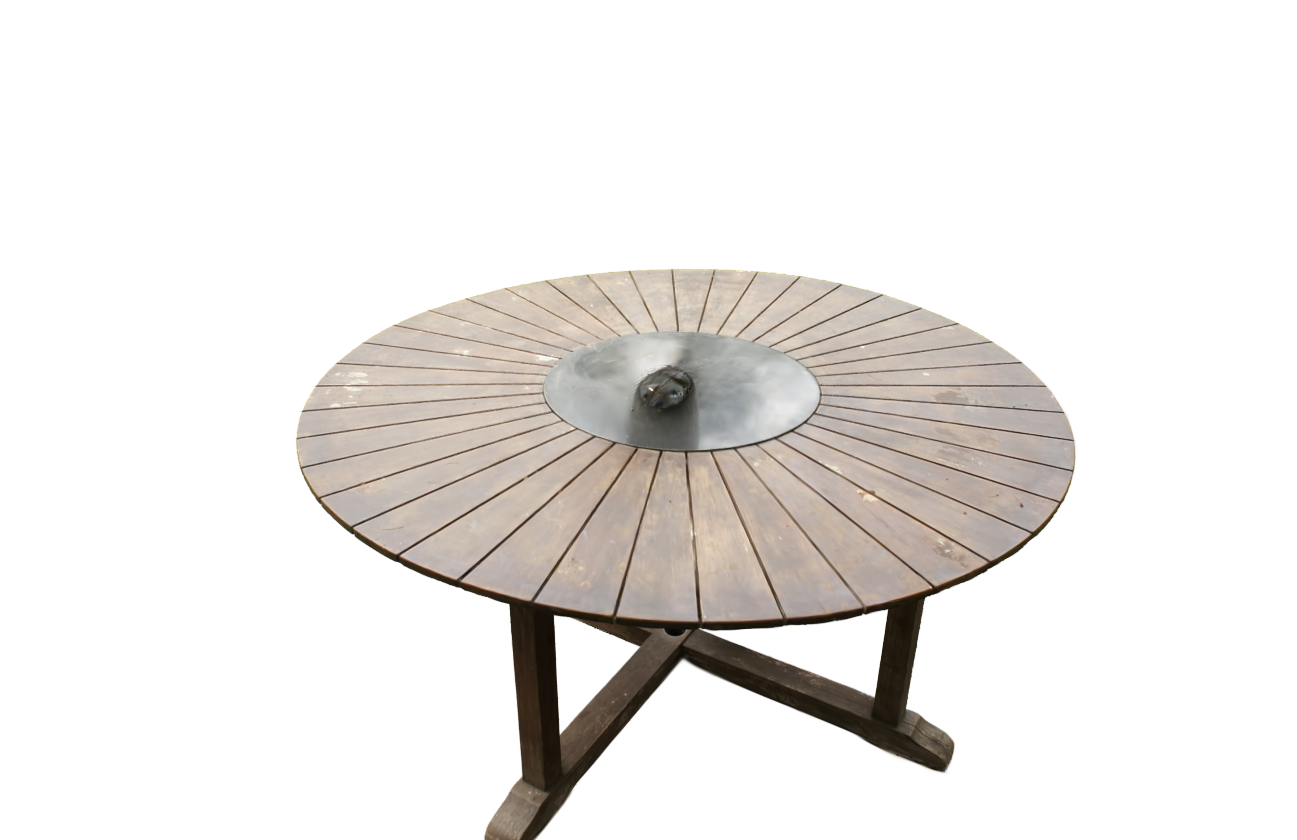} &
        \includegraphics[width=0.32\linewidth]{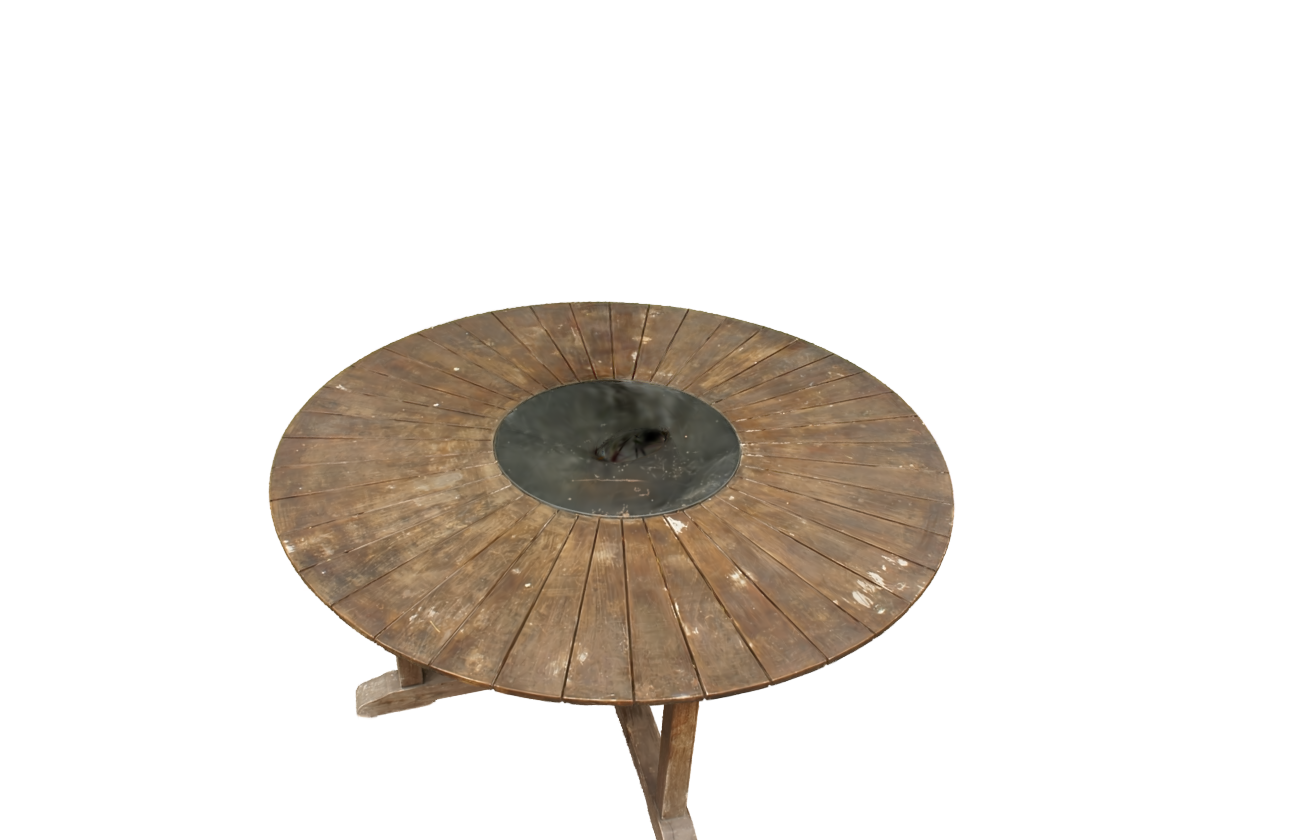} \\

        \includegraphics[width=0.32\linewidth]{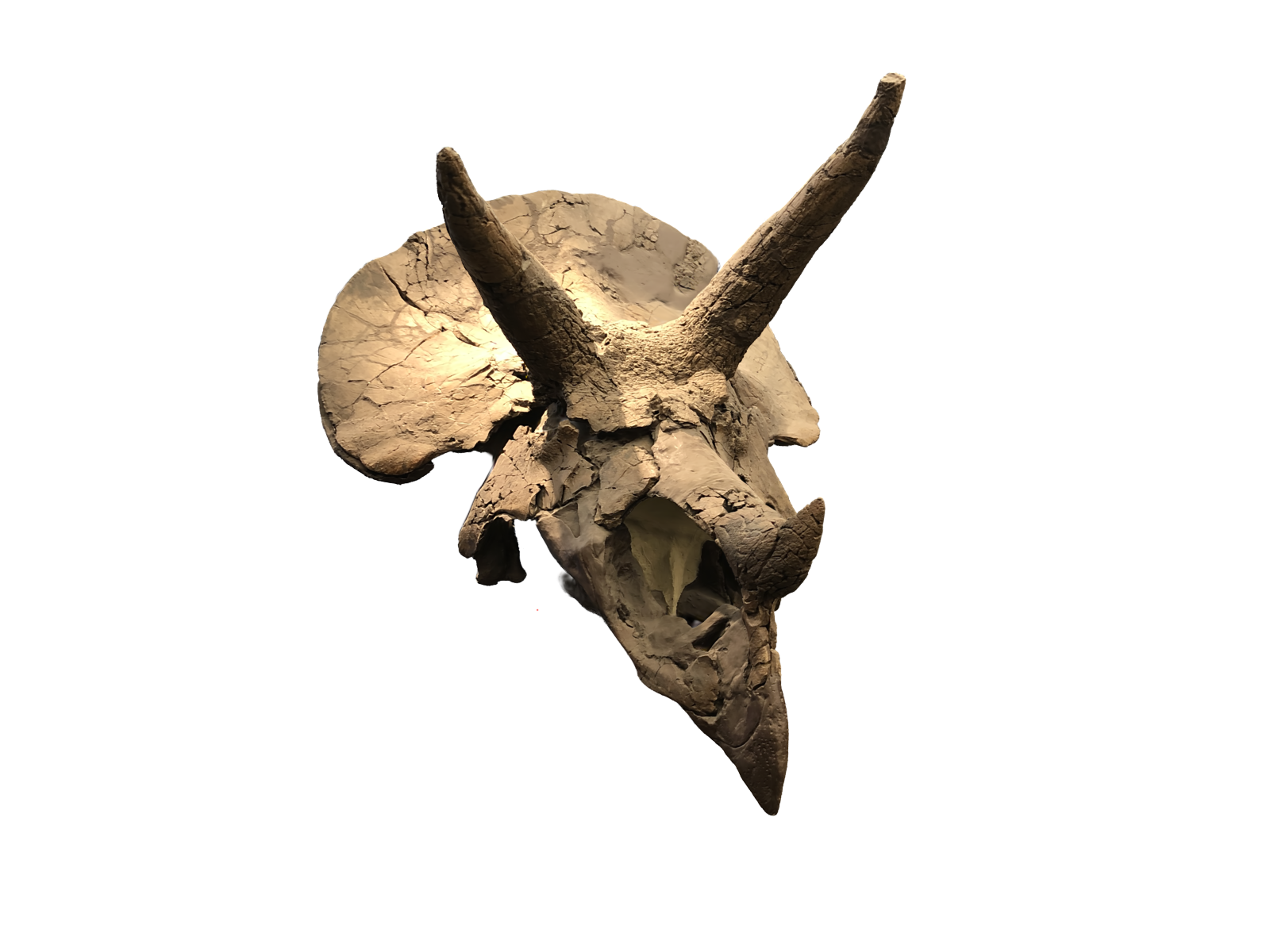} &
        \includegraphics[width=0.32\linewidth]{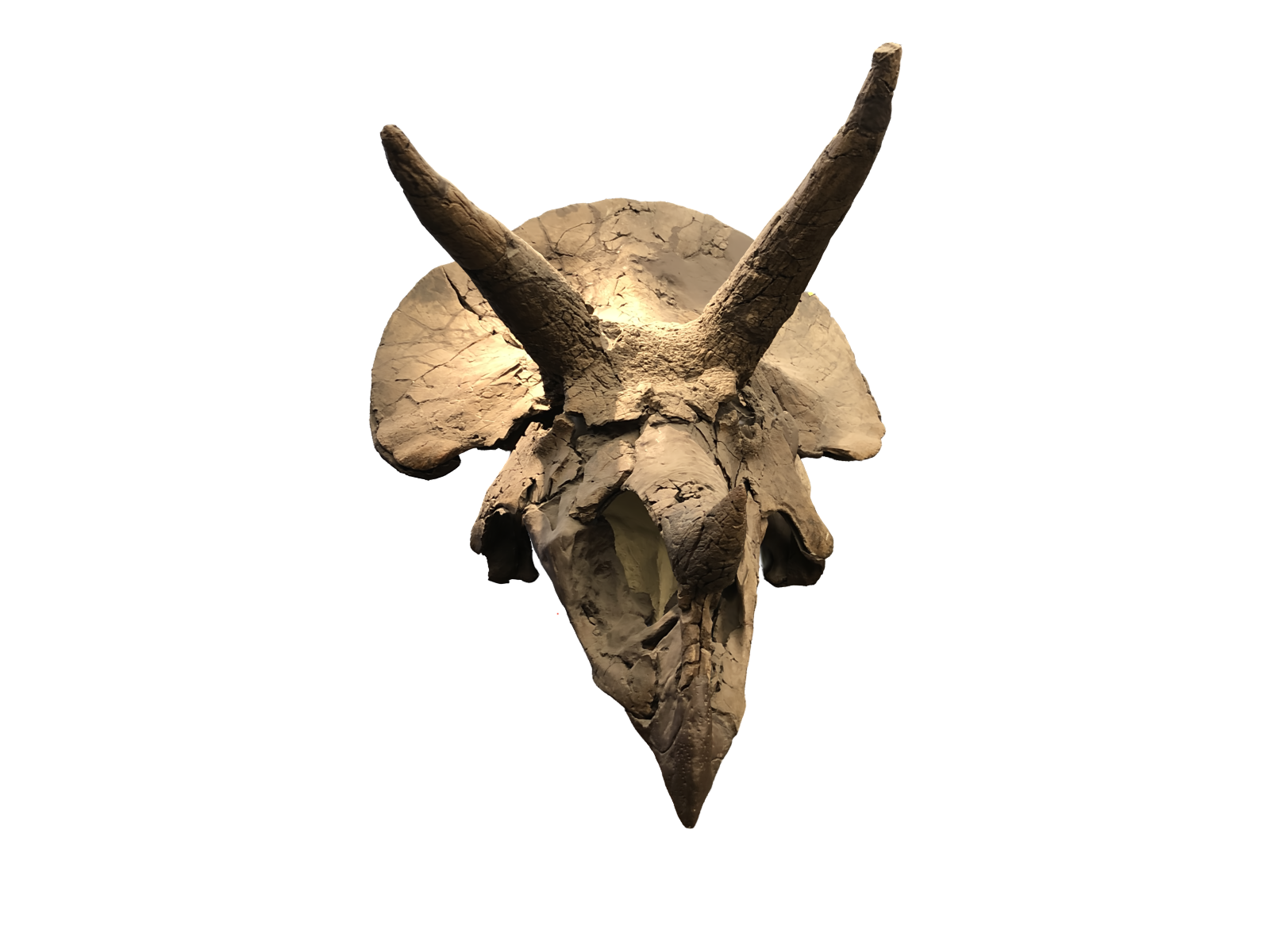} &
        \includegraphics[width=0.32\linewidth]{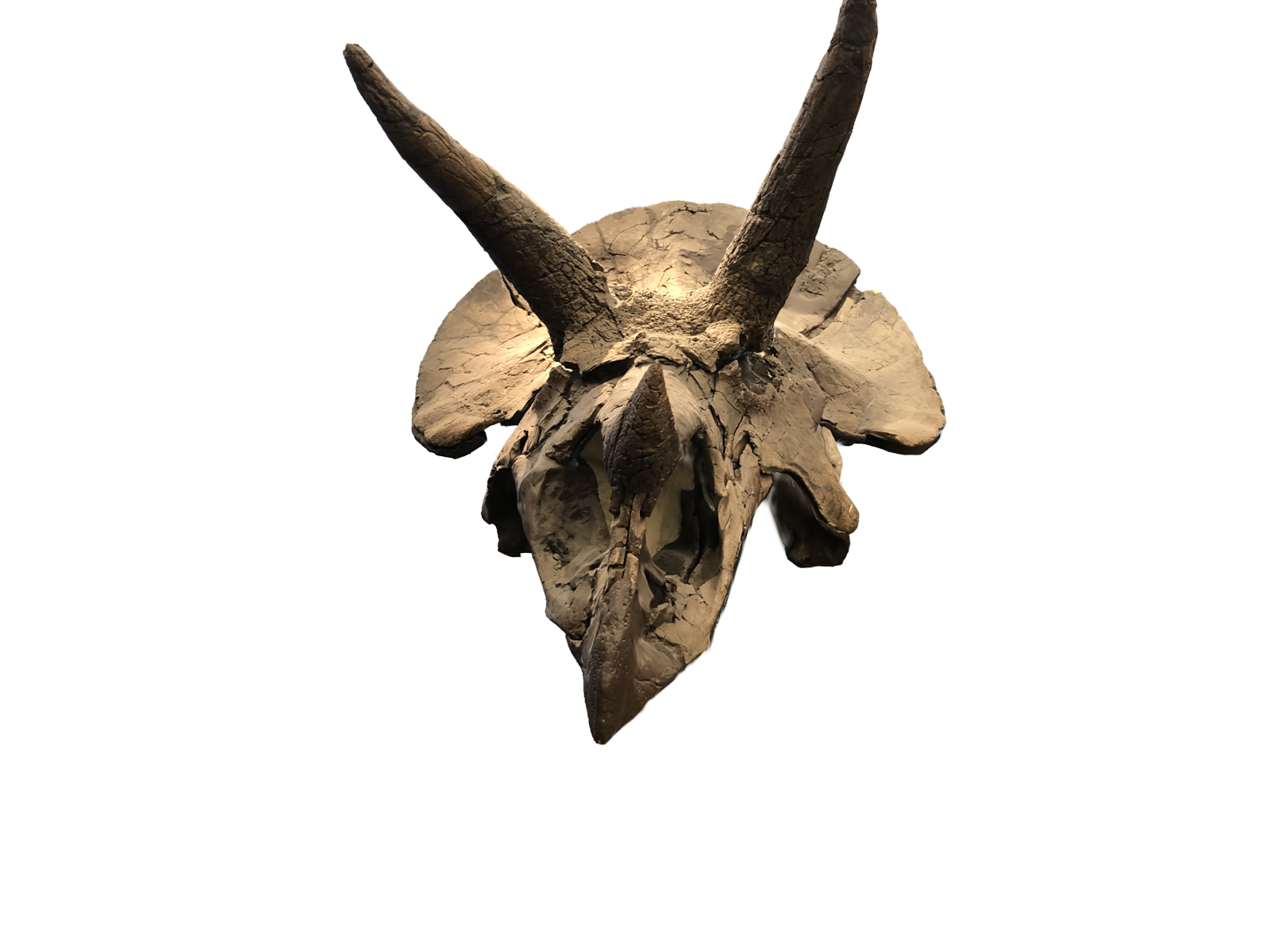} \\

        \includegraphics[width=0.32\linewidth]{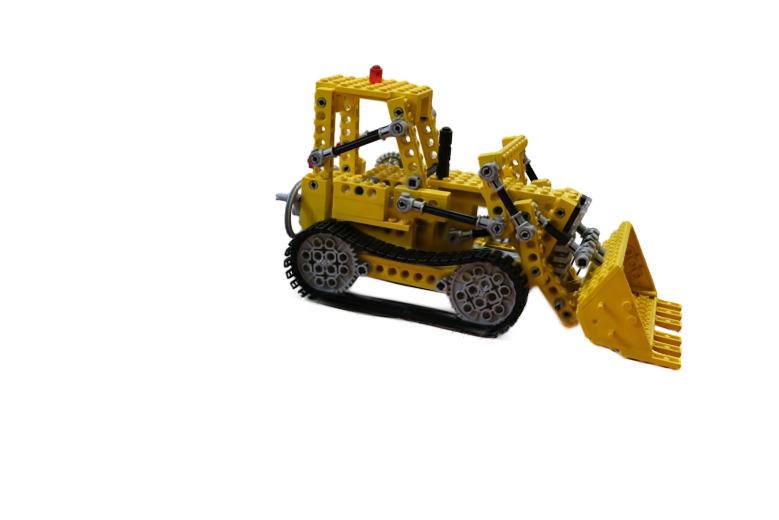} &
        \includegraphics[width=0.32\linewidth]{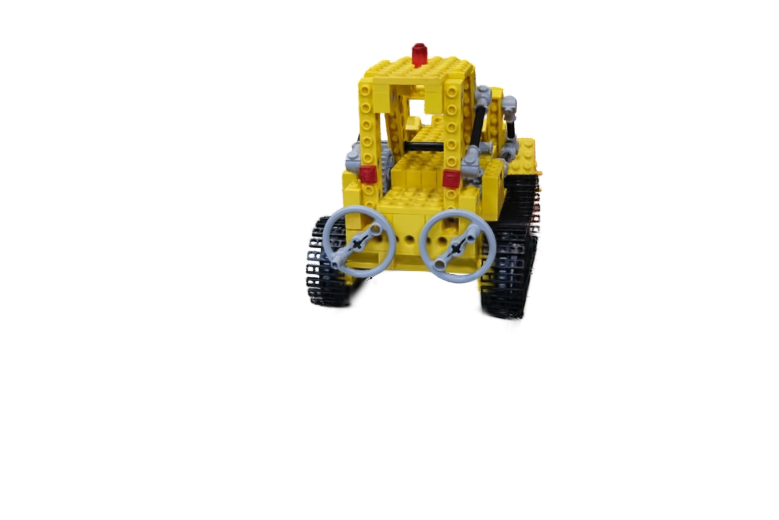} &
        \includegraphics[width=0.32\linewidth]{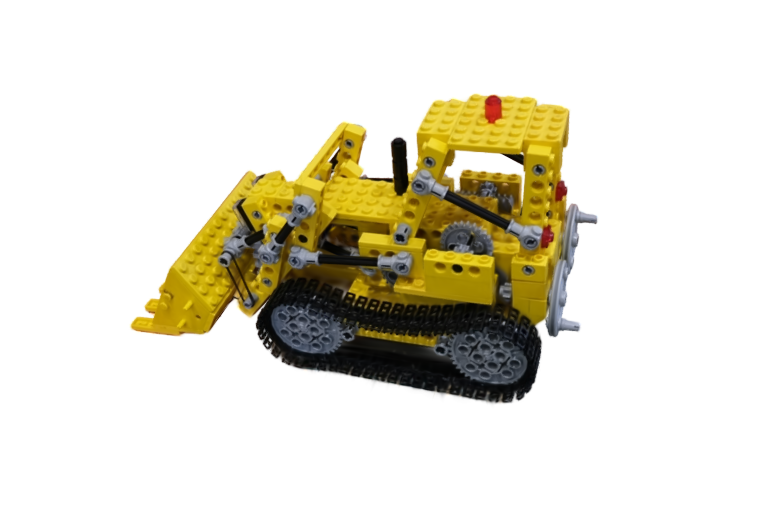} \\

        \includegraphics[width=0.32\linewidth]{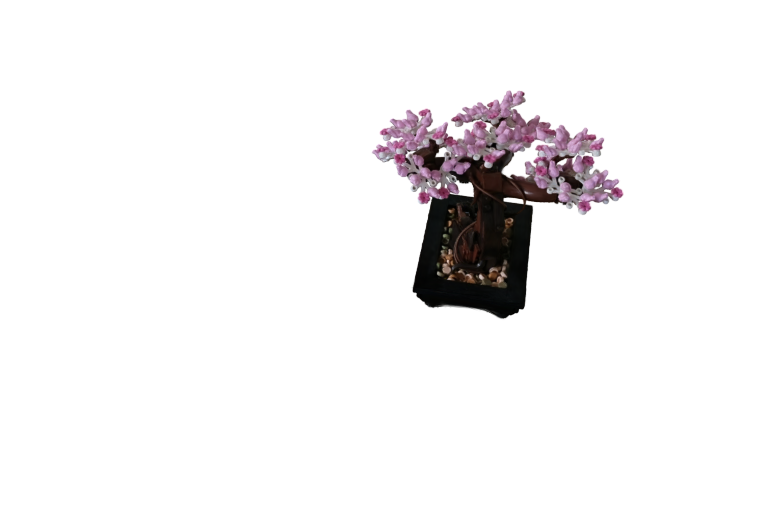} &
        \includegraphics[width=0.32\linewidth]{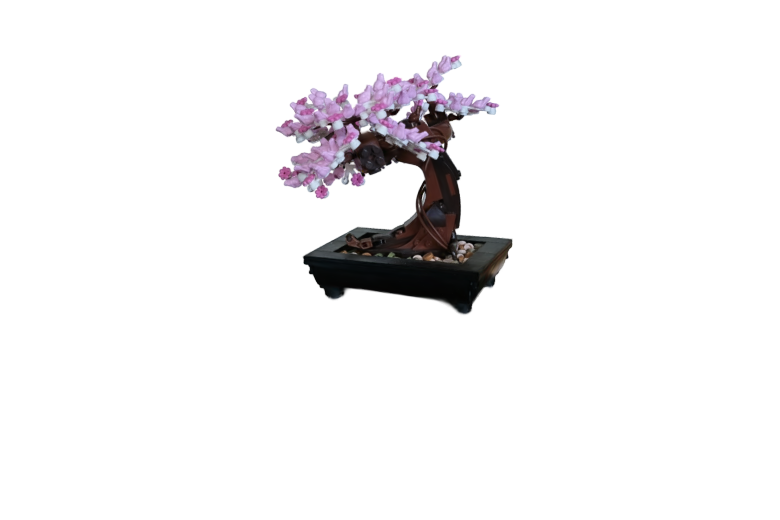} &
        \includegraphics[width=0.32\linewidth]{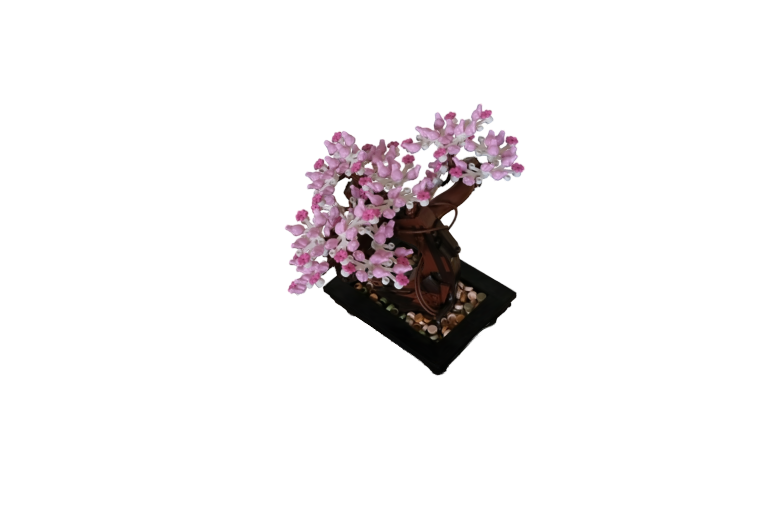} \\
        
    \end{tabular}

    \vspace{-2mm}
    \caption{Additional qualitative results obtained using our proposed approach}
    \label{fig:additional_qualitative2}
\end{figure*}

\begin{figure*}[t]
    \centering
    \setlength{\tabcolsep}{1pt}
    \renewcommand{\arraystretch}{0.95}

    \begin{tabular}{ccc}

        \includegraphics[width=0.32\linewidth]{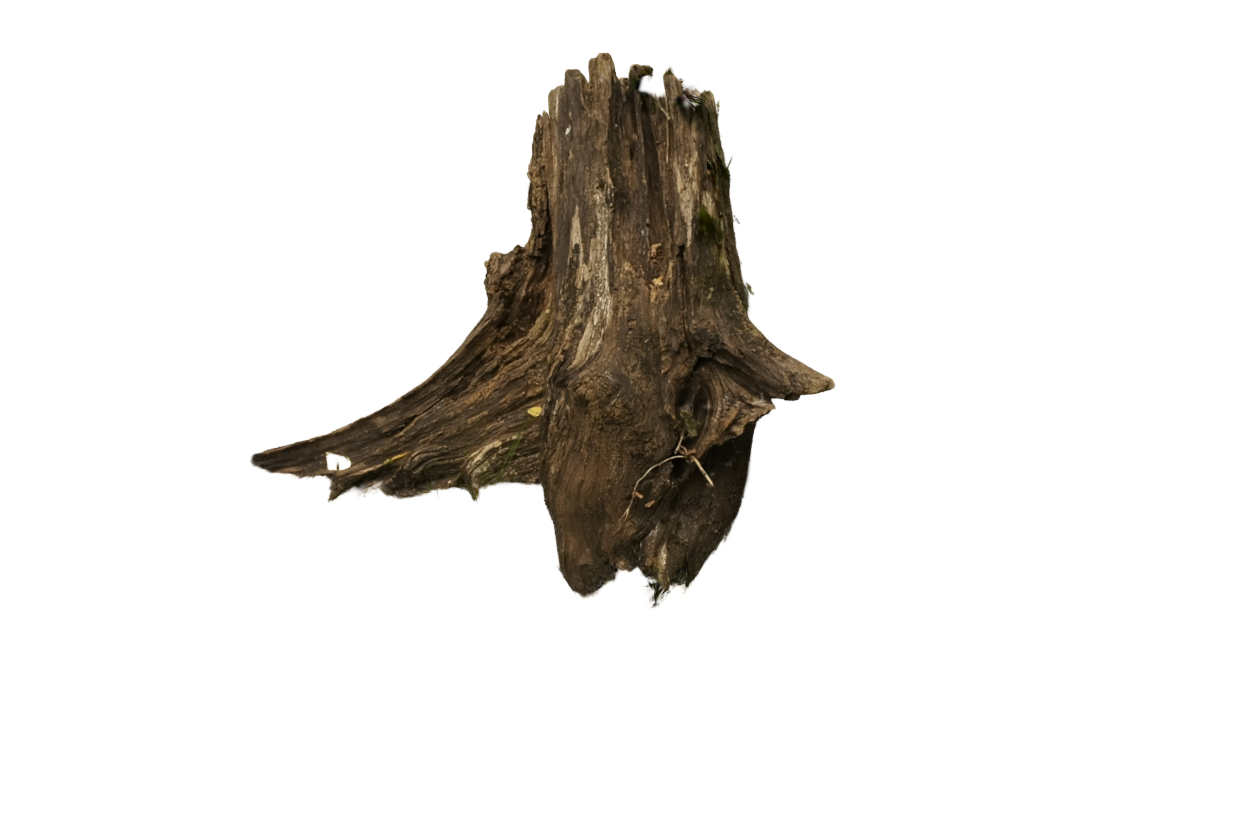} &
        \includegraphics[width=0.32\linewidth]{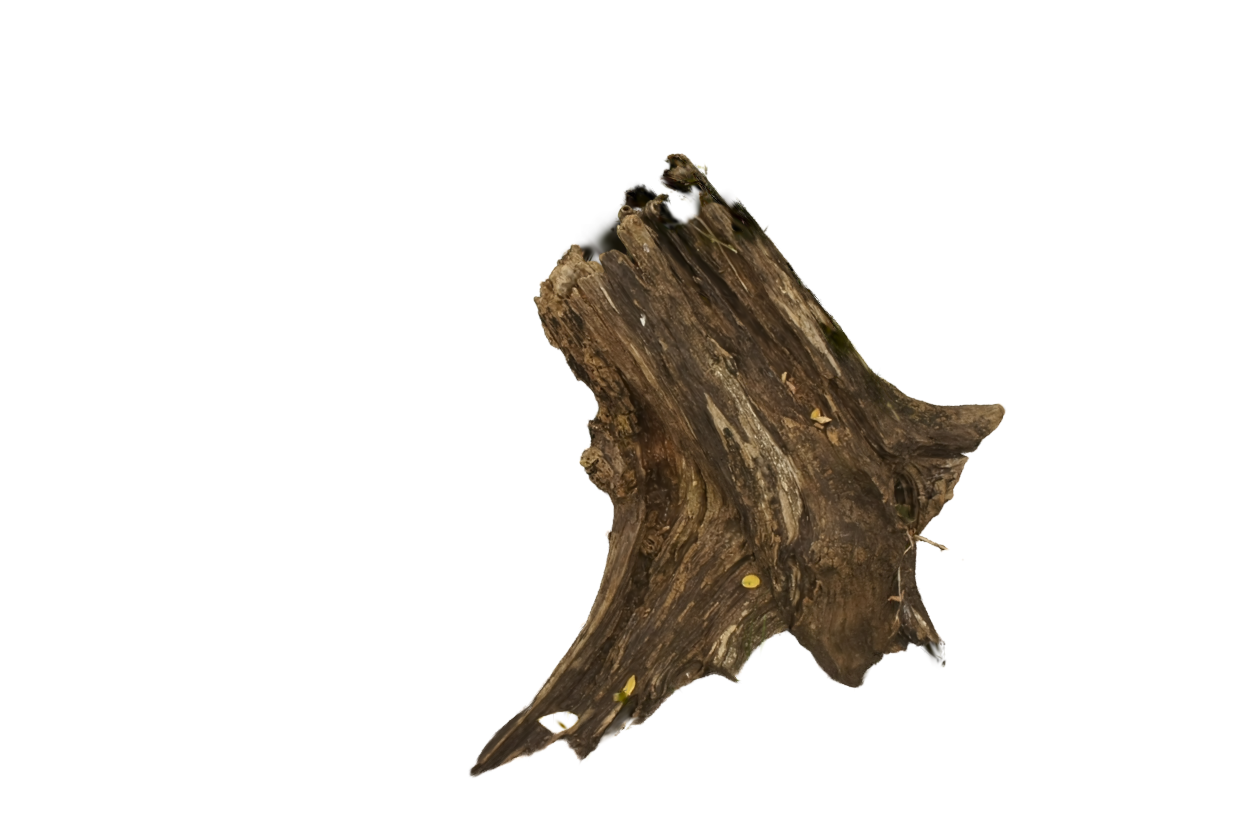} &
        \includegraphics[width=0.32\linewidth]{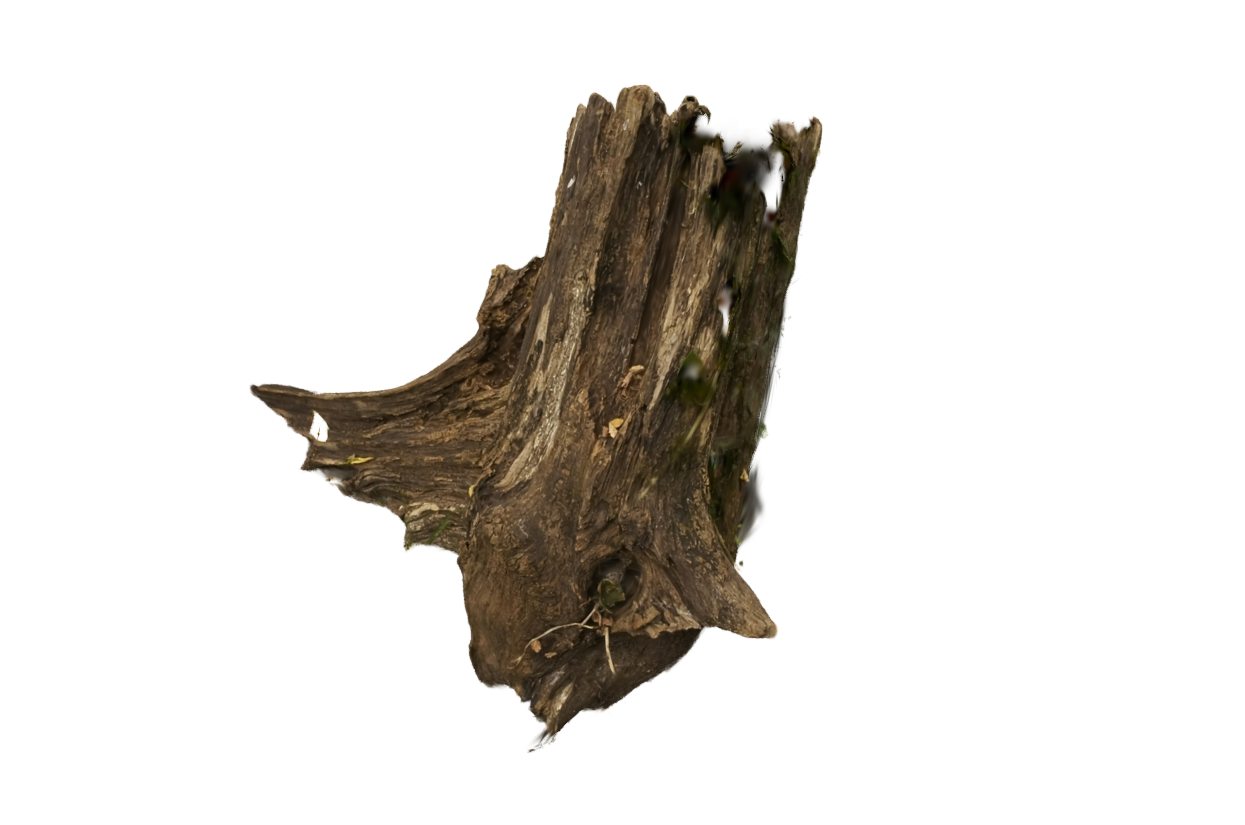} \\

         \includegraphics[width=0.32\linewidth]{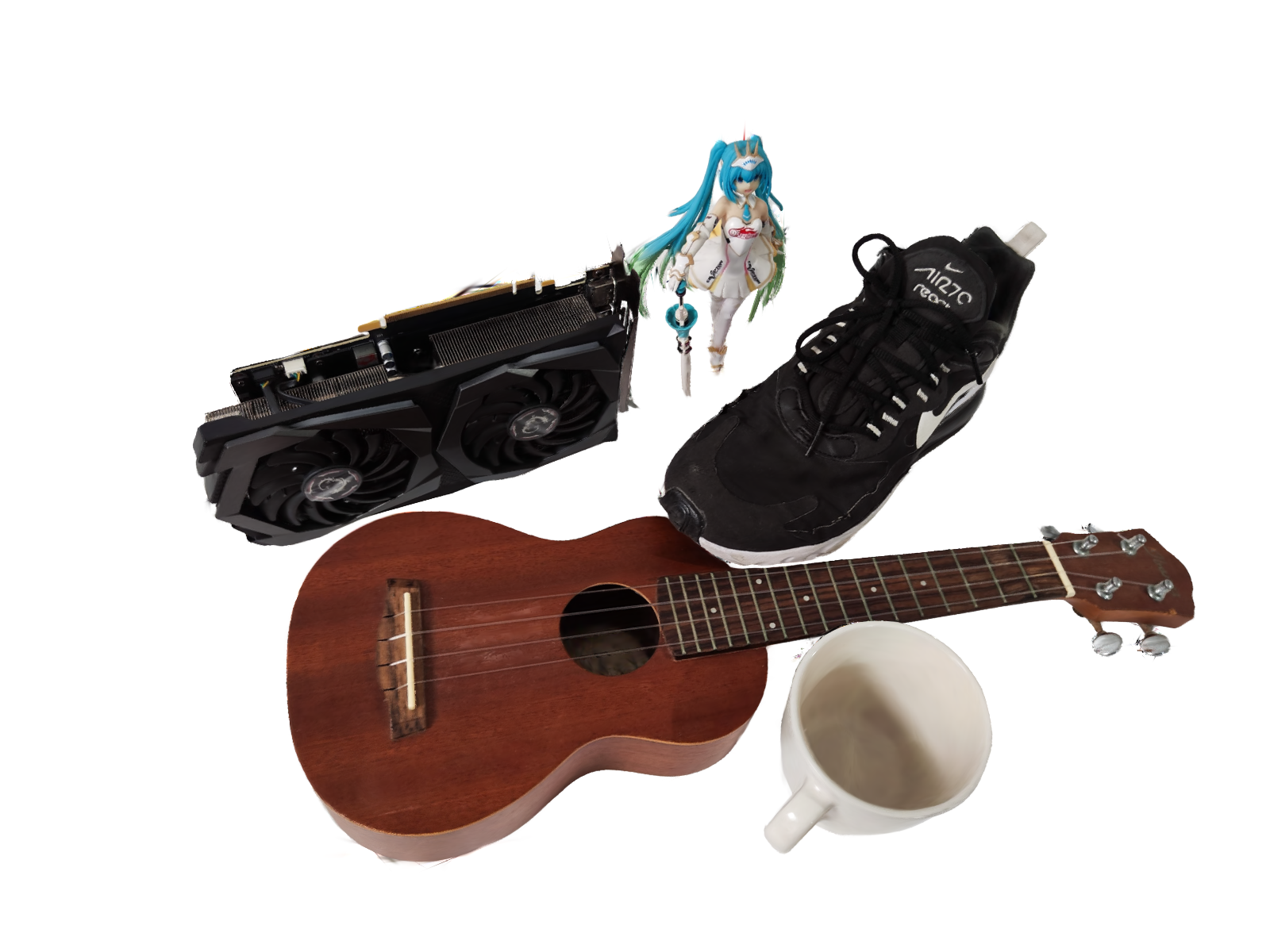} &
        \includegraphics[width=0.32\linewidth]{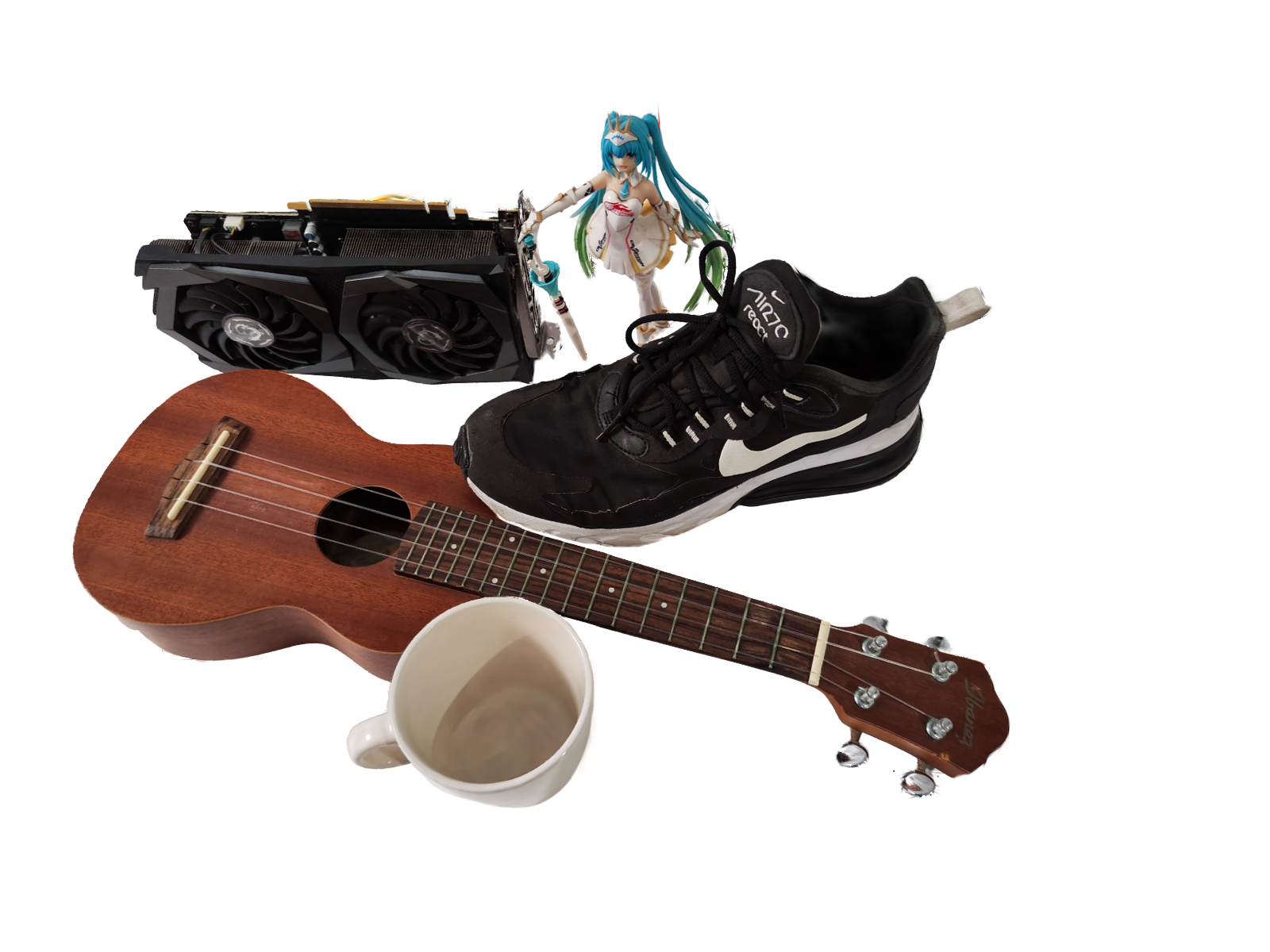} &
        \includegraphics[width=0.32\linewidth]{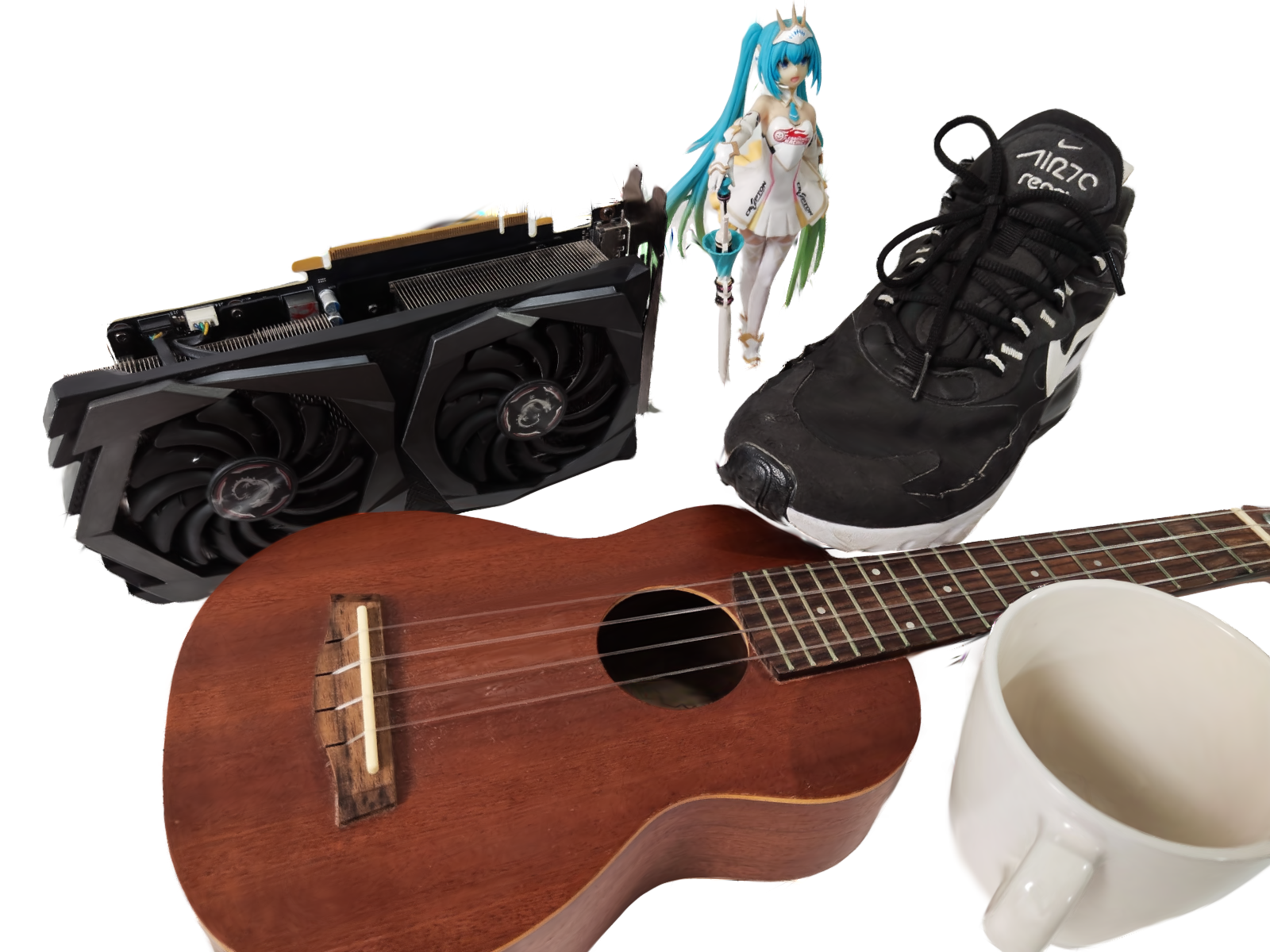} \\

         \includegraphics[width=0.32\linewidth]{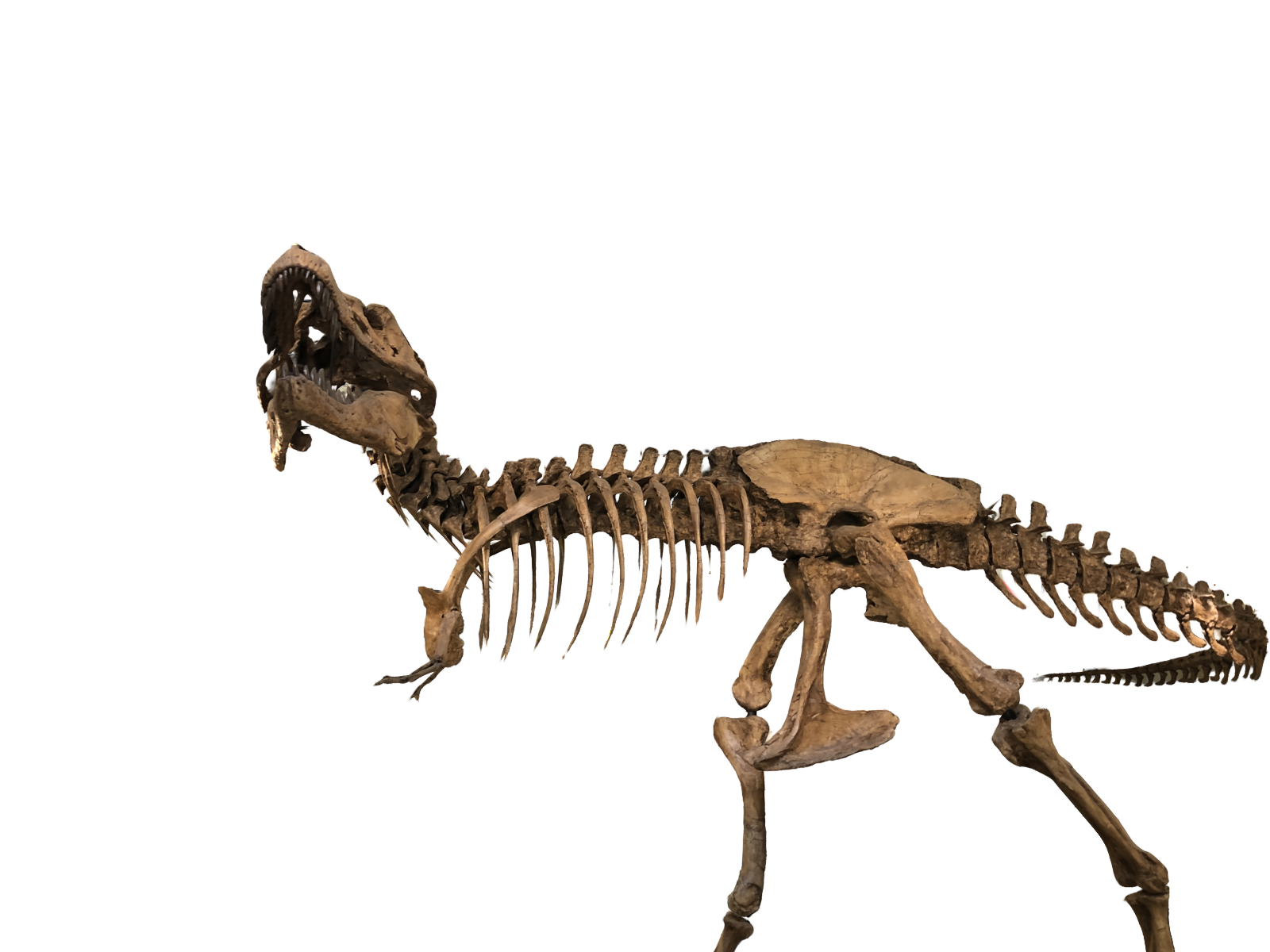} &
        \includegraphics[width=0.32\linewidth]{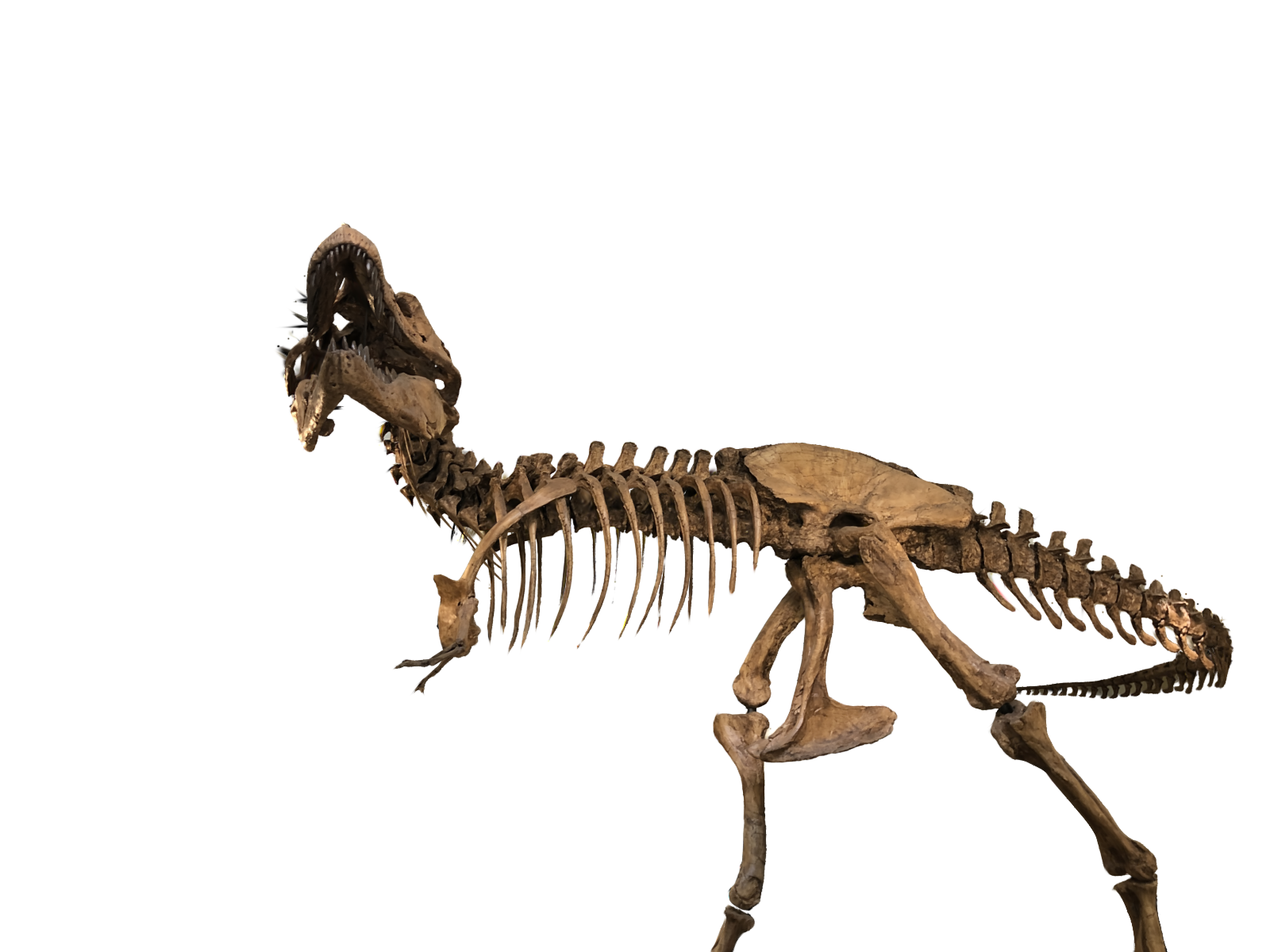} &
        \includegraphics[width=0.32\linewidth]{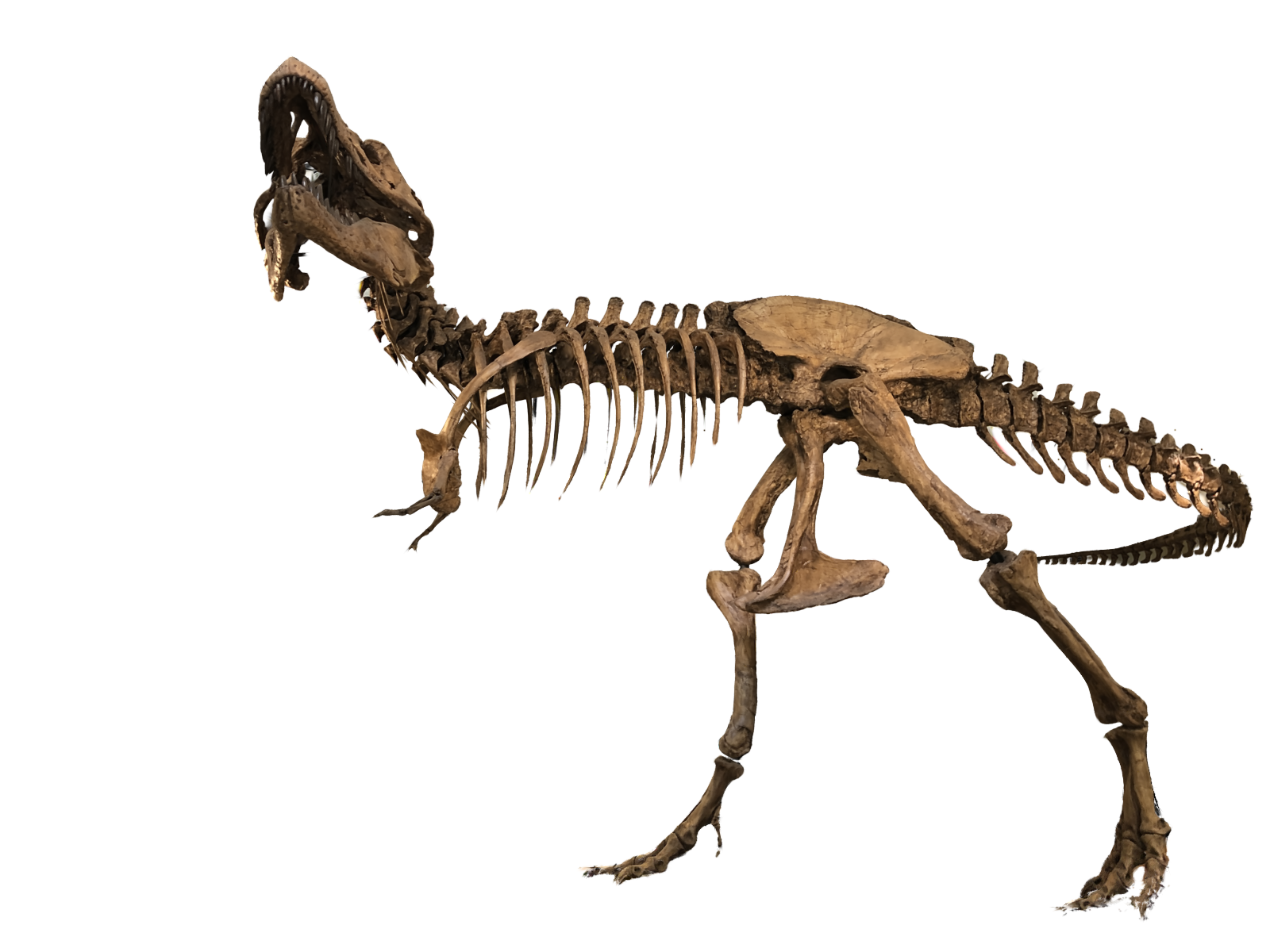} \\

         \includegraphics[width=0.32\linewidth]{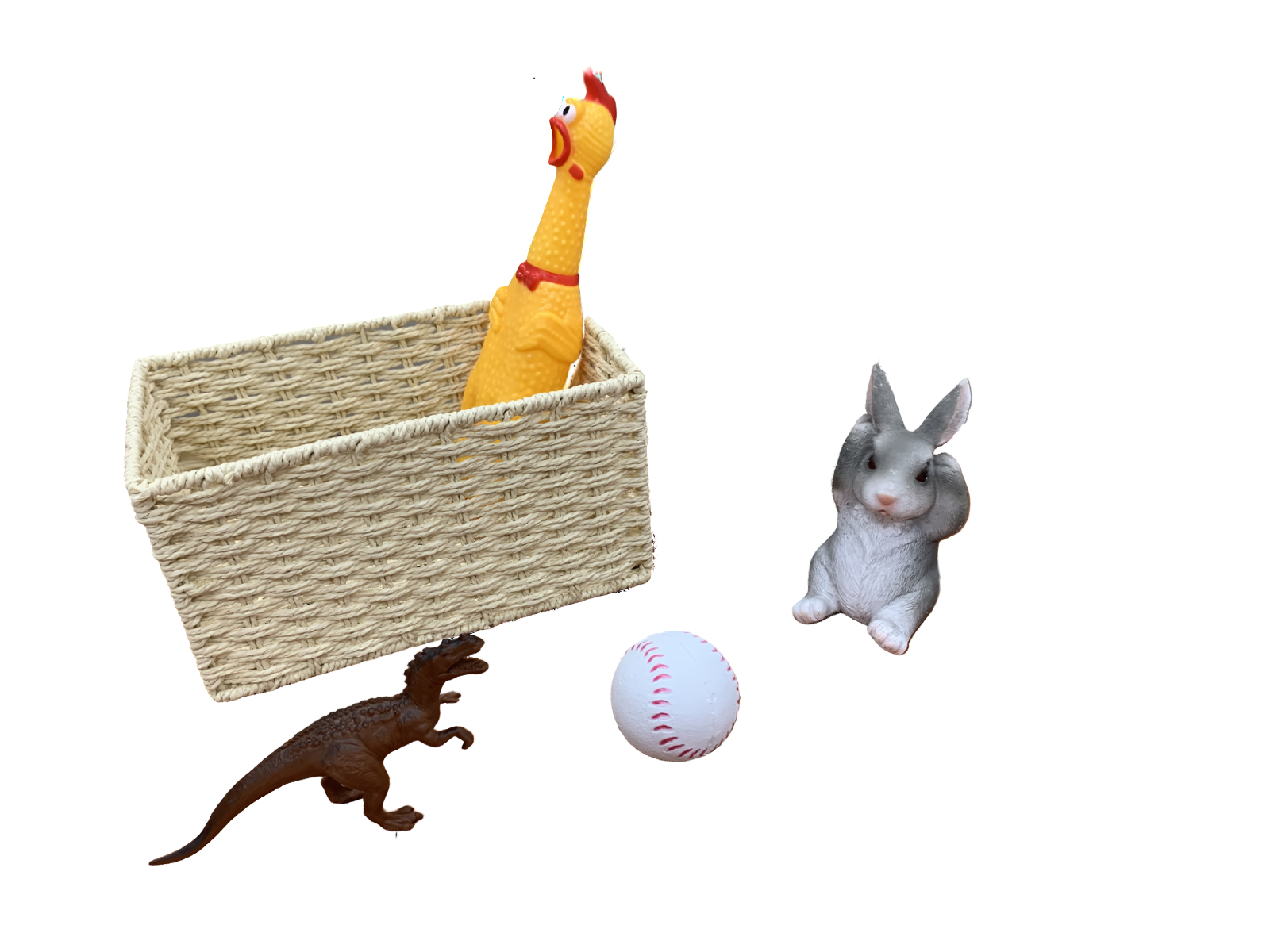} &
        \includegraphics[width=0.32\linewidth]{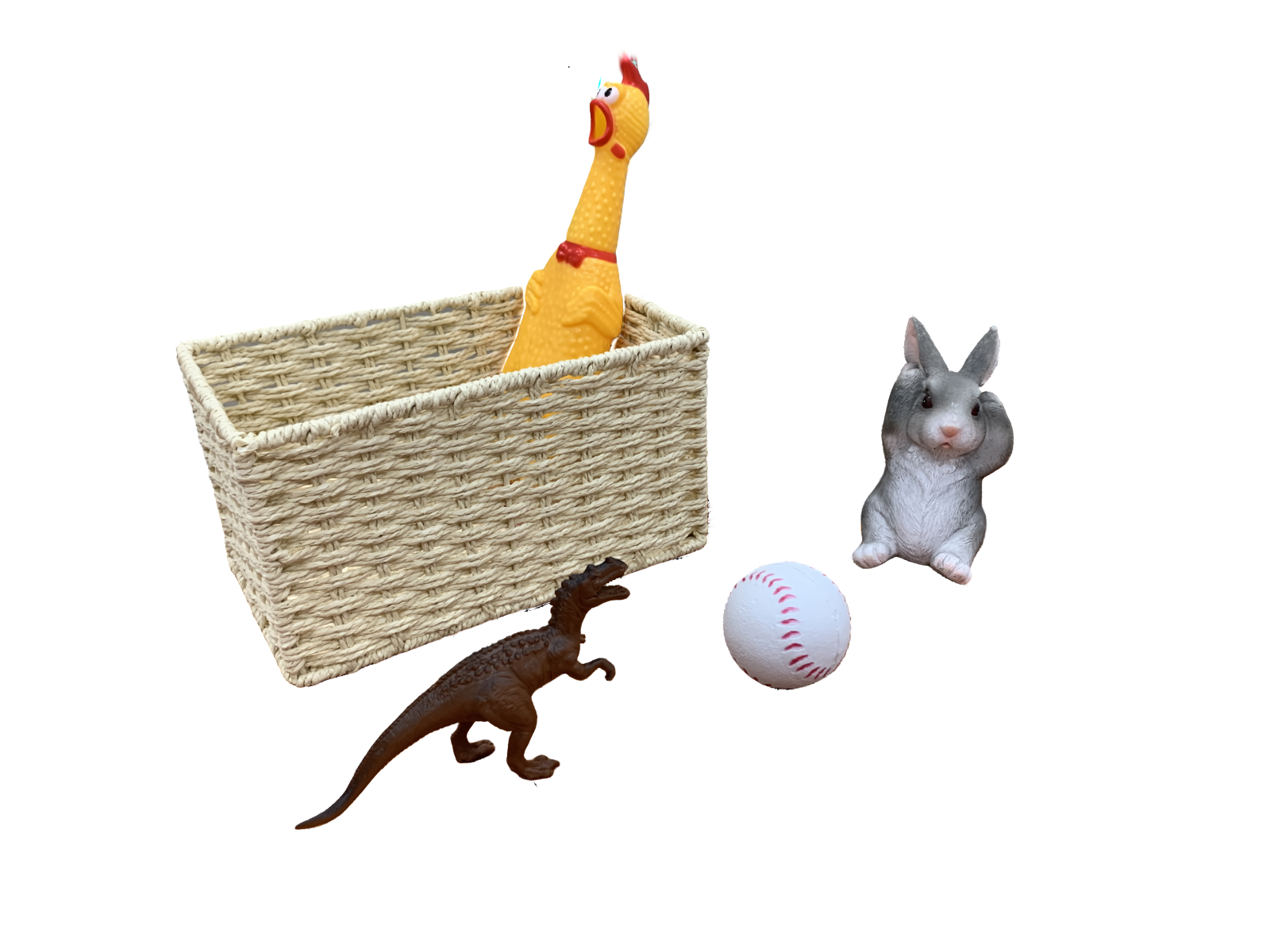} &
        \includegraphics[width=0.32\linewidth]{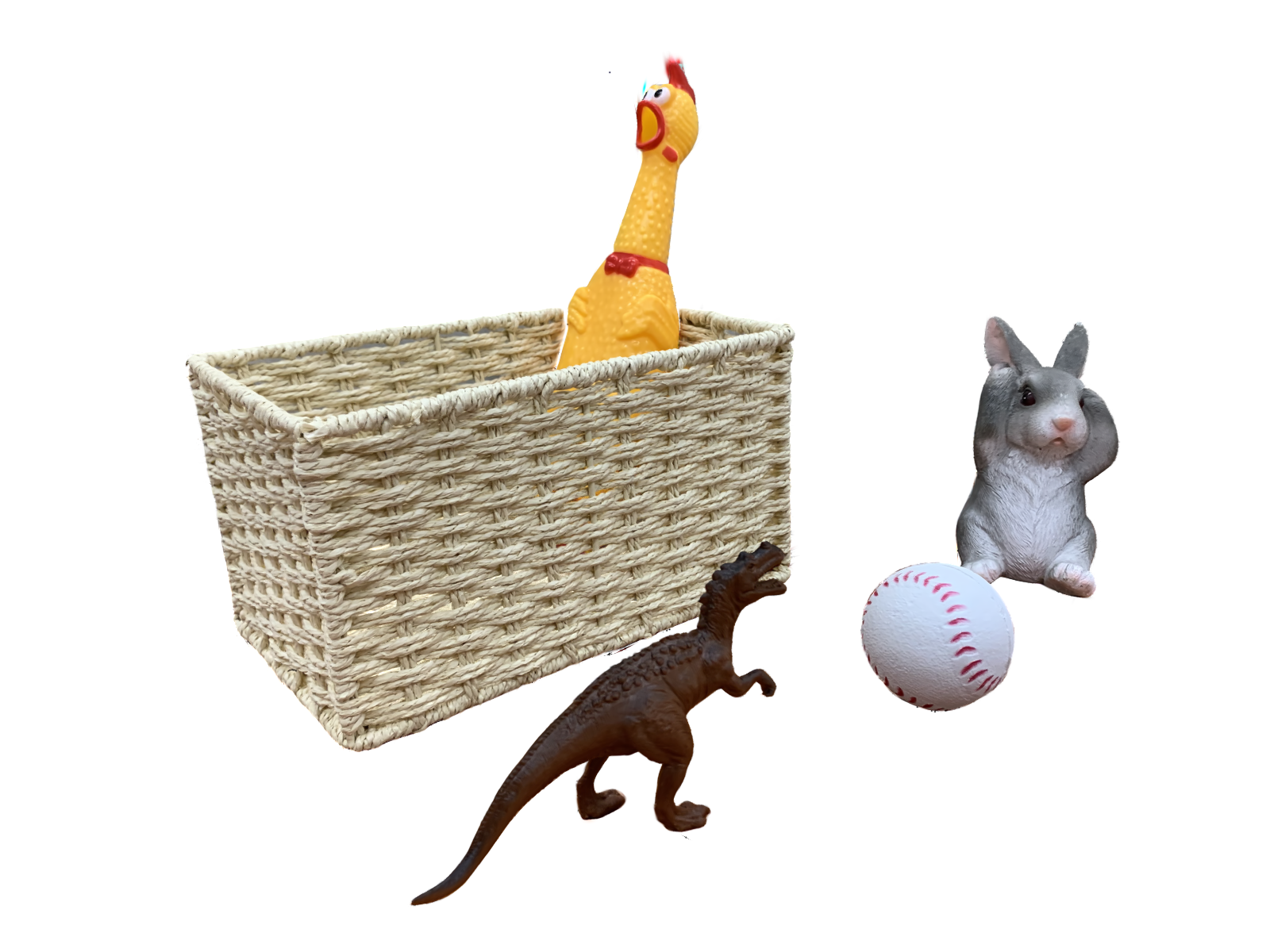} \\

        \includegraphics[width=0.32\linewidth]{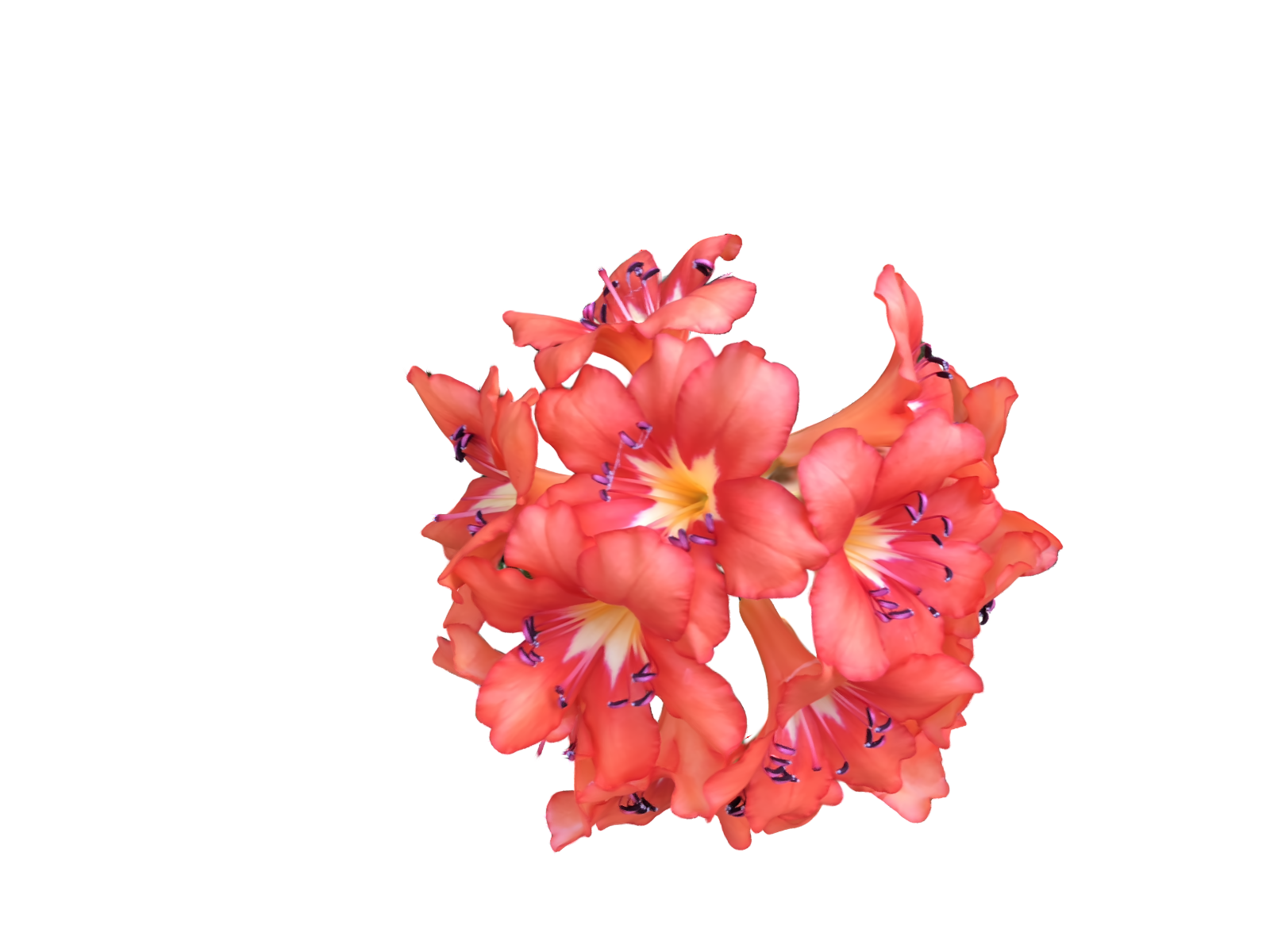} &
        \includegraphics[width=0.32\linewidth]{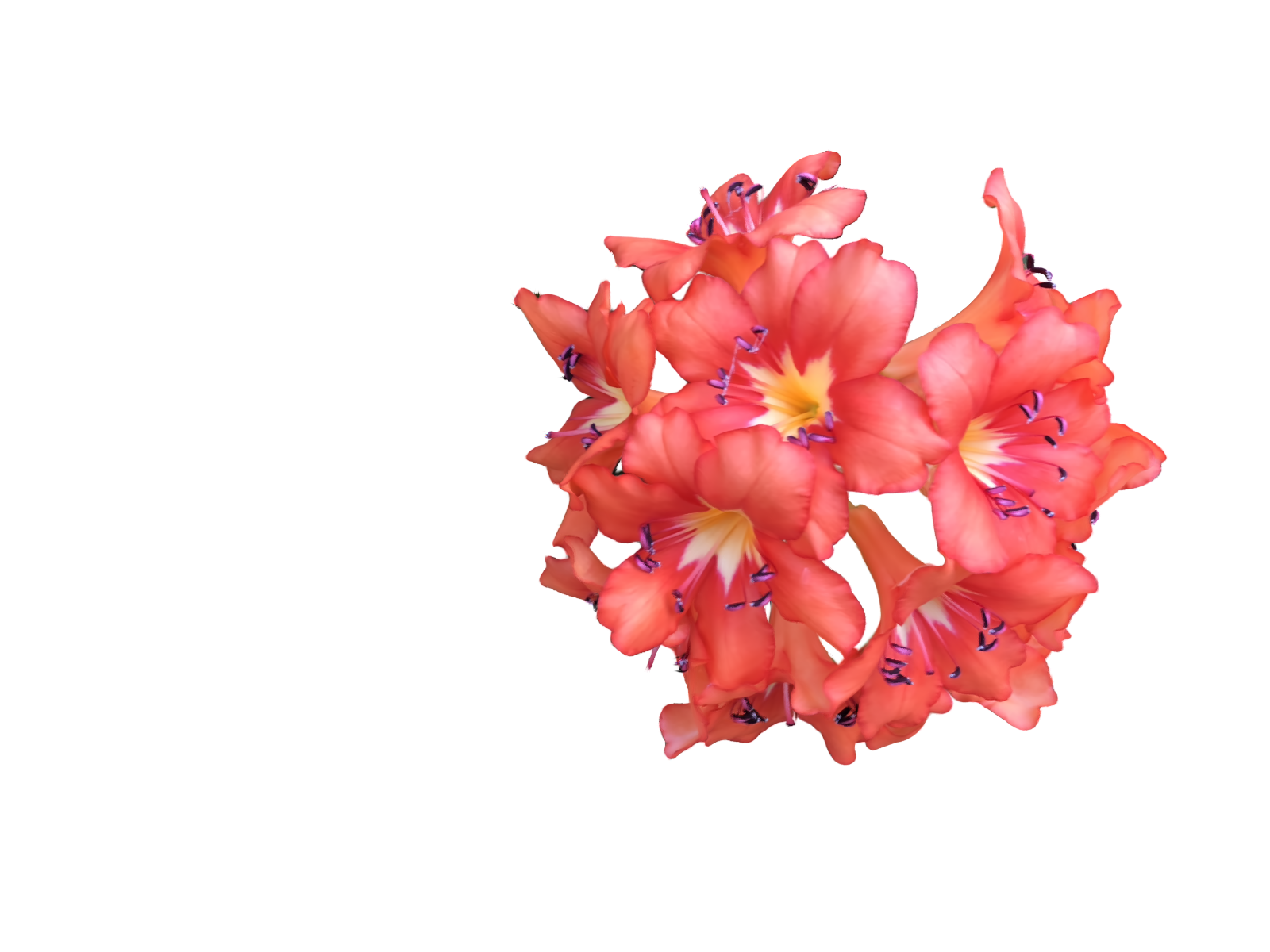} &
        \includegraphics[width=0.32\linewidth]{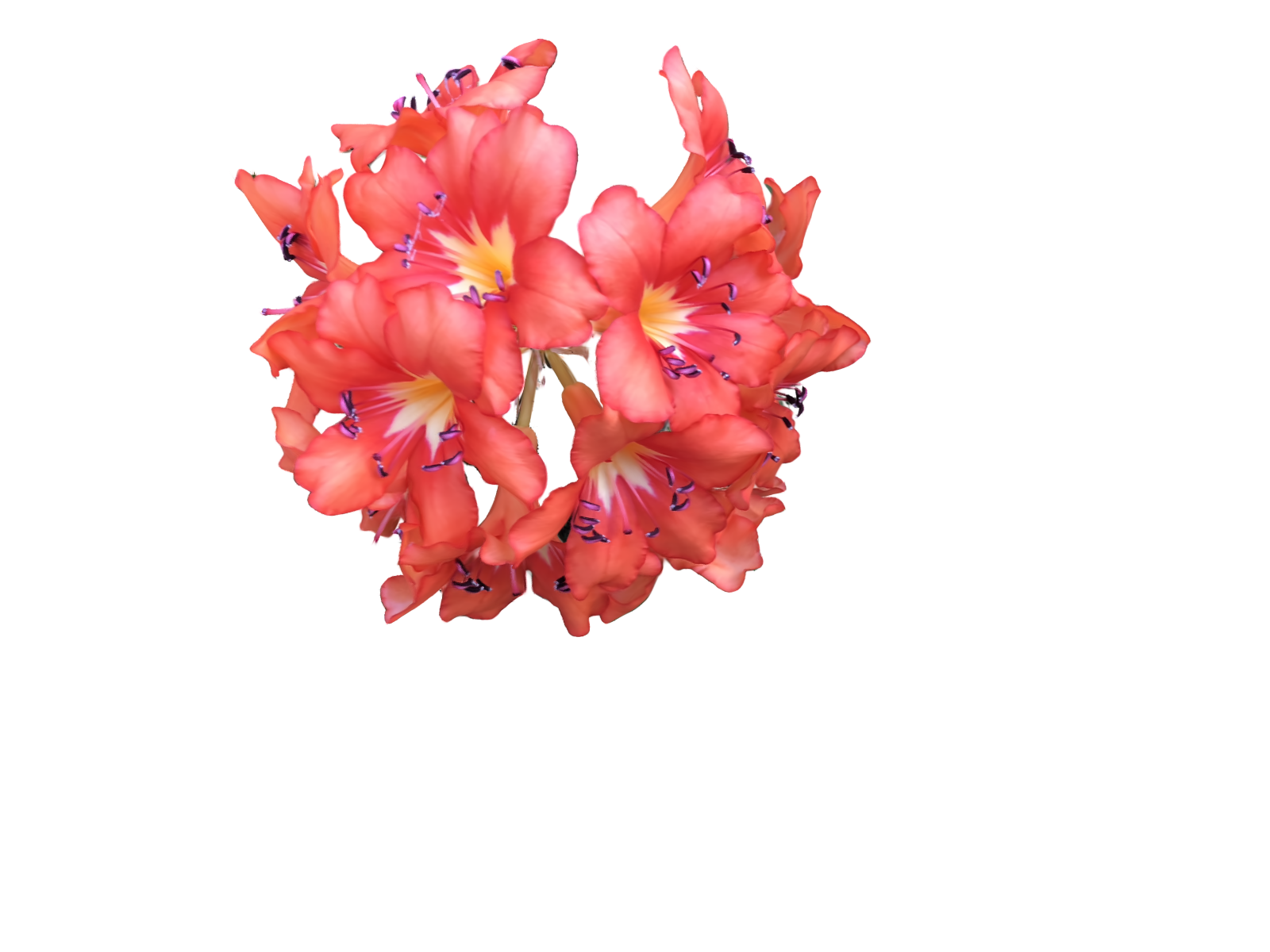} \\
        
    \end{tabular}

    \vspace{-2mm}
    \caption{Additional qualitative results obtained using our proposed approach}
    \label{fig:additional_qualitative3}
\end{figure*}

Due to space constraints in the main paper, we were unable to include as many qualitative results as desired. In Fig.~\ref{fig:additional_qualitative1}, Fig.~\ref{fig:additional_qualitative2}, and Fig.~\ref{fig:additional_qualitative3}, we therefore present additional objects extracted by our method, showing three rendered viewpoints for each scene. These examples demonstrate that our approach produces high-quality boundary segmentations across all four datasets: Mip-NeRF 360~\cite{barron2022mipnerf360}, LLFF~\cite{mildenhall2019llff}, LeRF~\cite{lerf2023}, and 3DOVS~\cite{3DOVS_dataset_paper}. For every scene, we remove the background and visualize the remaining object classes: multiple objects for LeRF and 3DOVS, and a single object for Mip-NeRF 360 and LLFF. The results highlight the method’s ability to recover fine-grained details, such as the T-Rex ribcage and the Lego bonsai leaves, which are challenging to infer from 2D segmentations alone. They also illustrate that non-visible Gaussians are no longer an issue during object extraction. Additional qualitative comparisons and results are included in the accompanying video.

%\begin{figure*}[htb]
%\begin{center}
%  \resizebox{\textwidth}{!}{\begin{tabular}{ l | c | r }
%    %\hline
%    \includegraphics{figures/qualitative_suppl_crop_resize/bed/bed1.png} & \includegraphics{figures/qualitative_suppl_crop_resize/bed/bed2.png} & \includegraphics{figures/qualitative_suppl_crop_resize/bed/bed3.png} \\ \hline
%    \includegraphics{figures/qualitative_suppl_crop_resize/bench/bench1.png} & \includegraphics{figures/qualitative_suppl_crop_resize/bench/bench2.png} & \includegraphics{figures/qualitative_suppl_crop_resize/bench/bench3.png} \\ \hline
%    \includegraphics{figures/qualitative_suppl_crop_resize/figurines/figurines1.png} & \includegraphics{figures/qualitative_suppl_crop_resize/figurines/figurines2.png} & \includegraphics{figures/qualitative_suppl_crop_resize/figurines/figurines3.png} \\
    %\hline
%  \end{tabular}}
%\end{center}
%\end{figure*}

\begin{comment}
MIP360 AVG 2DGS 30K TT 38.46 minutes -
MIP360 AVG 3K TT NO LOSSES 4.29 minutes -
MIP360 AVG 3K TT 4.29 minutes SAMP(0) -
MIP360 AVG 3K TT 5.3 minutes SAMP(1) -
MIP360 AVG 3K TT 5.57 minutes SAMP(5) -
MIP360 AVG 3K TT 6.42 minutes SAMP(20)

LERF AVG 2DGS 30K TT 40.3 minutes -
LERF AVG 3K TT NO LOSSES 2.5 minutes -
LERF AVG 3K TT 2.5 minutes SAMP(0) -
LERF AVG 3K TT 3.2 minutes SAMP(1) -
LERF AVG 3K TT 3.5 minutes SAMP(5) -
LERF AVG 3K TT 4.3 minutes SAMP(20)
\end{comment}

\end{document}